\documentclass{article}%\documentclass[a4paper]{article}
\usepackage{nac_preprint}

\usepackage[english]{babel}
\usepackage[utf8x]{inputenc}
\usepackage[T1]{fontenc}

\usepackage{comment}
\usepackage{booktabs}       % professional-quality tables
\usepackage{graphicx}
\usepackage{algorithm}
\usepackage{algorithmicx}
\usepackage[noend]{algpseudocode}
\usepackage{amssymb}
\usepackage{amsmath }
\usepackage{breqn}
\usepackage{multirow}
\usepackage{mathtools}
\usepackage{subcaption}
\usepackage{enumitem}
\usepackage{wrapfig}

\usepackage{amsbsy}

\usepackage{pifont}% http://ctan.org/pkg/pifont
\newcommand{\cmark}{\textcolor{green}{\ding{52}+}}%
\newcommand{\nmark}{\textcolor{blue}{\ding{51}}}%
\newcommand{\xmark}{\textcolor{red}{\ding{55}}}%

\algnewcommand{\LineComment}[1]{\State \(//\) #1}

\DeclareMathOperator*{\argmax}{arg\,max} % Jan Hlavacek

\usepackage[colorinlistoftodos]{todonotes}
\usepackage[colorlinks=true, allcolors=blue]{hyperref}

\title{Brain-Inspired Machine Intelligence: A Survey of Neurobiologically-Plausible Credit Assignment}

\author{%
  Alexander Ororbia \\
  Rochester Institute of Technology \\
  Rochester, NY 14623 \\
  \texttt{ago@cs.rit.edu}
}

\begin{document}

\setlength{\abovedisplayskip}{0.065cm}
\setlength{\belowdisplayskip}{0pt}
\maketitle

\begin{abstract}
\noindent
In this survey, we examine algorithms for conducting credit assignment in artificial neural networks that are inspired or motivated by neurobiology. These processes are unified under one possible taxonomy, which is constructed based on how a learning algorithm answers a central question underpinning the mechanisms of synaptic plasticity in complex adaptive neuronal systems: where do the signals that drive the learning in individual elements of a network come from and how are they produced? In this unified treatment, we organize the ever-growing set of brain-inspired learning schemes into six general families and consider these in the context of backpropagation of errors and its known criticisms. The results of this review are meant to encourage future developments in neuro-mimetic systems and their constituent learning processes, wherein lies an important opportunity to build a strong bridge between machine learning, computational neuroscience, and cognitive science.
%It is hoped that this review and unified treatment of the ever-growing set of brain-inspired credit assignment schemes will prove useful in guiding the creation of more efficient, novel learning procedures for neuro-mimetic intelligence. 

\small{}
\end{abstract}

%%% POPS: talk about what what has been done as opposed to what is done (and draw connections) - narrow focus on 2-3 questions and build taxonomy based on these --> contus chain of development of algorithms (80/90's continue to what is now) what builds on what and this what we need to do to make a major breakthrough (what does math/neuroscience tell us) -- bottom-up + top-down (as opposed to separate), looked at what others did and creatively changed it based on solidly proven concepts in past (or things that were abandoned and things that were there back in the day), focus on way of doing things and synthesize based on what answers looking for, what questions we should be answering, and what was found out -- nobody knows why or how to make things work better (no fundamental answers as to how and why are not given)
% bio-inspiration may answer the questions as to why it does work, there's something in machine we don't understand (or in our brain) now that we are learning how we think ourselves maybe we can make a better thinking machine

\section{Introduction}
\label{sec:intro}

One of the central goals that underpins research in biologically-inspired, or neuro-mimetic, machine intelligence is to create a complete theory of inference and learning that emulates how the brain can learn complex functions from the environment. Such a theory would be one that is not only biologically-plausible but also one that makes sense from a machine learning point-of-view \cite{bengio2017stdp}. Consequentially, this theory would be credible from both neuroscience and statistical perspectives on learning, and as a result, could be empirically tested and validated from both perspectives. Furthermore, progress in this burgeoning line of scientific inquiry could serve as a central basis for the grander goals of neuro-mimetic cognitive architectures \cite{eliasmith2012large,oreilly2016leabra,roelfsema2018control,ororbia2022cogngen,salvatori2023brain,bernaez2023incorporate} and the type of embodied brain-inspired machine intelligence behind the more recently cast ``NeuroAI'' initiative \cite{richards2019deep,payeur2021burst,zador2023catalyzing,momennejad2023rubric}.

A challenging aspect in constructing the aforementioned theory centers around the development of a plausible mechanism for conducting credit assignment -- in support of optimizing a behavioral scoring function(s) -- within computational neural systems. Credit assignment itself refers to the act of assigning ``credit'' and/or ``blame'' to individual processing elements, e.g., neuronal units, within a complex adaptive system based on their contribution to final behavioral output(s). With respect to networks of neurons, credit assignment is particularly difficult given that the impact or effect of early-stage neurons depends on downstream synaptic connections and neural activities. This challenge has also been referred to as the \emph{credit assignment problem} \cite{bengio1993credit}.

Although strong criticism of its biological plausibility has existed for several decades \cite{grossberg1987competitive,zipser1988back,crick1989recent,oreilly2000computational,harris2008stability,urbanczik2009reinforcement}, the algorithm known as backpropagation of errors \cite{linnainmaa1970representation,rumelhart1986learning}, or backprop, has been almost exclusively used to train modern-day, state-of-the-art artificial neural networks (ANNs) in supervised, unsupervised, and reinforcement learning tasks/problems. Though elegant and powerful in its own right, as well as one of the driving forces behind the ``deep learning revolution'' \cite{lecun2015deep}, it is one of the aspects of artificial neural computation that is the most difficult to reconcile with the current insights and findings that we have from cognitive neuroscience. Furthermore, many mechanisms and elements used to construct the computational architectures of current ANNs, such as normalization operations including batch and layer normalization \cite{ioffe2015batch,ba2016layer}, have been largely designed to address issues in credit assignment rather than acting as architectural features in service of solving the task at hand \cite{ororbia2019biologically,ororbia2023backpropagation}. As we will see throughout in our treatment of various alternative algorithms, although initially inspired by properties/behaviors of real neurons in the brain \cite{mcculloch1943logical}, even the processing elements that constitute deep neural networks (DNNs) %, including their underlying computation(s), 
omit many of the details that characterize actual neurobiological mechanisms and dynamics. Ultimately, considering the formalization and integration of other mechanisms and elements of computation that underlie biological neurons \cite{kording2000learning,kording2001supervised} might prove important in creating more powerful forms of credit assignment and facilitating more human-like generalization abilities in DNNs, addressing issues such as reliability/model calibration, robustness, and sample inefficiency. It is the view of this work that biological plausibility in the learning, inference, and design of artificial neural systems is not solely a (niche) property of interest to neuroscientists and cognitive scientists; it will play critical a role in the future of machine intelligence, such in efforts that seek implementation on low-energy analog and neuromorphic chips \cite{davies2018loihi,grollier2020neuromorphic,kumar2022dynamical,yi2023activity}. Furthermore, these biologically-inspired computational frameworks will need to complemented with examination and evaluation  from a behavioral point-of-view \cite{bartunov2018assessing,ororbia2023backpropagation,lillicrap2020backpropagation}, either considering how approaches to credit assignment scale to higher-dimensional, complex tasks \cite{bartunov2018assessing} or investigating how particular algorithms generalize in the context of modular cognitive architectures \cite{ororbia2022cogngen,ororbia2022maze}. As a result, new forms of analysis and benchmarking, enriched by ideas and concepts from computer science, cognitive science, and computational neuroscience \cite{linsker1988self,marblestone2016toward,richards2019deep,chavlis2021drawing}, will be needed to make consistent progress as well as new breakthroughs that advance the current state of research towards flexible, robust brain-inspired intelligent systems. %In service of this argument, this survey will highlight directions that could advance the current state of neuro-mimetic systems in progressing towards flexible, robust brain-inspired intelligent systems, enriched by ideas and concepts from computer science, cognitive science, and computational neuroscience \cite{linsker1988self,marblestone2016toward,richards2019deep,chavlis2021drawing}.

Concretely, this review will focus on algorithms that have been proposed over the past several decades to train ANNs without backprop; these methodological frameworks have sometimes referred to as ``backprop-free'' or ``biologically-inspired'' learning algorithms. In order to better compare and organize the ever-growing plethora of approaches, we construct a taxonomy based on answering one of the key questions that centrally motivates the development of biologically-plausible forms of credit assignment: 
%\textit{Where do the teaching signals needed for adapting the processing elements in a network come from?} 
\textit{Where do the driving forces, or signals, behind the synaptic plasticity needed for adapting a network's processing elements come from and how are they produced?} The way that backprop answers this question, as we will shall soon examine, is what stands in contrast to many theories of the brain \cite{grossberg1982does,rao1999predictive,huang2011predictive,clark2013whatever}. In this survey, algorithm clusters/families (``themes'') will be formulated with respect to how various schemes attempt to answer this question; in the scope of this work, this yields six different families. Along the way, we will consider how, and the extent to which, different biological algorithms address or resolve the neurophysiological and engineering criticisms of backprop. It is hoped that the taxonomy and unified treatment that we present of extant algorithmic proposals for biologically-motivated credit assignment will inform, inspire, and aid the field in generating new ideas that extend, combine, or even supplant current methods. 
%Attempting to build models and algorithms that resolve some of the above criticisms, from both biological and practical perspectives, might open the door to learning approaches that generalize better, allowing us to tackle even greater open challenges, such as that of lifelong (machine) learning.
% I could, at the end, even point to how lifelong learning/catastrophic forgetting might be resolved by even just considering a better form of credit assignment that is better connected to neuro-physiology since we know that humans generally only experience gradual forgetting and not massive interference

\noindent
\textbf{Structure of the Survey.} This article is organized in the following manner. First, to contextualize our review, we provide, in Section \ref{sec:framing}, our survey's notation/symbology, its running-example neural architecture, and a characterization of backpropagation of errors (backprop); we furthermore elucidate backprop's core problems and sources of biological implausibility. Following this, in Section \ref{sec:taxonomy}, we present our framing question and the corresponding taxonomy it induces to organize the approaches we will examine. In Section \ref{sec:algo_families}, we then review the various families of neuro-mimetic credit assignment from the perspective of our taxonomy. To conclude, in Section \ref{sec:discussion}, we then move to synthesize the results in terms of addressing the issues inherent to backprop as well as highlight several important future directions for research in neurobiological credit assignment and brain-inspired machine intelligence.
%we consider available software tools and libraries that support bio-CA (neuroAI) and discuss how research on hardware could realize the full potential bio-CA -- in appendix/supplement

\subsection{Algorithmic Framing}
\label{sec:framing}

\noindent 
\textbf{Notation.} We start by defining key notation and symbols that will be commonly used throughout this survey (see the supplement for tables containing symbol/operator and acronym definitions). A capital bold symbol $\mathbf{M}$ represents a matrix while a lowercase bold one $\mathbf{v}$ is a vector; note that $M_{ij}$ retrieves a scalar at position $(i,j)$. A matrix-matrix/vector multiplication is denoted by $\cdot$, a Hadamard product is $\odot$, and $(\mathbf{v})^{\mathsf{T}}$ represents the transpose of $\mathbf{v}$. We will represent elementwise functions (applied to matrices or vectors) in terms of notation $g(\mathbf{v})$ with the first derivative (with respect to its input argument) denoted as $\partial g(\mathbf{v}$). %In the supplementary material, we provide a table containing key symbols and abbreviations used throughout this review article.

\begin{figure}[!t]
\centering  
%\vspace{-5mm}
\includegraphics[width=1.0\linewidth]{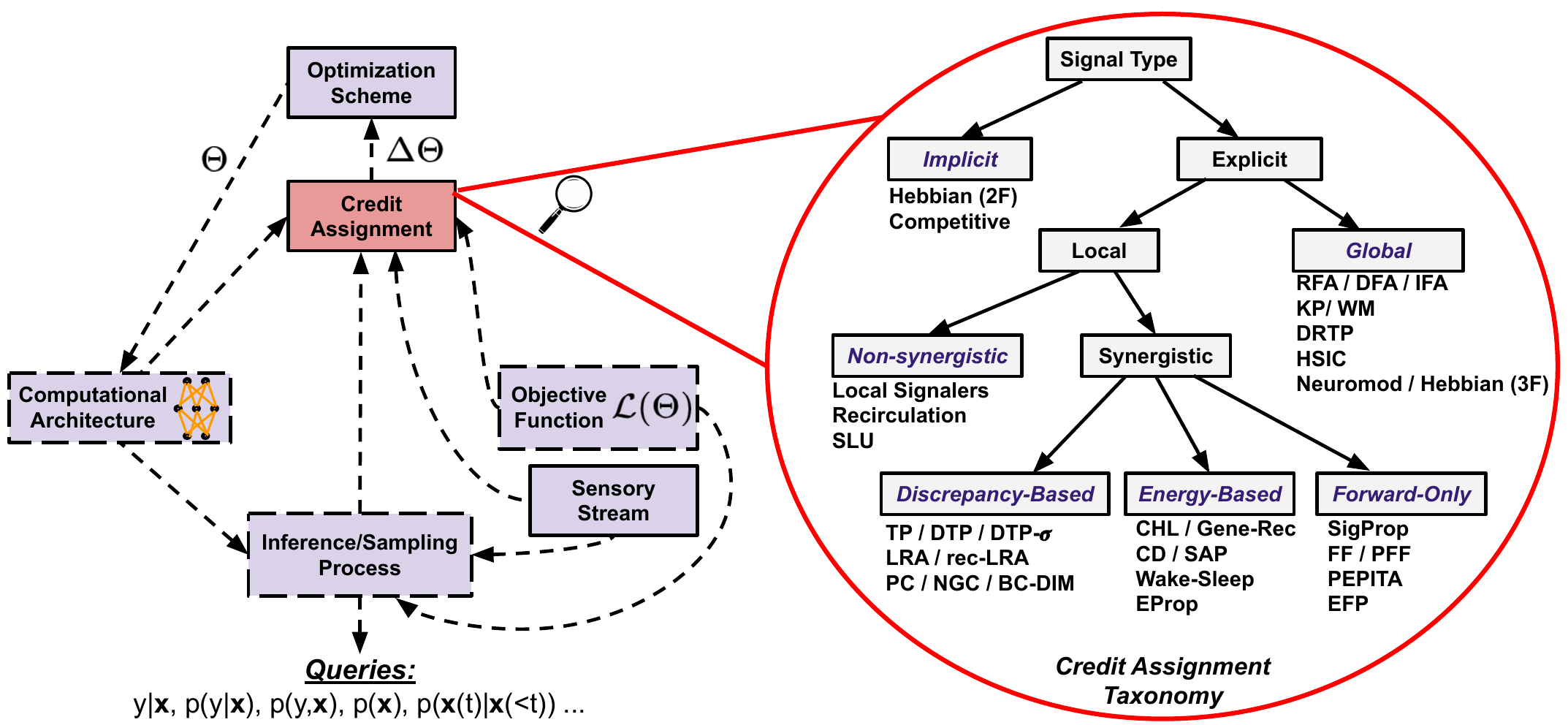}
%\caption{Neural system pipeline and learning algorithm taxonomy.}
%\vspace{-0.3cm}
\caption{\small{
\textbf{The neural system context and the proposed algorithm taxonomy shown.} Our proposed credit assignment taxonomy (see red zoom-in) centers around the question: \textit{What are the driving forces or  signals that underlie synaptic plasticity?} Components in dashed box outlines indicate possible (direct or indirect) involvement in the design of the credit assignment. Purple italicized text in the taxonomy indicate a leaf or algorithm family. 
Note: See supplement for acronym definitions. 
}}
\label{fig:system_and_taxonomy}
\vspace{-0.5cm}
\end{figure}

% CONNECTION to PDP book/literature !!!!!!!!!
% Pops comment - organize list below into bullets...
\noindent
\textbf{Algorithmic Framing: The Neural System Context.} To contextualize and unify the various learning processes studied in this review, we introduce what we call the `neural system context': this is the neuronal model framing and information-processing pipeline that a credit assignment scheme would be situated and operate within. We consider a neural system context to be one that specifies and implements (at least) the following elements: 
\begin{itemize}[noitemsep,nolistsep]
    \item a computational architectural consisting of processing elements or set of interacting neuronal constructs, e.g., such as an MLP graphical model (including its weight initialization scheme);
    \item an inference or information communication procedure across its elements, e.g., forward propagation of activities in an ANN model;
    \item a credit assignment process which calculates/produces updates/adjustments to (synaptic) parameters (the learning algorithm), e.g., backprop in the case of a standard DNN;
    \item a parameter optimization function, or update rule, that takes in updates provided by the credit assignment process and directly changes the values of the parameters $\Theta$, e.g., stochastic gradient descent (SGD), RMSprop \cite{tieleman2012lecture}, Adam \cite{kingma2014adam}; 
    \item a problem-specific (global) cost function, usually determined by the desired task; 
    \item a sensory stream, which could be a collection of patterns, as in a dataset $\mathcal{D}$, or those drawn online from a streaming data generating process.
\end{itemize}
The above points are graphically depicted in Figure \ref{fig:system_and_taxonomy}, and as indicated by the red-tinted box ``credit assignment'', this survey will focus on the learning process of a neural system context. %, i.e., the algorithm that produces adjustments to parameters. 
As indicated by the dashed boxes, the ``computational architecture'', ``task objective'', and ``inference/sampling process'' will also be considered to the extent in which they are involved in or influenced by the learning scheme.\footnote{This relationship is what we will refer to later as the \emph{degree of entanglement}.} However, we remark that other components not only provide the context for the learning and inference processes but, depending on setup, can significantly affect their performance/behavior, e.g., choice of parameter optimization scheme, e.g., RMSprop versus Adam.

\begin{wrapfigure}{r}{0.5\textwidth}
\vspace{-0.455cm}
  \begin{center}
    \includegraphics[width=0.4\linewidth]{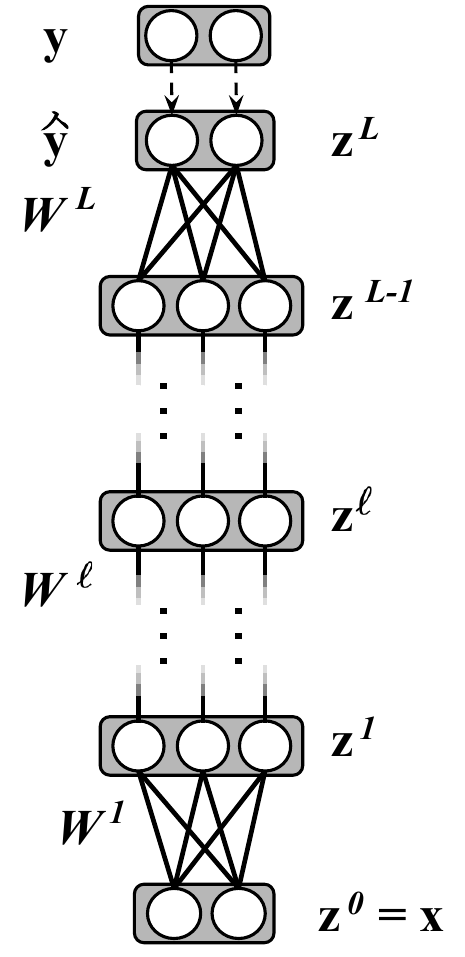}
  \end{center}
  %\vspace{-0.255cm}
  \caption{\small{
  An example MLP, processing sample $(\mathbf{y},\mathbf{x})$.
  }}
  \label{fig:mlp}
  \vspace{-0.25cm}
\end{wrapfigure}

\noindent 
\textbf{The Computational Architecture Running Example.} Here, we will present the base form of our running example of the computational neural architecture that backprop, as well as brain-inspired alternatives, perform credit assignment with respect to.\footnote{The caveat here is that some algorithm families examined in this survey essentially alter the nature/form of this architecture.} Specifically, we examine the popularly-used ANN structure known as the feedforward network; see Figure \ref{fig:mlp}. In essence, a feedforward ANN can be represented as a stack of nonlinear transformations, or $f_\Theta(\mathbf{x}) = { \{f_\ell(\mathbf{z}^{\ell-1};\theta_\ell)\}^L_{\ell=1} }$, where the input is $\mathbf{z}^0 = \mathbf{x}$. The dimension of any vector within this construct is $\mathbf{z}^\ell \in \mathbb{R}^{\mathcal{J}_\ell \times 1}$ while $\mathbf{x} \in \mathbb{R}^{\mathcal{J}_0 \times 1}$ (assuming a mini-batch size of one) and $\mathbf{y} \in \mathbb{R}^{C \times 1}$. If we further simplify the ANN to be a multilayer perceptron (MLP), each transformation $\mathbf{z}^\ell = f_\ell(\mathbf{z}^{\ell-1})$ produces an output $\mathbf{z}^\ell$ from the value $\mathbf{z}^{\ell-1}$ of the previous layer using a matrix of synaptic weight values $ \theta_\ell = \{\mathbf{W}^{\ell}\}$ where $\mathbf{W}^\ell \in \mathbb{R}^{\mathcal{J}_{\ell} \times \mathcal{J}_{\ell-1}}$. Each layer-wise function $f_\ell$ of the MLP $f_\Theta(\mathbf{x})$ is made up of two operations (biases omitted for clarity):
\begin{align}
\mathbf{h}^\ell &= \mathbf{W}^{\ell} \cdot \mathbf{z}^{\ell-1}, \quad \mathbf{z}^\ell = \phi^\ell(\mathbf{h}^\ell) \,,\label{eqn:mlp}
\end{align}
where $\phi^\ell$ is an elementwise activation function, ${ \mathbf{z}^\ell \in \mathcal{R}^{J_\ell} }$ is the post-transformation of layer $\ell$, and ${ \mathbf{h}^\ell \in \mathbb{R}^{\mathcal{J}_\ell} }$ is the pre-transformation vector of layer $\ell$.\footnote{In this survey, we will utilize ``pre-activation'' to refer to an incoming vector/set of neural activity values, i.e., $\mathbf{z}^{\ell-1}$, and ``post-activation'' to refer to an outgoing vector/set of neural activity values, i.e., $\mathbf{z}^\ell$} Note that this computational framing is treating neurons as rate-based units; the activity of any neuron $i$ in a layer $\ell$ of a network is described by its real-valued firing rate $z^\ell_i$. The final output prediction made by the MLP $f_\Theta(\mathbf{x})$ is $\mathbf{z}_L$. The central goal of any credit assignment process will be adjust the values within $\Theta$ so as minimize a task-specified loss/cost functional, i.e., a function that evaluates the quality of $f_\Theta(\mathbf{x})$'s data-fitting ability and overall performance on a given task. Note that the parameters of the entire network would be written out as $\Theta = \{ \mathbf{W}^1,...,\mathbf{W}^\ell,...,\mathbf{W}^L\}$ and that some learning algorithms might introduce extra parameters beyond those of the original architecture, e.g., inverse mapping or error feedback synapses. %\footnote{Note that each internal neuron does not need to map perfectly to a single neuron in the brain. It could, for example, map to a group of neurons that compose a cortical microcircuit.}

%%%%%%%%%%%%%%%%%%%%%%%%%%%%%%%%%%%%%%%%%%%%
% MLP computational arch figure
%%%%%%%%%%%%%%%%%%%%%%%%%%%%%%%%%%%%%%%%%%%%

The activation function for the output layer $L$, as well as the cost function, of a feedforward network are chosen based on the problem at hand. For instance, in the case of regression over unbounded continuous values, an identity activation function is typically employed, i.e.,  $\mathbf{z}^L = \phi^{L}(\mathbf{h}^{L}) = \mathbf{h}^{L}$, paired with a squared error cost function, i.e., $\mathcal{L}(\mathbf{y},\mathbf{z}^L) = \frac{1}{2}\sum_i (\mathbf{z}^L_i - \mathbf{y}_i)^2$. For task of classification, the activation function of the output is generally set to be the softmax: $p(\mathbf{y}|\mathbf{z}^L) ={ \phi_L(\mathbf{z}^L) = \exp(\mathbf{z}^L) / ( \sum_j \exp(\mathbf{z}^L_j ) }$. Any element in the output vector, i.e., $\mathbf{y}_j \equiv { \phi_L(\mathbf{v})_j = p(j|\mathbf{v}) }$, is the specific scalar probability for class $j$. In this case, the cost function typically chosen is the categorical (multinoulli) negative log-likelihood: ${ \mathcal{L}(\mathbf{y}, \mathbf{z}^L) = -\sum_i \big( \mathbf{y} \odot \log p(\mathbf{y}|\mathbf{z}^L) \big)_i }$. Note that $\mathbf{y}$ is the task-specific target vector (sometimes this will be referred to as the `context' vector), which could be a label/regression target, as is common in supervised learning setups, or even the data input itself, as is the case for some unsupervised learning contexts.

\subsection{Backpropagation of Errors} 
\label{sec:backprop}

\begin{figure}[!t]
\centering     %%% not \center
\includegraphics[width=0.475\linewidth]{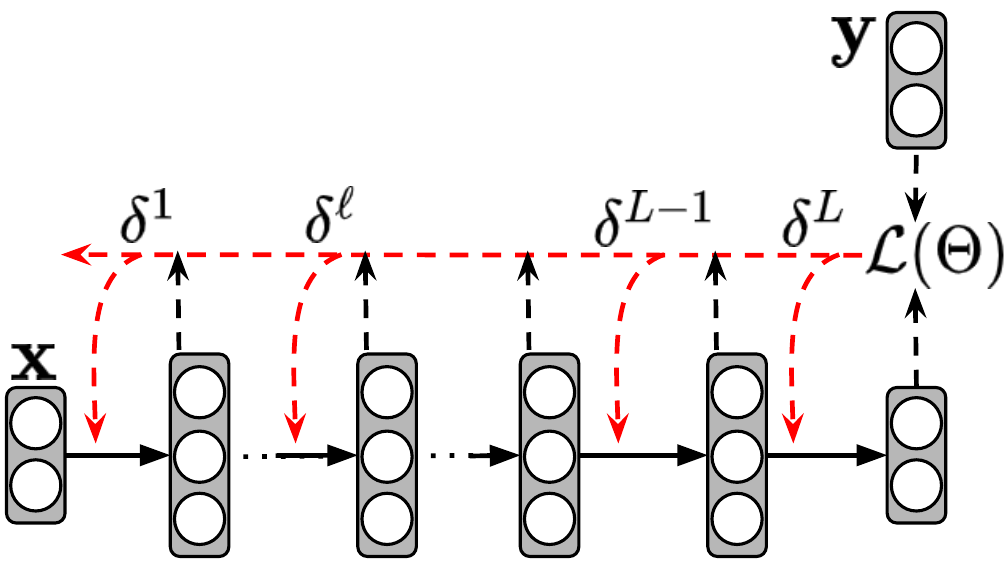}
%\vspace{-0.3cm}
\caption{\small{\textbf{The global feedback pathway of backpropagation of errors.} Depicted is the global feedback pathway, or backward flow of recursively computed teaching signals, that characterizes backprop-based credit assignment.
}}
\label{fig:global_feedback_pathway}
\vspace{-0.5cm}
\end{figure}

According to backprop-based adaptation \cite{linnainmaa1970representation,rumelhart1986learning}, and its many variations \cite{fahlman1988faster,riedmiller1993direct}, to conduct credit assignment in a model such as the MLP, we take gradients of $\mathcal{L}(\mathbf{y}, \mathbf{z}^L)$ with respect to each matrix in $\Theta$, i.e., for each synaptic matrix, our goal is to obtain another matrix $\frac{\partial \mathcal{L}(\mathbf{y}, \mathbf{z}^L)}{\partial \mathbf{W}^\ell}$ which contains synaptic adjustments; this adjustment matrix can then be used in an optimization rule such as SGD. To obtain these necessary adjustments, we employ the chain rule of calculus, i.e., reverse-mode differentiation \cite{baydin2018automatic}, starting by computing gradients of the cost function (at the output) and moving recursively backward through the computational graph of operations that defines the network. Working back through the consequently long chain of operations\footnote{This is also what is referred to as moving back along the ``global feedback pathway'' \cite{bengio2015towards,ororbia2019biologically}; also see Figure \ref{fig:global_feedback_pathway}.} step-by-step, layer-by-layer, the required gradient $\nabla_\Theta \mathcal{L}(\mathbf{y},\mathbf{z}^L)$, or $\Delta \Theta$, is computed. %; see Figure \ref{fig:global_feedback_pathway}. 
Formally, calculating the required delta matrices is done as follows:
\begin{align}
    % output delta matrix
    \Delta \mathbf{W}^L \propto \frac{\partial \mathcal{L}(\mathbf{y}, \mathbf{z}^L)}{\partial \mathbf{W}^L} &= \frac{\partial \mathcal{L}(\mathbf{y}, \mathbf{z}^L)}{\partial \mathbf{h}^L} \frac{\partial \mathbf{h}^L}{\partial \mathbf{W}^L} = \frac{\partial \mathcal{L}(\mathbf{y}, \mathbf{z}^L)}{\partial \mathbf{h}^L} \cdot (\mathbf{z}^{L-1})^{\mathsf{T}} \\
    &= \bigg( \frac{\partial \mathcal{L}(\mathbf{y}, \mathbf{z}^L)}{\partial \mathbf{z}^L} \frac{\partial \mathbf{z}^L}{\partial \mathbf{h}^L} \bigg) \cdot (\mathbf{z}^{L-1})^{\mathsf{T}} = \Big( \delta^L \odot {\partial\phi}^L(\mathbf{h}^L) \Big) \cdot (\mathbf{z}^{L-1})^{\mathsf{T}} \label{eqn:backprop_out_delta}
    \\
    % intermediate delta matrices
    \Delta \mathbf{W}^\ell \propto \frac{\partial \mathcal{L}(\mathbf{y}, \mathbf{z}^L)}{\partial \mathbf{W}^\ell} &= \frac{\partial \mathcal{L}(\mathbf{y}, \mathbf{z}^L)}{\partial \mathbf{h}^\ell} \frac{\partial \mathbf{h}^\ell}{\partial \mathbf{W}^\ell} = \bigg( \frac{\partial \mathcal{L}}{\partial \mathbf{h}^L} \frac{\partial \mathbf{h}^L}{\partial \mathbf{z}^{L-1}} \frac{\partial \mathbf{z}^{L-1}}{\partial \mathbf{h}^{L-1}} \cdots \frac{\partial \mathbf{h}^{\ell+1}}{\partial \mathbf{z}^{\ell}} \frac{\partial \mathbf{z}^{\ell}}{\partial \mathbf{h}^{\ell}} \bigg) \frac{\partial \mathbf{h}^{\ell}}{\partial \mathbf{W}^{\ell}} \\
    &= \bigg( \Big( (\mathbf{W}^{\ell+1})^{\mathsf{T}} \cdot \frac{\partial \mathcal{L}}{\partial \mathbf{h}^{\ell+2}} \Big) \odot \phi^\prime(\mathbf{h}^\ell) \bigg) \cdot (\mathbf{z}^{\ell-1})^{\mathsf{T}} = \Big(\delta^\ell \odot {\partial\phi}^\ell(\mathbf{z}^L)\Big) \cdot (\mathbf{z}^{\ell-1})^{\mathsf{T}}\,, \label{eqn:backprop_hidden_delta}
\end{align}
where, above, we show first how to compute the adjustment for the output matrix $\mathbf{W}^L$ and then how to compute the model's internal intermediate synaptic parameter matrices for any layer $\ell \neq L$, i.e., $\mathbf{W}^{\ell}$. 
Note that ${\partial\phi}^\ell(\mathbf{h}^\ell)$ is the first derivative of activation $\phi^\ell()$ with respect to its pre-transformed vector $\mathbf{h}^\ell$ argument. In the above equations, we group sets of differentiation operations (the gradient of the cost with respect to a layer $\ell$'s pre-transformed activities $\mathbf{h}^\ell$) into the delta construct $\delta^\ell$ to further represent what is sometimes referred to as a ``teaching signal'' from the backprop perspective. 
%Furthermore, we set $\mathbf{\delta}^L = \frac{\partial \mathcal{L}(\mathbf{y}, \mathbf{z}^L)}{\partial \mathbf{z}^L} \frac{\partial \mathbf{z}^L}{\partial \mathbf{h}^L}$.
It is common to speed up the simulation of the above backprop credit assignment equations by using more than one data point at a time, i.e., a mini-batch, for the requisite calculations. That is, one could either use one single sensory input $\mathbf{x}$ for the equations above, i.e., $\mathbf{z}^0 = \mathbf{x}$ as in online learning, or one could substitute $\mathbf{x}$ with matrix $\mathbf{x} \in \mathbb{R}^{J_0 \times B}$ (and possibly $\mathbf{y} \in \mathbb{R}^{C \times B}$) where $B > 1$ is the batch size, i.e., this is sometimes referred to as batch-based learning. Figure \ref{fig:global_feedback_pathway} illustrates the resulting credit assignment induced by backprop on an MLP. 

Once the gradient matrices have been calculated using Equations \ref{eqn:backprop_out_delta} and \ref{eqn:backprop_hidden_delta}, resulting in $\Delta \Theta = \{\Delta \mathbf{W}^\ell\}^L_{\ell=1}$, we may then appeal to an optimization rule that changes the actual values of the MLP's synaptic values (or efficacies) in a subsequent step, e.g., through SGD with step size $\eta$ which entails $\mathbf{W}^\ell \leftarrow \mathbf{W}^\ell - \eta \Delta \mathbf{W}^\ell$. % or through a more sophisticated procedure, such as Adam \cite{kingma2014adam} or RMSprop \cite{tieleman2012lecture}.
We remark that, although we have shown weight updates computed via the chain rule for an MLP network, the scheme of backprop-based credit assignment can be applied to any type graph so long as it is restricted to be acyclic and its constituent mathematical operators, e.g., linear transforms, elementwise activations, are differentiable. This includes long operation chains, including the kind that compose unfolded recurrent neural networks or deep autoencoders; in these cases, however, the task of credit assignment becomes much more difficult as the processing depth, or length, of the chain increases. Notably, many variations of backprop have been proposed over the years, including \cite{fahlman1988faster,riedmiller1992rprop,hertz1997nonlinear}.

\subsection{The Problems with Backpropagation of Errors}
\label{sec:backprop_problems}

As presented before, credit assignment entails identifying the degree to which neuronal processing units, within a system, have an impact on a particular objective/cost function and, after doing so, adjusting their synaptic values (efficacies) in order to improve performance in the future. In terms of error, this would mean that credit assignment is engaged with assigning (partial) error values to every neural unit in service of minimizing a task-specific objective while, in terms of reward, this would mean allocating (partial) reward values to each unit in service of maximizing a task-central reward function. The updates made to synaptic parameters are made in terms of these computed per-unit error/reward allocations; in the abstract, a similar process, at least with respect to error/reward-centric optimization, has been theorized/observed to occur in the brain \cite{rao1999predictive, friston2003learning, friston2005theory,summerfield2006predictive,summerfield2008neural}. However, in the context of deep learning, the manner in which backprop carries out this distributed allocation of credit is largely considered to be implausible, with little to no neurophysiological evidence to support this form of learning. Over the decades since its earliest critiques \cite{grossberg_resonance_1987,crick1989recent}, it has become more apparent that the backprop-based adaptation is unlikely to occur in systems of real neuronal cells. In what follows, we explain several of backprop's core problems and long-standing issues, many of which are a mixture of both practical engineering issues and neurobiological criticisms.

\noindent
\textbf{The Global Feedback Pathway Problem.} Deep neural models trained with backprop have long been known to struggle with what has been labeled as \emph{vanishing} and \emph{exploding gradients} \cite{bengio1993credit,pascanu2013difficulty} (we will also refer to these collectively as ``unstable gradients''), which lead to instability in the training/fitting process of a DNN. Specifically, the issue of unstable gradients refers to the fact that the (Frobenius) norm of the gradients produced by backprop to update DNN parameters grows (exponentially explodes) or shrinks (exponentially vanishes) throughout the course of training; mathematically, a product of (Jacobian) matrices can grow towards infinity or shrink to zero (along a particular vector direction) much in the same way that a similar length series of numbers would \cite{pascanu2013difficulty}. The issue of unstable gradient values stems from the fact that a backprop-based scheme is attempting to conduct allocate per-unit credit across a deep hierarchy of computational elements, % multiplying a long sequence of numbers either smaller or greater than one...%(coming down from the output layer to lower hidden layers) is purely linear, whereas biological neurons interleave linear and non-linear operations
percolating information recursively by coming down/back from the entire system's output layer to the lower hidden layers. The resulting long chain or pathway of recursive operations has been referred to as the ``global feedback pathway'' \cite{ororbia2017learning,ororbia2019biologically} (see Figure \ref{fig:global_feedback_pathway}) and have been argued to be an important aspect of error-driven learning inherent to backprop that needs to be addressed in order to emulate the more robust, stable learning that characterizes natural neuronal networks. %it is a driving mechanism of error-driven learning that needs to be dispensed with if more parallel, local learning is desired.
Note that the issues related to credit assignment over deep hierarchies of computational units are exacerbated when training temporal neural models on sequence data, such as recurrent neural networks (RNNs) \cite{bengio1994learning}, which require unfolding the neural model backwards across time (an act in of itself that has been criticized to be quite biologically implausible \cite{ororbia2018continual}). The resulting instability created by the global feedback pathway created by backprop (through time) makes it extremely difficult for an RNN to learn correlations between temporally distant events.
%% INTEGRATE THIS TEXT INTO ABOVE (global fbk path is also reason for many of the problems below)
%The global feedback pathway \cite{bengio2015towards,ororbia2019biologically} presents the key problematic issue underlying backprop -- there is no evidence that the neural circuits in the brain transmit backward error derivatives back along the same pathways that information was transmitted along. This crucially prohibits parallel computing across layers in the system.

\begin{figure}[!t]
\centering     %%% not \center
\begin{subfigure}{0.325\textwidth}
    \centering
    %\vspace{-5mm}
     \includegraphics[width=0.63\linewidth]{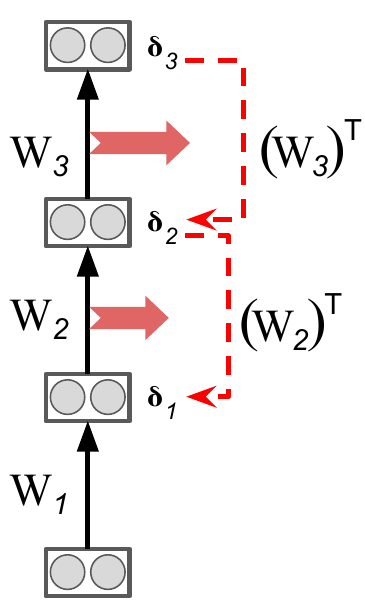} 
    \caption{The weight transport problem.}
    \label{fig:weight_transport}
\end{subfigure}
\begin{subfigure}{0.325\textwidth}
    \centering
    %\vspace{-5mm}
     \includegraphics[width=0.4\linewidth]{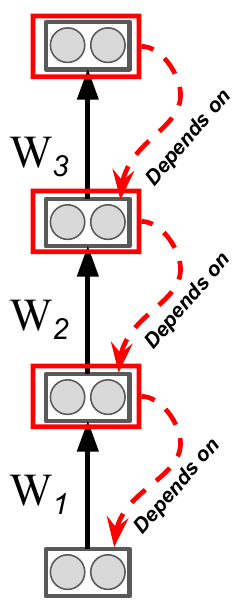} 
    \caption{The forward-locking problem.}
    \label{fig:forward_lock}
\end{subfigure}
\begin{subfigure}{0.325\textwidth}
    \centering
    %\vspace{-5mm}
     \includegraphics[width=0.545\linewidth]{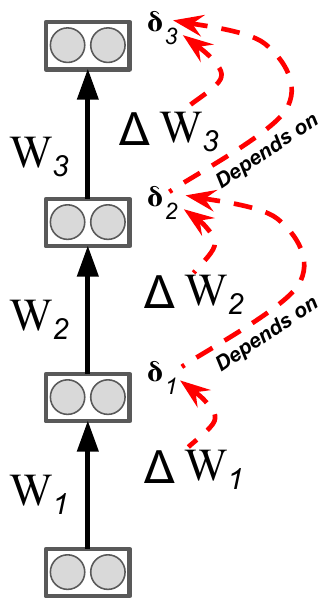} 
    \caption{The update-locking problem.}
    \label{fig:update_lock}
\end{subfigure}
%\vspace{-0.3cm}
\caption{\small{\textbf{Problems with backpropagation of errors.} 
Depicted are three of the core issues underlying credit assignment that is based on backprop (from Left to Right): the weight transport, the forward-locking, and the update-locking problem.
}}
\label{fig:backprop_issues}
\vspace{-0.5cm}
\end{figure}

\noindent 
\textbf{The Weight Transport Problem.} Weight transport, or the problem of symmetric synaptic connections, refers to the fact that, in backprop, the same synaptic parameter matrices that are used to conduct inference are reused in communicating information for the learning phase  (see Figure \ref{fig:weight_transport}). This means that in a backprop-adapted DNN, pre-synaptic neurons receive error information from post-synaptic ones through the very same synapses that transmitted (input) information forward \cite{grossberg1987competitive}. In neurobiology, this operation is not feasible in bio-chemical synapses given that neurotransmitters and receptors enforce a unidirectional flow of information. As a consequence, synaptic feedback loops in the brain are the result of leveraging two different sets of synapses \cite{lillicrap2016random,lillicrap2020backpropagation,ororbia2017learning,whittington2019theories, bengio2015towards}. Furthermore, it is entirely possible that feedback loops are not event present in some cases \cite{song05random}, which is an aspect that certain brain-inspired algorithms attempt to address (e.g., forward-only learning; see end of Section \ref{sec:synergistic_local_explicit_algos}).

\noindent
\textbf{The Inference-Learning Dependency Problem.} This refers to the central dependency of the pathways between those of learning and those of inference . This is a consequence of the fact that backprop-based models require the presence of two heterogeneuos computational/calculation phases -- a forward and a backward pass -- each characterized by their own constituent operations \cite{rumelhart1986learning, lillicrap2020backpropagation}. This means that the adjustments made to synaptic parameters, which driven by the backward transmission of error gradients (or ``teaching signals'') of a global cost function, inherently depend on the statistics produced by the forward propagation of information (the inference phase), a conditional dependency that is not observed in real neuronal structures. In terms of the brain, this relationship between synaptic adjustments and neural activity values produced by forward propagation-based inference in backprop would implausibility mean that biological neurons would require storage capabilities for memorizing the forward signals in direct support of learning. Furthermore, the computation underwriting inference and learning in backprop needs to be precisely clocked to alternate between forward and backward propagation phases \cite{bengio2015towards} whereas, in the brain, there is no need for such external control \cite{whittington2017approximation}, i.e., neurons perform computation autonomously with little external routing information applied at externally-decided times. % desire minimal external control - the computation would have to be precisely clocked to alternate between feedforward and back-propagation phases (since the latter needs the former’s results) 
%The neurons perform the computation autonomously with as little external control routing the information that runs through them at any given time.%in different ways at different times as possible.

\noindent 
\textbf{The Problem of Locality and Locking.} Another critical issue inherent to backprop-based learning is that the rules/mechanisms which dictate synaptic updates are non-local (this is also a consequence of the inference-learning dependency problem above) and are dependent upon the minimization of a globally defined cost function that itself depends on the value of neural activities across the network, including those that are near the bottom of the hierarchy. This strongly contrasts with how plasticity is believed to occur in biological neurons and the synapses that connect them, i.e., neuronal adaptation and plasticity is believed to rely on information that is local in both space and time \cite{ramon1894croonian,hebb1949organization}. The non-locality inherent to backprop-based credit assignment further gives rise to three related sub-problems: the forward-locking, backward-locking, and update-locking problems \cite{jaderberg2016decoupled,czarnecki2017understanding}. The forward-locking problem is the result of the very nature of (feed-forward) inference in modern-day DNNs; activity values of one layer of processing elements depend on all of the existence/activities of the layers below/that come before them (see Figure \ref{fig:forward_lock}); in other words, no single layer can process incoming information before the previous layers in a directed graph have been executed. Update-locking itself stems from the problem of forward-locking -- the adjustments to one portion of the computational graph depend on computations that come after them (see Figure \ref{fig:update_lock}); no layer's synapses can be updated before all of the dependent modules have executed in forward inference mode and the error gradient information has been back-transmitted in the layers above/ahead. The backward-locking problem is effectively the forward-locking problem but applied to the chain of operations that characterize backprop-based learning in a DNN; no neuronal layer can be updated before all dependent neuronal layers have been executed in both forward inference and backwards propagation mode. Resolving these three locking problems inherent to backprop-centric  learning would open the door to a plethora of biological and practical computational possibilities; this includes the ability to make adjustments to portions of a neural system in parallel and asynchronously.
% Local computation. A neuron performs computation only on the basis of the inputs it receives from other neurons weighted by the strengths of its synaptic connections. 
%Local plasticity. The amount of synaptic weight modification is dependent on only the activity of the two neurons the synapse connects (and possibly a neuromodulator).
%% There may be other "local hacks" to choose parameter updates that incorporate some sort of credit assignment, instead of allowing all parts of a model race to incorporate any available information including noise. In handwaving terms, we could optimize parts of a model to pass on information that ought to eventually be useful instead of greedily taking a step in the direction that's most useful right now.

\noindent
\textbf{The Problem of Constraint and Sensitivity.} % the problem of arbitrary topologies
Through the use of automatic differentiation \cite{margossian2019review}, it is possible to train different types of computational neural structures (beyond linear chains) consisting of multiple types of operations. Many modern-day DNNs consist of multiple, various kinds of layers, including those that leverage convolution, as in convolutional networks \cite{fukushima1982neocognitron,ciregan2012multi,Krizhevsky2012}, or multiple heads of self-attention, as in neural transformers \cite{Vaswani17}. However, despite its flexibility, a backprop-centric form of learning does impose several constraints and functional requirements on the architectures that can be built: 
\textbf{1)} model must be fully differentiable, and 
\textbf{2)} backprop is limited to training networks that take the form of directed acyclic graphs. With respect to the first constraint, a model being fully differentiable means that all of its constituent operations must also be differentiable (including its elementwise activation functions); otherwise, it is not possible to carry out backprop transmission due to the fact that the matrix chain rule of calculus entails a chained product of local first derivatives/Jacobians. This makes it difficult to utilize discrete-valued functions and stochastic elements (such as Bernoulli sampling), making it challenging to design systems that communicate using discrete/spike values (as in spiking neuronal systems), thus further hindering our ability to construct more energy efficient neural systems \cite{ororbia2019spiking,yi2023activity}. Biologically, this means that the feedback pathways that characterize the underlying brain structures %(with their own synapses and maybe their own neurons)
would require precise knowledge of the derivatives of the nonlinear dynamics (at particular operating points) of the neurons that transmit information in the corresponding feedforward computations of the DNN's inference pathway. %%if the feedback paths known to exist in the brain (with their own synapses and maybe their own neurons) were used to propagate credit assignment by backprop, they would need precise knowledge of the derivatives of the non-linearities at the operating point used in the corresponding feedforward computation on the feedforward path
In terms of the second constraint, if a cycle is present inside a DNN's neural structure, an infinite loop is created in the forward pass which makes learning impossible. To address this limitation, such as for temporal/sequential data, researchers have developed a time-dependent variation of backprop which stores vectors on neural activity values across time \cite{hochreiterLSTM1997} -- this is known as backpropagation-through-time (BPTT). However, this further increases the implausibility of this type of error-driven learning since the neural activity values that were earlier stored for error computations now must be further stored over time \cite{hinton2022forward}. This limits biological plausibility and prevents emulating much of the message-passing structure that characterizes processing in brain given that the structure of biological networks is extremely complex, full of cycles, and heterarchically organized with small-world connections \cite{avena18,salvatori2023brain}. These topologies are likely highly optimized by evolution, where different topological attributes promote different types of communication mechanisms \cite{suarez2020linking,farahani2019application, boguna2021network}. In addition, the requirement of an acyclic form makes it difficult to incorporate mechanisms such as lateral competition and cross-layer interactions (which further hinder the use of mechanisms useful for fighting off catastrophic forgetting). 

%%high sensitivity to initialization
Finally, another aspect of this problem is that DNN's are highly sensitive to their initial conditions as well as the choice of normalization. In particular, it is well-known in DNN optimization that a key ingredient in guaranteeing convergence (and subsequent generalization) is the initialization scheme used to randomly instantiate the synaptic weights. However, DNNs trained with backprop are particularly sensitive to the random initial values chosen for its synapses  \cite{kolen1990back}, which ultimately hinders its overall final performance. A great deal of research has gone into crafting effective initialization setups \cite{glorot2010understanding,saxe2013exact,he2015delving,hu2020provable} (including data-level/dependent ones \cite{lecun2002efficient,saxe2013exact}); nevertheless, this issue still persists to this day. Furthermore, widely-used initialization schemes in modern-day DNNs, e.g., Glorot initialization \cite{glorot2010understanding}, have been shown to have their limitations, e.g., \cite{kumar2017weight} and \cite{jaiswal2022old} demonstrated that blindly using Glorot initialization can result in (long-term) suboptimal generalization performance. Furthermore, in practice, modern DNNs depend heavily depend not only on initialization but also on the normalization schemes applied at the data and activation levels \cite{lecun2002efficient,ba2016layer,ioffe2015batch}; the choice of a good normalization scheme can aid in ensuring faster convergence and possibly better generalization. Nevertheless, batch or activity-level normalization introduces further issues -- some schemes only work well with small batch sizes \cite{ba2016layer}, others only work when operating across certain input patterns or particular activation functions, and some do not work well when used jointly with other normalization strategies and/or noise-injection processes. In addition, the dependence on normalization results in leads to a training process that struggles when there are dependencies between samples within a mini-batch (since necessary statistics are calculated at the batch-level) and further can result in model operational instability when the distribution during test-time inference is different from or strongly drifts away from the training distribution.

\noindent 
\textbf{The Problem of Short-Term Plasticity.} This problem pertains to the nature of how inference is typically carried out in DNNs. Concretely, DNNs do not model nor offer an account for short-term plasticity; in effect, neural activities do not ``exist'' until data is clamped at the input layer(s) of the system and then forward propagation calculations are performed (this also is partly the cause of the forward-locking problem described earlier). Furthermore, in backprop, the error information that is propagated backwards only effect changes in synaptic values and do not result in any modification of the neural activities themselves \cite{betti2018backpropagation,lillicrap2020backpropagation}. In contrast, in the brain, inherent feedback connectivity is observed to operate quite differently, e.g., feedback synapses in the cortex influence feedforward neural dynamics in a top-down modulating fashion \cite{oreilly2000computational}. Although the problem of short-term plasticity centers around the way that inference is carried out in DNNs, since inference is a critical aspect of functionality that (directly or indirectly) affects learning, it is an important aspect of neural computation that many credit assignment approaches try to address, as we will see throughout this survey. This problem, which we have made explicit, also relates to another important aspect that brain-inspired machine intelligence has begun to consider -- biologically-plausible architectural design, which involves considering that connectivity patterns within a neural model should be consistent/in accordance with basic constraints of brain connectivity (in the neocortex) \cite{whittington2017approximation}.
%\item functional requirement of sufficient linearity -- this makes model vulnerable to adversarial samples;

\subsection{On the Target Signals that Drive Synaptic Plasticity} % and Other Issues}
\label{sec:taxonomy}

% Tie-in w/ central question; addressal of criticisms and desiderata 
As foregrounded at the start of this article, any computational process designed for conducting credit assignment within a neural system context must answer the question: where do the (target) signals for inducing learning/adaptation, for each processing unit within a neuronal network, come from?\footnote{Note that target signals could come in many forms, ranging from separate pools of neuronal activities produce mismatch values all the way to control signals that trigger or modulate changes in synaptic strengths.} There are many possible answers to this question, and in this survey, we will focus on six of the more prominent answers to it, each of which motivates a different family of biologically-inspired algorithms and frameworks.

As shown in Figure \ref{fig:system_and_taxonomy}, we start by breaking down the signal type into two general categories, \emph{implicit} and \emph{explicit}. In the case of learning and adaptation that makes use of implicit signals, which is also one of the six computational families, the information used to adjust synaptic parameters is completely local in both time and space; this is what centrally characterizes pure Hebbian-type rules \cite{hebb1949organization} or what is also known as ``correlation learning'' \cite{oreilly1998sixprinciples}. 
Specifically, synaptic change depends only on information that is immediately usable at pre- and post-synaptic sites/locations, i.e., the incoming and outgoing neurons that a synaptic cable connects.\footnote{In neurobiology, a synapse is a specialized junction that carries/mediates the information between neurons; it is many of these ``cables'' that facilitate communication between an individual neuron and its small subset of pre-synaptic transmitters and post-synaptic receivers.} Consequentially, there is no explicit signal or external information, such as that based on error (such as a signal created by comparing a prediction to a reference value), that is transmitted to a particular neuron. 

Alternatives to credit assignment based on implicit signals are naturally those based explicit values; this encompasses many sets of procedures that rely on information beyond what is locally available to any single synapse. Any biologically-inspired process under this general partition typically creates signal values based on some sort of process (which itself could be local), such as those based on the message passing of error/mismatch measurements. Within the cluster of explicit signal algorithms, we partition credit assignment schemes based on whether or not they make use of a \emph{local} or \emph{global} signal to induce synaptic weight change. Under the global category (which is a leaf in the taxonomic tree and thus a distinct family), we have many possible frameworks, ranging from feedback alignment to neuromodulation. 
Within the local category, procedures can be further sorted based on ``how local'' they really are and, in this survey, we propose two sub-partitions -- %(each of these will then result in leaves, or actual algorithm families of the taxonomic tree) -- 
`non-synergistic' and `synergistic'. A \emph{non-synergistic local} scheme operates with information that is local in the sense that it is spatially and temporally ``near'' the neurons it will affect but, unlike an implicit signal scheme, this is information or target values beyond what a synapse and the two neurons it connects could possibly provide. 
%they make use of information that is spatially and temporally available (to the two neurons that a synaptic weight connects).
Mechanisms capable of creating this information generally involve additional neurons and synaptic parameters, typically forming a local predictor of some kind, e.g., a classifier that has immediate access to a label context, or specialized local feedback synapses that enable a pair of layers to form an encoder/decoder sub-system. 
%However, the locally available information is provided by a mechanism beyond the statistics of the neurons themselves -- this could mean that an error signal of some form is produced, involving comparing neuronal activity to some locally produced target value (possibly using specialized, local feedback synaptic parameters) or using global information that is immediately accessible to all neurons (such as a label vector).
On the other hand, a \emph{synergistic local} learning process is one where the signaling information or target values are produced using some indirect knowledge of the neural system state; this kind of information is usually acquired through a message passing scheme or iterative settling process, typically (though not always) involving additional neural circuitry to construct feedback loops. 
%, although the synaptic weight connecting any pair of neuronal units is also updated using locally located/computed information (again, such as comparing activities to corresponding goal values), the targets are generated with some indirect knowledge of the system state, acquired usually through a message passing scheme or iterative settling process that utilizes feedback loops. 
%activities of layers above and below, possibly through feedback loops or nearby error communication. 
Within this classification, there are three primary sub-paradigms (or taxonomic leaves): discrepancy-based  \cite{friston2010free,ororbia2017learning}, energy-based \cite{lecun2006tutorial}, and forward-only frameworks \cite{kohan2018error}. % <-- this connects to O'Reilly's interactive net stuff
% each family presupposes something about the brain (or does not) which could be accounted for actual structure or some chemical process...

\begin{comment}
While the key differentiating element between algorithmic families is how they decide what the targets are and where they come from, along the way, we will consider and note beneficial or neurobiologically-faithful mechanisms/properties offered by various algorithms within each family in turn. Concretely, these properties (sometimes side-effects of the nature of a particular credit assignment scheme) will be examined in the context of the problems/criticisms related to backprop-based learning (as in Section \ref{sec:backprop_problems}), e.g., whether or not a scheme requires differentiable neural activities/dynamics, whether or not symmetric feedback pathways are required, or whether  training-time and testing/inference-time computations are different or the same. 
\end{comment}

\section{Families of Neuro-mimetic Credit Assignment}
\label{sec:algo_families}

%%Family # 1: 
\subsection{Implicit Signal Algorithms}
\label{sec:implicit_algos}
The first family of procedures that we review offers a simple answer to this  survey's central question: there is no target or external signal. Rather, the target is implicit and not produced by external processes, unlike the other algorithms that we will investigate. These schemes utilize information that is readily available to each individual synaptic connection of the architecture. This could mean, in the case of Hebbian learning, that only the activities of the pre-synaptic and post-synaptic neuronal elements are needed, and adaptation is effectively a form of correlation-based learning. 

The schemes, sometimes referred to as `update rules' in the literature, that fall under this family only make use of the information that is produced by the inference process of the neural model, e.g., a network's feedforward pass. This is particularly attractive not only from a practical perspective, given that computation only involves neural activity vectors immediately available from the inference process of the system (with no further signals provided by mechanisms such as feedback), but also from a neurophysiological standpoint, since synaptic changes can be calculated with exclusively local information. Furthermore, activation derivatives are typically not required in any of the methods within this family and, furthermore, training and test-time computations are identical. 
\begin{comment}
Despite these big advantages, the primary drawback of many of these schemes is that an implicit positive feedback loop is created that results in ever-increasing weight magnitudes (particularly for Hebbian rules). Resolving this issue often requires significant heuristics or modifications to normalize the system and prevent unbounded weight growth.
\end{comment}

\noindent
\textbf{Hebbian Learning.} Hebbian-based adjustment is a classical, biologically-plausible synaptic modification rule. It is based on the idea that synaptic plasticity is the result of a pre-synaptic neuronal cell's persistent and repeated stimulation of a post-synaptic cell \cite{hebb1949organization,kuriscak2015biological}\footnote{Specifically, it was observed in \cite{hebb1949organization} that: ``When an axon of cell A is near enough to excite a cell B and repeatedly or persistently takes part in firing it...A's efficiency, as one of the cells firing B, is increased''.}, popularly summarized by the phrase: ``neurons wire together if they fire together'' \cite{lowel1992selection}. Each and every occurrence of input activity patterns (from pre-synaptic specializations) strengthen the ability of the related synaptic parameters to recall or reproduce the pattern later on. In effect, Hebbian learning allows neural structures to encode memories in their synaptic connectivity. What is quite attractive about Hebbian-like rules is that they operate completely locally -- this resolves one of the central problems inherent to backprop, the problem of locality and locking. In essence, synaptic adjustments only require information readily available and in close proximity to the (location of the) synaptic weight of interest. In other words, Hebbian rules are generally cell-by-cell rules, where information regarding some aspect of pre-synaptic activation and post-synaptic activation -- such as magnitudes of neuronal activation patterns -- as well as possibly a dependency on current synaptic efficacies, are used to calculate and update the matrix of weights connecting layer $\ell-1$ to $\ell$. Formally, this type of adjustment can be expressed in scalar (single synapse) form, i.e., $\tau \frac{\partial W_{ij}}{\partial t} = W_{ij} (z^\ell_i z^{\ell-1}_j)$,  
or in matrix-vector form, i.e., $\tau \frac{\partial \mathbf{W}^\ell}{\partial t} = \Delta \mathbf{W}^\ell = \mathbf{W}^\ell \odot \Big( \mathbf{z}^\ell \cdot (\mathbf{z}^{\ell-1})^{\mathsf{T}} \Big)$ ($\tau$ is a time constant that would be refactored into what is known as the 
%%%%%%%%%%%%%%%%%%%%%%%%%%%%%%%%%%%%%%%%%%%%%%%%%%%%%
\begin{wrapfigure}{r}{0.525\textwidth}
\vspace{-0.455cm}
  \begin{center}
    \includegraphics[width=0.41\textwidth]{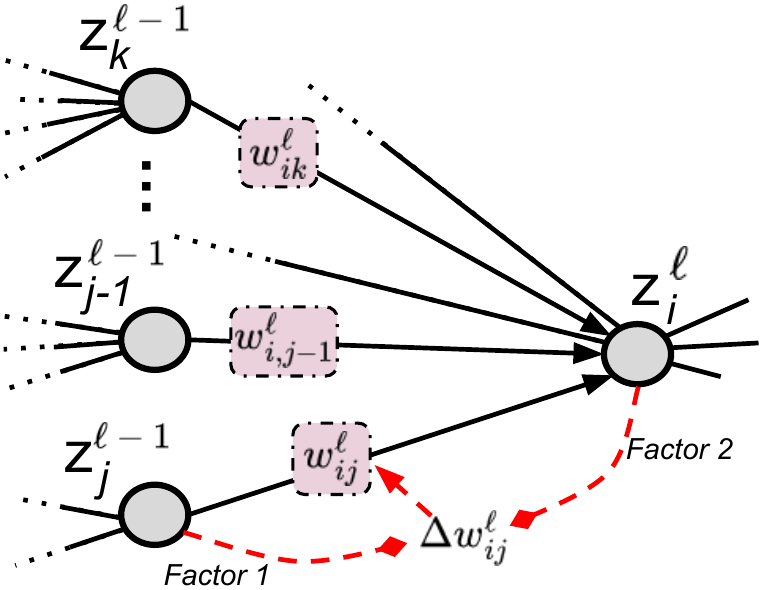}
  \end{center}
  \vspace{-0.255cm}
  \caption{\small{Two-factor Hebbian plasticity: here, only pre- (neuron $j$) and post-synaptic (neuron $i$) measurements of a specific synaptic connection ($w_{ij}$) are used to produce its adjustment $\Delta w_{ij}$.
  }}
  \label{fig:hebb_plasticity}
  \vspace{-0.4cm}
\end{wrapfigure}
%%%%%%%%%%%%%%%%%%%%%%%%%%%%%%%%%%%%%%%%%%%%%%%%%%%%%
``learning rate''). This update equation has also be referred to as a two-factor Hebbian rule; one factor is the pre-synaptic activity while the other is the post-synaptic activity (the use of the synapse with the rule is generally not counted as a separate factor).

Although this scheme is simple and efficient, requiring only the activity values produces by the neurons in the pairing of layers $\ell-1$ and $\ell$, as well as the current value of the synapses connecting them, one critical drawback of na\"ively-interpreted Hebbian update rules is possibility of exploding weight magnitudes; an implicit positive feedback loop is created by repeated application of the rule, resulting in ever-increasing weight magnitudes. This can be taken care of with the introduction of plasticity constraints \cite{miller1994role}, e.g., some form of weight decay \cite{yuille1989quadrature}, normalization \cite{von1973self,rubner1989self}, or both \cite{oja1982simplified,hyvarinen1998independent} (as in Oja's rule\footnote{Oja's rule \cite{oja1982simplified} is a generalized Hebbian plasticity rule which introduces a mechanism for ensuring the norms of each vector synaptic weights are approximately constant after adjustment.}). 
For example, one simple way to introduce a depressing force on the synaptic magnitudes is via: $\Delta \mathbf{W}^\ell = \mathbf{W}^\ell \odot \Big( \mathbf{z}^\ell \cdot (\mathbf{z}^{\ell-1})^{\mathsf{T}} \Big) - \gamma \mathbf{W}^\ell$, where $\gamma$ is a non-negative decay factor. 
Beyond decay, synaptic strengths can further be bounded to a magnitude range, yielding the full Hebbian plasticity update below:
\begin{align}
    \tau \frac{\partial \mathbf{W}^\ell}{\partial t} = \Big( w_{max} - \mathbf{W}^\ell \Big) \odot \Big( \mathbf{z}^\ell \cdot (\mathbf{z}^{\ell-1})^{\mathsf{T}} \Big) - \gamma \mathbf{W}^\ell
\end{align}
where $w_{max}$ is a scalar bound on the maximal value any synaptic strength in $\mathbf{W}^\ell$ can take on and $\mathbf{W}^\ell$ is assumed to only take on non-negative values. 
The above differential equation results in softly-constrained, multi-term Hebbian plasticity update rule. % (cast in terms of a differential equation).  % add citations to multi-factor rules

Another way to correct the explosive nature of Hebbian-like rules is to incorporate a local mechanism for weight depression, thus leading to the incorporation of anti-Hebbian counter-pressures \cite{foldiak1990forming} or the use of gated Hebbian rules \cite{gerstner2002mathematical}. For example, one could use the post-synaptic gated update rule, or $\Delta \mathbf{W}^\ell = (\mathbf{z}^\ell - \mathbf{g}^\ell) \cdot \big(\mathbf{z}^{\ell-1}\big)^{\mathsf{T}}$, where $\mathbf{g}^\ell$ is a set of thresholds, one-per-neuron in layer $\ell$, that allow the post-synaptic activity to change the sign/direction of the weight change but still emphasizes the importance of pre-synaptic activity for any change. Synaptic gating serves as a building block for more sophisticated Hebbian-like adjustment rules, such as the Bienenstock-Cooper-Monroe (BCM) update rule \cite{bienenstock1982theory,cooper2004theory}; $\Delta \mathbf{W}^\ell = \alpha \Phi(\mathbf{z}^\ell - \mathbf{g}^\ell) \cdot \big(\mathbf{z}^{\ell-1}\big)^{\mathsf{T}}$ where $\Phi$ is a nonlinearity applied to the post-synaptic gating. Alternatively, there is the pre-synaptically gated rule: $\Delta \mathbf{W}^\ell = \mathbf{z}^\ell \cdot \big(\mathbf{z}^{\ell-1} - \mathbf{g}^{\ell-1}\big)^{\mathsf{T}}$ \cite{gerstner2002mathematical}. 
Other formulated schemes introduce stabilizing mechanisms, e.g., calculated values based on statistics of tracked neural dynamics, into the original Hebbian framework. An example of such a rule is the Hebbian covariance rule \cite{sejnowski1989hebb}:
\begin{align}
    \Delta \mathbf{W}^\ell = (\mathbf{z}^\ell - \langle \mathbf{z}^\ell \rangle ) \cdot (\mathbf{z}^{\ell-1} - \langle \mathbf{z}^{\ell-1} \rangle )^{\mathsf{T}} \label{eqn:hebb_covar}
\end{align}
where $\langle\mathbf{z}^\ell\rangle$ indicates a short-term running average of each neural firing rate within $\mathbf{z}^\ell$. However, these types of rules generally mean incorporating  statistics that partially violate the property of locality -- Equation \ref{eqn:hebb_covar} breaches locality in time -- in order to adjust the synaptic weight values. Further note that generalizations of Hebbian learning have been explored in the context of complex vision architectures, such as those that employ convolution \cite{lagani2022comparing}. 

%% there are two further points to make:
%% 1) there are supervised variants of Hebbian rules (that force externally introduced targets like labels)
%% 2) Hebb rules/themes pop up again in other algos like CHL and PC, however, these fall under a different algo family (seen later on)
Although Hebbian learning has historically been presented as an unsupervised process for synaptic plasticity, variants introduce means of conducting supervised learning in the presence of a desired goal $\mathbf{y}$, e.g., a label. Instances of supervised Hebbian learning, some of which are reminiscent of the principle of ``teacher forcing'' \cite{toomarian1992learning}, include the perceptron learning rule \cite{rosenblatt1958perceptron}, the delta rule \cite{hanson1990stochastic,andersen1995modified,voelker2015solution}, and the Widrow-Hoff rule \cite{widrow1960adaptive,widrow1983adaptive}. The last one takes the form $\Delta \mathbf{W}^\ell = (\mathbf{z}^\ell - \mathbf{y}) \cdot (\mathbf{z}^{\ell-1})^{\mathsf{T}}$; notice that this rule happens to correspond to the first derivative of a mean squared error cost -- it is sometimes referred to as the least mean squared errors rule. In addition, notions of Hebbian plasticity have appear in other credit assignment paradigms; frameworks such as contrastive Hebbian learning \cite{baldi1991contrastive} or predictive coding \cite{rao1999predictive,salvatori2023brain}, of which we will review later, also entail final synaptic adjustments that are made using information such as pre- and post-activity signals (although such schemes require external mechanisms such as message passing). %However, although the final connection adjustments look ``Hebbian'', they are the result of and depend on message passing/settling dynamics to produce the requisite target/driving signals. 
It is interesting to note that the delta rule \cite{andersen1995modified}, which extends the Widrow-Hoff rule by incorporating the derivative of the activation function, can be viewed as a building block that gave rise to the full generalization that is known today as backprop \cite{rumelhart1986learning}. Note that modern research efforts have crafted schemes based on supervised Hebbian plasticity to train deeper, multi-layer neural models \cite{gupta2021hebbnet,alemanno2023supervised}.

Although raw Hebbian plasticity has its drawbacks and, as of today, is rarely used in isolation to directly to train complex neural systems (except in some more recent cases \cite{journe2022hebbian}), it still plays an invaluable role in emulating other aspects of neurobiological organization and functionality.  For instance, a powerful idea behind Hebbian-based and neuro-correlational rules is that they are engage a form of ``model-learning'' \cite{oreilly2000computational}. Rather, such rules facilitate the general extraction of co-occurrence statistical structure from a neural system's environment, making it particularly attractive to utilize for unsupervised dimensionality reduction. For instance, prior work has established a strong connection between Hebbian rule variants and principle components analysis (PCA) \cite{oja1982simplified}. Beyond this, Hebbian learning has crucially been shown to be a powerful means of constructing models of memory, such as those based on Minerva-2 \cite{hintzman1984minerva} and sparse distributed memory \cite{kanerva1988sparse}.

%% make connection to STDP (which is another advantage of multi-factor Hebbian rules)
Another important aspect of Hebbian plasticity is its generalization to the spiking temporal domain, i.e., spike-timing-dependent plasticity (STDP) \cite{levy1983temporal,abbott2000synaptic,bi2001synaptic}, which facilitates the adaptation of networks based on communication of discrete (action potential) values, such as those composed of spiking neuronal cells such as spiking neural networks (SNNs) \cite{maass1997networks}. Hebbian adjustment through STDP specifically entails using the relative timing of the action potentials (or spike emissions) of pre- and post-synaptically located neurons; a sliding temporal window is used to determine if a pre-synaptic spike arrives before a post-synaptic one which results in a positive increase in a synapse's efficacy (long-term potentiation) while, if this timing is reversed, i.e., post- occurs before pre-synaptic spike, then a decrease is applied (long-term depression). STDP-centric Hebbian plasticity notably facilitates the capture of temporal correlations inherent to sensory data streams\footnote{The integral over an STDP window, assuming slowly changing neuronal firing rates, can be shown to recover a Hebbian correlational update similar to the ones described earlier in this section.} and serves to this day as a useful biophysical mathematical model of neuronal organization and synaptic adaptation \cite{zhang1998critical,muller2010plasticity,koch2013hebbian,tazerart2020spike,andrade2023timing}. %neuronal organization during development and a driving force for receptive field formation
However, STDP and general Hebbian-based plasticity largely work as unsupervised forms of adaptation and are not directly useful for modeling behavioral learning (though there are supervised variants, as mentioned before). In general, to make Hebbian adjustments work for action-oriented functionality and behavioral conditioning experimental settings, it is typically extended to include a third additional factor; a three-factor Hebbian adjustment \cite{pawlak2010timing} combines the original two factors, i.e., pre- and post-synaptic activity values, with a third one typically labeled as the modulator, e.g., a dompanie neuromodulatory signal (such as an encoding of a reward function that conveys the degree of success of actions taken).\footnote{Three-factor Hebbian schemes furthermore integrate what is called an eligibility trace in order to handle the temporal correlations inherent to sequences of actions taken over time in the presence of delayed reward signals.} However, adding an additional factor to Hebbian plasticity technically results in a credit assignment mechanism that falls under a different category in our taxonomy, i.e., global explicit target algorithms, of which we will review/cover later (see Section \ref{sec:global_explicit_algos}).

\noindent
\textbf{Competitive Hebbian Learning.} One of the simplest rules entails adding intra-layer connections between neurons, particularly inhibitory connections. The idea is that neurons in a given layer will compete with each other for the chance to represent particular input patterns; this is the essence of what is historically known as `competitive learning' \cite{rumelhart1985feature,desieno1988adding,white1991competitive,martinetz1993competitive,dayan1995competition,wang1997competitive,gliozzi2018self}. In winner-take-all style competition \cite{kohonen1988neural}, the unit with the highest level of activation will be declared the winner, warranting an update to its incoming synaptic weights while the rest of the losing neurons receive no adjustment to their incoming weights. Note that weights and inputs must be normalized (unless they contain bipolar values, i.e., values in the set $\{-1,0,1\}$). For a winner-take-all (WTA) block of neurons -- which involves an entire layer or be limited to a specific sub-group of neurons, i.e., local WTA units \cite{srivastava2013compete}) -- any processing element $z^{\ell}_j$ in layer $\ell$ is updated according to a Hebbian update that operates in tandem with a hard interaction function as follows:
\begin{align}
  \Delta W^\ell_{ij} = 
    \begin{cases} 
      z^{\ell-1}_{j} - W^\ell_{ij} & \mbox{if} \quad i = \argmax_{\{1,2,..,\mathcal{J}_\ell\}} \Big( \mathbf{z}^\ell \Big)
      \\
      0 & \mbox{otherwise,} 
    \end{cases}
\end{align}
where we note that a vector update is produced for the $i$th row of $\mathbf{W}^\ell$ as a result of the local update. Notably, it was the general use of WTA-driven synaptic change in a multi-layer neural model that served as a key part of the classical Neocognitron \cite{fukushima1982neocognitron,fukushima1988neocognitron}, the historical predecessor to the modern-day convolutional network. 
The above is a variation of the `instar update rule' \cite{grossberg1969embedding,kohonen1988neural}, which, simply leads to modifying the synapses connecting to the neuron that took on the largest post-activation value for a given input $\mathbf{z}^{\ell-1}$; the other synaptic connections are left unchanged. Note that this scheme may be extended to leverage the top $K$ highest activity values instead (a $K$-winners-take-all scheme). To extend the instar algorithm to supervised learning settings, the `outstar update rule' was proposed \cite{grossberg1969embedding}, where, instead, outgoing weights are updated so that neuronal output matches a desired target pattern, such a label vector $\mathbf{y}$. The instar and outstar forms of competitive plasticity ultimately gave rise to what is known as adaptive resonance theory \cite{grossberg1987competitive,grossberg2013adaptive} -- and its plethora of variants, such as fuzzy ART \cite{carpenter1991fuzzy}, ARTMAP \cite{carpenter1991artmap}, and ART-C 2A \cite{he2004modified} -- as well as the counter-propagation generalization for credit assignment in multi-layer models \cite{hechtNielsen1987counterprop}. 
Other ways to induce competitive neural dynamics include anti-Hebbian learning \cite{foldiak1990forming,ororbia2021continual} and explicit lateral/cross-layer synaptic connections (with influences on more modern deep learning architectures \cite{krotov2018unsupervised}) or by enforcing a sparse kurtotic prior distribution (or penalty), e.g., a Laplacian or Cauchy prior, within an iterative inference process \cite{olshausen1997sparse,rao1999predictive} (the latter of which is used in sparse coding, which is technically part of a different class of credit assignment schemes that will be covered in Section \ref{sec:discrep_reduction}).

%%%%%%%%%%%%%%%%%%%%%%%%%%%%%%%%%%%%%%%%%%%%
\begin{wrapfigure}{r}{0.525\textwidth}
\vspace{-0.455cm}
  \begin{center}
    \includegraphics[width=0.5\textwidth]{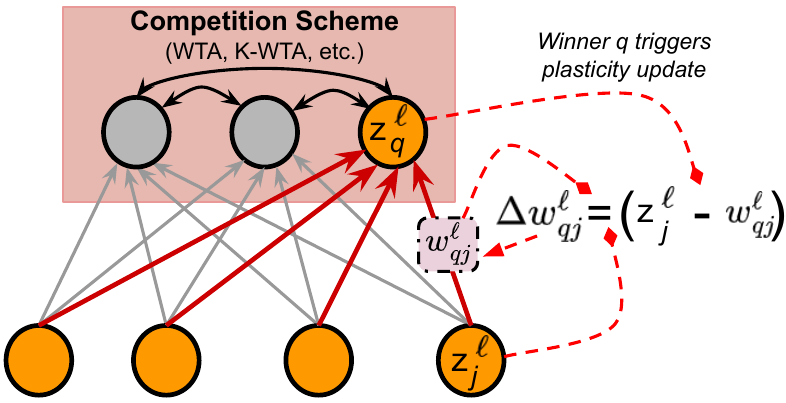}
  \end{center}
  \vspace{-0.255cm}
  \caption{\small{Competitive learning focuses on layer dynamics where neurons compete for the right to compute, i.e., they laterally inhibit or excite one another to form sparse distributed representations.
  }}
  \label{fig:comp_learning}
  \vspace{-0.3cm}
\end{wrapfigure}
%%%%%%%%%%%%%%%%%%%%%%%%%%%%%%%%%%%%%%%%%%%%

\begin{algorithm}[!t]
\caption{Procedure for selecting the (maximal) $K$ best matching units. }
\label{algo:competition}
\begin{algorithmic}
   \State {\bfseries Input:} Activation vector $\mathbf{z}^\ell$, $K$ BMU desired, winner\_type string flag
   \Function{FindBMU}{$\mathbf{z}^\ell$, $K$} \Comment Run $K$-winners selection process
        %\State $\mathbf{z} = \mathbf{0}$
        \State $w = \emptyset$, \; $z_{max} = \max(\mathbf{z}^\ell)$
        \Comment{Initialize statistics}
        \For{$k = 1$ to $K$}
        \State $q = \arg\max_i \mathbf{z}^\ell$, \; $\mathbf{z}^\ell[q,1] = z_{max}$ \Comment winner\_type is unit w/ maximum value
        \State $w \leftarrow w \cup \{q\}$ \Comment Record index $q$ of $k$th  winning neuron
        %\State $\mathbf{z}[1,i] = 1$ \Comment Set the index of the winning neuron to be $1$
        \EndFor
        \State \textbf{Return} $w$ \Comment Output set of $K$ winning neuronal units
    \EndFunction
\end{algorithmic}
\end{algorithm}

In general, the premise of competitive learning could be stated to center around the idea of `neuronal template matching' or clustering \cite{forgy1965cluster,lloyd1982least} -- a given pool of neuronal units fight for the right to activate leading to different (sets of) units activating for different clusters/partitions of patterns. As more and more sensory input patterns are presented to this pool of competing units, each neuron within the pool will converge to the center of the emergent cluster is has come to model. In other words, each neural unit will activate more strongly for sensory patterns strongly correlated with its cluster ``template'' and more weakly for those related to other cluster templates. As discussed in \cite{ororbia2021continual}, systems that operate under competitive neural dynamics are variations of the above story; this includes classical supervised systems such as vector quantization \cite{gray1984vector} to explicit topology clustering systems such as the venerable self-organizing map (SOM) \cite{kohonen1990self,luttrell1994bayesian,kohonen1995learning,gliozzi2018self} and self-organizing (incremental) neural networks \cite{luttrell1992self,furao2007enhanced}. Some schemes generalize the competitive dynamics to localized forms of competition or facilitating the emergence of multiple winning neurons (i.e., $K > 1$) as is the case for competitive Hebbian learning \cite{white1991competitive,martinetz1993competitive} and variations of compression systems based on neural gas  \cite{martinetz1993neural,fritzke1995growing}. %where neural prototypes are generated incrementally based on several criterion and are, crucially, guided by a relation graph between the units that is incrementally constructed and constantly updated

To fully characterize competitive neural systems, we borrow several perspectives \cite{rumelhart1986parallel,ororbia2021continual} to organize them under a few core foundational principles. In essence, a competitive model's computation can be broken down into:
% Scheme
% 1) calc set of measurements (distances, dot products)
% 2) apply selection function (also judgement function in literature) to get BMUs
% 3) calc synaptic adjustment (rule) and update
\begin{enumerate}[noitemsep,nolistsep]
    \item \textit{A Response Specificity Measurement}: the neuronal units start as highly similar (except for a randomized initial condition, which makes each unit respond slightly differently to a set of inputs). There is also a limit to the ``strength'' of each unit and activation in this context is produced by a set of measurements either as an array of distance values:
    \begin{align}
        \mathbf{z}^\ell[i,1] = \Big|\Big| \mathbf{z}^{\ell-1} - (\mathbf{W}^\ell)^{\mathsf{T}}[:,i] \Big|\Big|_p \text{for } i = 1,2,...,\mathcal{J}_\ell, \label{eqn:sub_dist}
    \end{align}
    or as a set of (parallel) dot products:
    \begin{align}
        \mathbf{z}^\ell[i,1] = \mathbf{W}^\ell[i,:] \cdot \mathbf{z}^{\ell-1} \; \text{for } i = 1,2,...,\mathcal{J}_\ell, \label{eqn:dot_prod}
    \end{align} %\mathbf{z}^\ell = \mathbf{W}^\ell \cdot \mathbf{z}^{\ell-1}
    where the $i$th element of the activation vector $\mathbf{z}^\ell$ is either a dot product (or, as in Equation \ref{eqn:sub_dist}, a subtraction fed into a $p$-norm function, e.g., $p=2$ yields the Euclidean distance) of the current pattern vector and the $i$th column of the transposed synaptic (memory) matrix; % The second approach follows a form of computation based on parallel dot products;
    \item \textit{A Competition Mechanism}: units compete for the right to respond to a particular subset of inputs -- this requires mechanisms for selecting what is known as the `best matching unit' (BMU), or the `prototype' (template) that satisfies a particular constraint.\footnote{Note that multiple BMUs may be selected/win the competition, as in growing neural gas \cite{fritzke1995growing}.} 
    A typical function for choosing a winning neuron(s) entails using the maximum, i.e., the $\max( )$ and $\arg\max( )$, out of a set of $\mathcal{J}_\ell$ activation values and is typically dependent on how the neural post-activities are computed in the first place. Once the values for $\mathbf{z}^\ell$ have been computed for neuronal layer $\ell$, the selection function is applied, as formally depicted in Algorithm \ref{algo:competition} (note that this algorithm is depicted as picking $K$ maximal neurons BMUs, stored in the list/set $w$).
    \item \textit{A Synaptic Adjustment Rule}: Given the results of the competition function or dynamics, synaptic efficacies are adjusted, typically in the form of a Hebbian or anti-Hebbian rule \cite{foldiak1990forming}, e.g., for a subtractive distance form of specificity, a synaptic update would be $\Delta \mathbf{W}^\ell = \mathbf{z}^{\ell-1} - (\mathbf{W}^\ell)^{\mathsf{T}}[:,q]$ where $q$ is a BMU index. The resultant update matrix $\Delta \mathbf{W}^\ell$ is then used to alter the values within $\mathbf{W}^\ell$ as in: $\mathbf{W}^\ell \leftarrow \mathbf{W}^\ell + \alpha \Delta \mathbf{W}^\ell$ 
    where $0 < \alpha < 1$ controls the magnitude of the update applied to parameters $\mathbf{W}^\ell$. %Note that other rules for changing the values of $\mathbf{M}$ are possible, such as an adaptive learning rate, e.g., Adam \cite{kingma2014adam} or RMSprop \cite{tieleman2012lecture}.
\end{enumerate}
%\textbf{1)} a best matching unit function, and \textbf{2)} a synaptic weight adjustment scheme. 
The above three components are important in the design of a minimal model of neural competitive learning. Competition dynamics across neuronal units notably results in sparse activity patterns, which has been demonstrated to be an invaluable biological property of neurons useful in preventing forgetting \cite{mccloskey_catastrophic_1989,french1999catastrophic} in systems such as sparse distributed memory \cite{kanerva1988sparse}. Many neural models driven by competitive learning embody the core principles above to varying implementational degrees, including incremental WTA models (e.g., vector quantization-based systems) \cite{gray1984vector}, self-organizing maps \cite{kohonen1988neural,kohonen1990self}, competitive neural Gaussian mixture models \cite{pinto2015fast,mclachlan2019finite}.
%%%%%%%%%%%%%%%%%%%%%%%%%%%%%%%%%%%%%%%%%%%%%%%%%%%%%%%%%%%%%%%%

%%Family # 2
\subsection{Global Explicit Signal Algorithms}
\label{sec:global_explicit_algos}
The next credit assignment family we examine embodies a different answer to our organizing question -- there is an explicit target that drives the learning process. These schemes effectively take a completely global approach to playing the credit assignment game, which is what backprop does; take a signal, such as an error measurement, originating at the output units and transmit (a transformation of) this signal back to each neuron inside the network. However, although these schemes do rely on a single (global) feedback pathway to ferry along teaching/adjustment signals, the design and nature of this pathway generally varies in form and nature in contrast to backprop. Within this category, more biologically-plausible variations of backprop have been proposed, e.g., random feedback alignment \cite{lillicrap2016random}, which notably offer means of resolving the problem weight transport, as well as methods based on either implicitly/explicitly modeling the act of (chemical) neuromodulation, e.g., three-factor Hebbian plasticity \cite{kusmierz2017learning}. 

\noindent
\textbf{Feedback Alignment.} Feedback alignment \cite{lillicrap2016random,launay2019principled,frenkel2021learning} -- also referred to as random feedback alignment (RFA) -- and its variants \cite{nokland2016direct,baldi2018learning,liao2016important} have shown that random feedback weights can also, surprisingly, still deliver useful teaching signals. In other words, feedback alignment algorithms resolve the weight-transport problem described earlier, showing that coherent learning is possible with asymmetric forward and backward pathways. Rather, the back-projection pathways for carrying backwards derivative information need not be transpositions of the connection weights used to carry out forward propagation; the process of credit assignment can instead be viewed as focused on the (partial) alignment of feedforward weights with (complementary) feedback weights. When the feedback and forward weights undergo a form of synaptic normalization and forced to approximate sign concordance \cite{baldi2016theory,baldi2018learning}, this form of learning can result, across various tasks, in networks with performance nearly as strong as those learned via backprop \cite{liao2016important}. %We will examine two particular instantiations of feedback alignment: random feedback alignment (RFA) and direct feedback alignment (DFA).% Pops - need to define acronyms for RFA/DFA...

\begin{figure}[!t]
\centering
\includegraphics[width=2in]{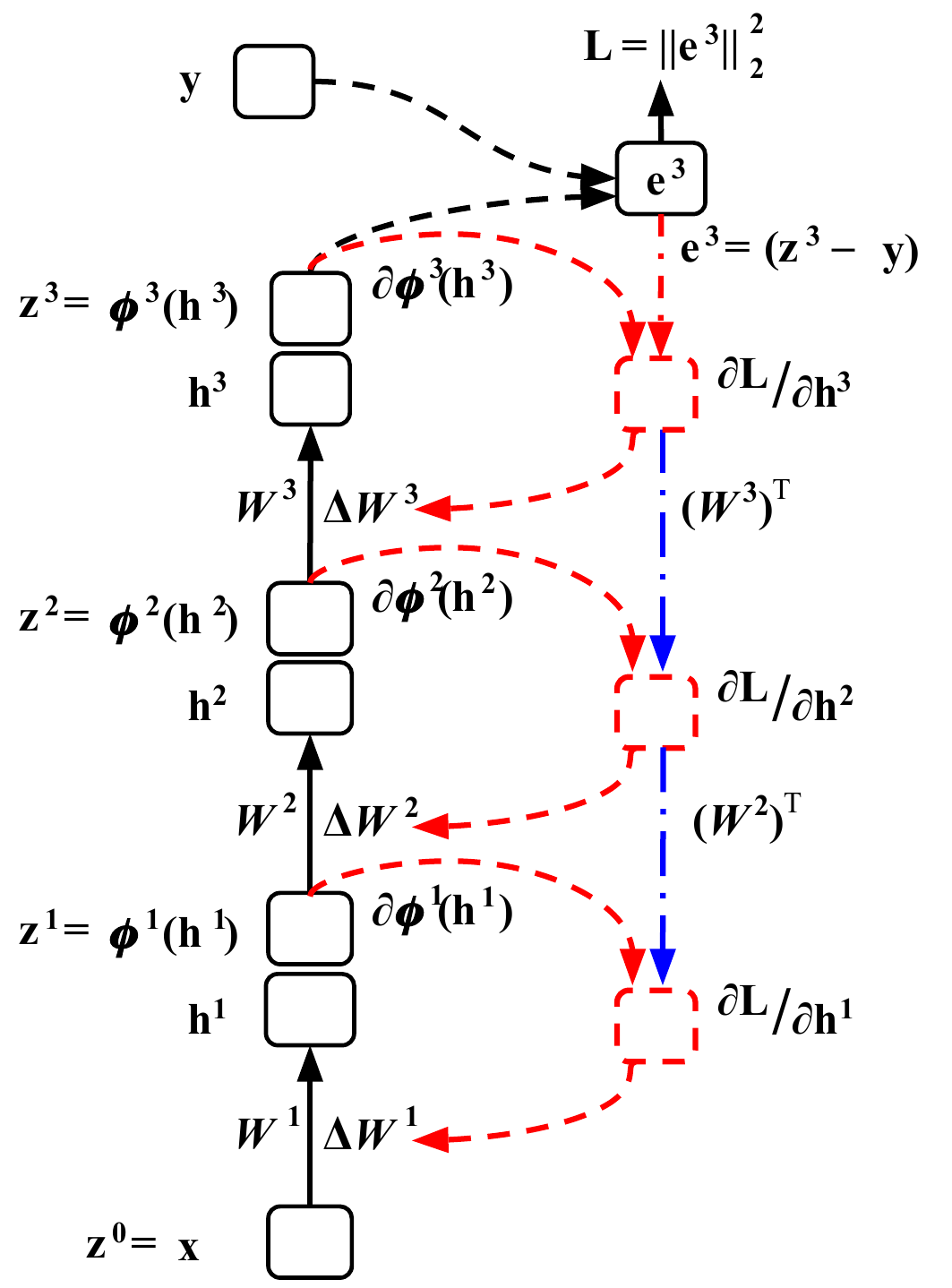}
~
\includegraphics[width=2in]{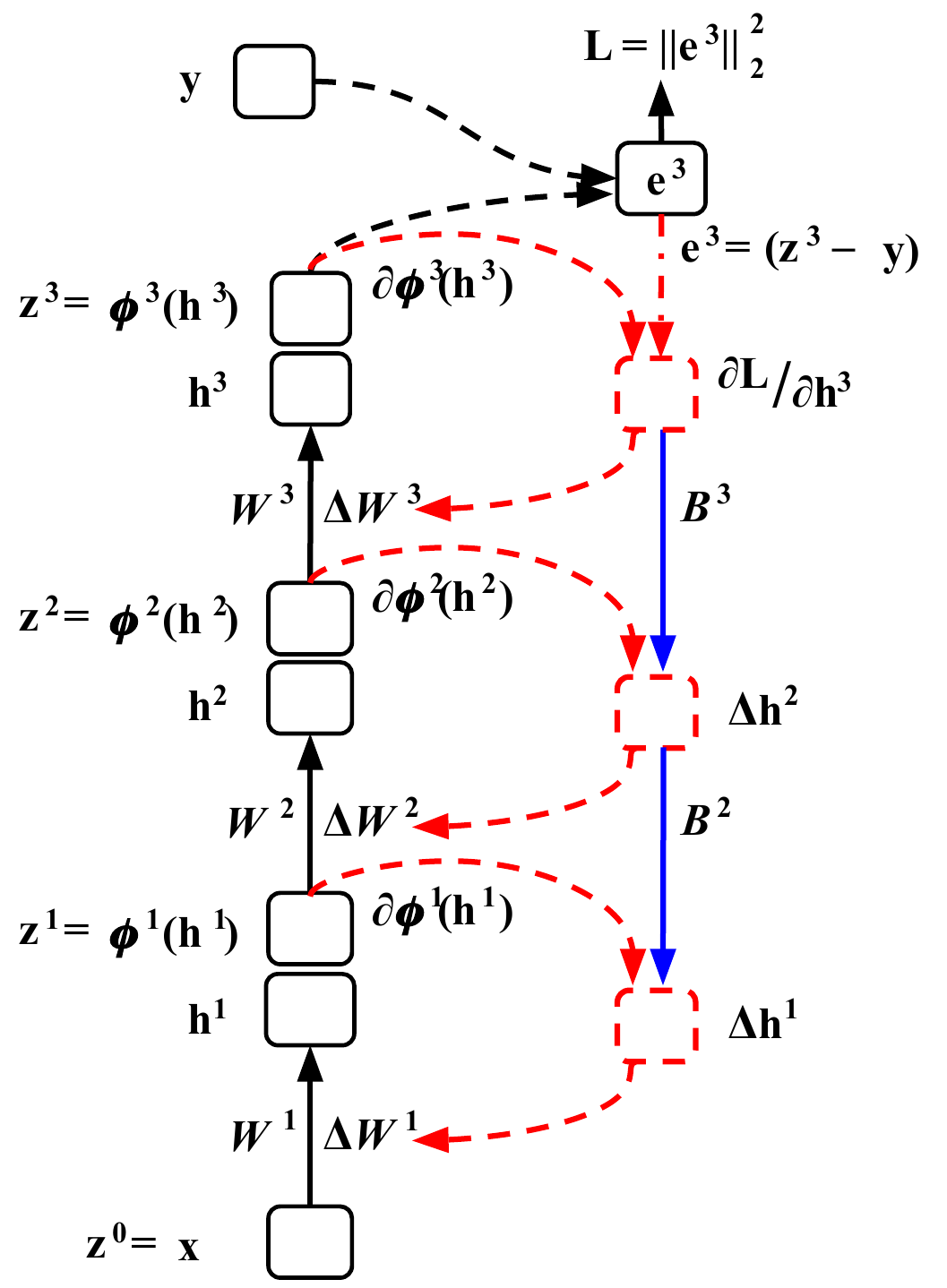}
~
\includegraphics[width=2in]{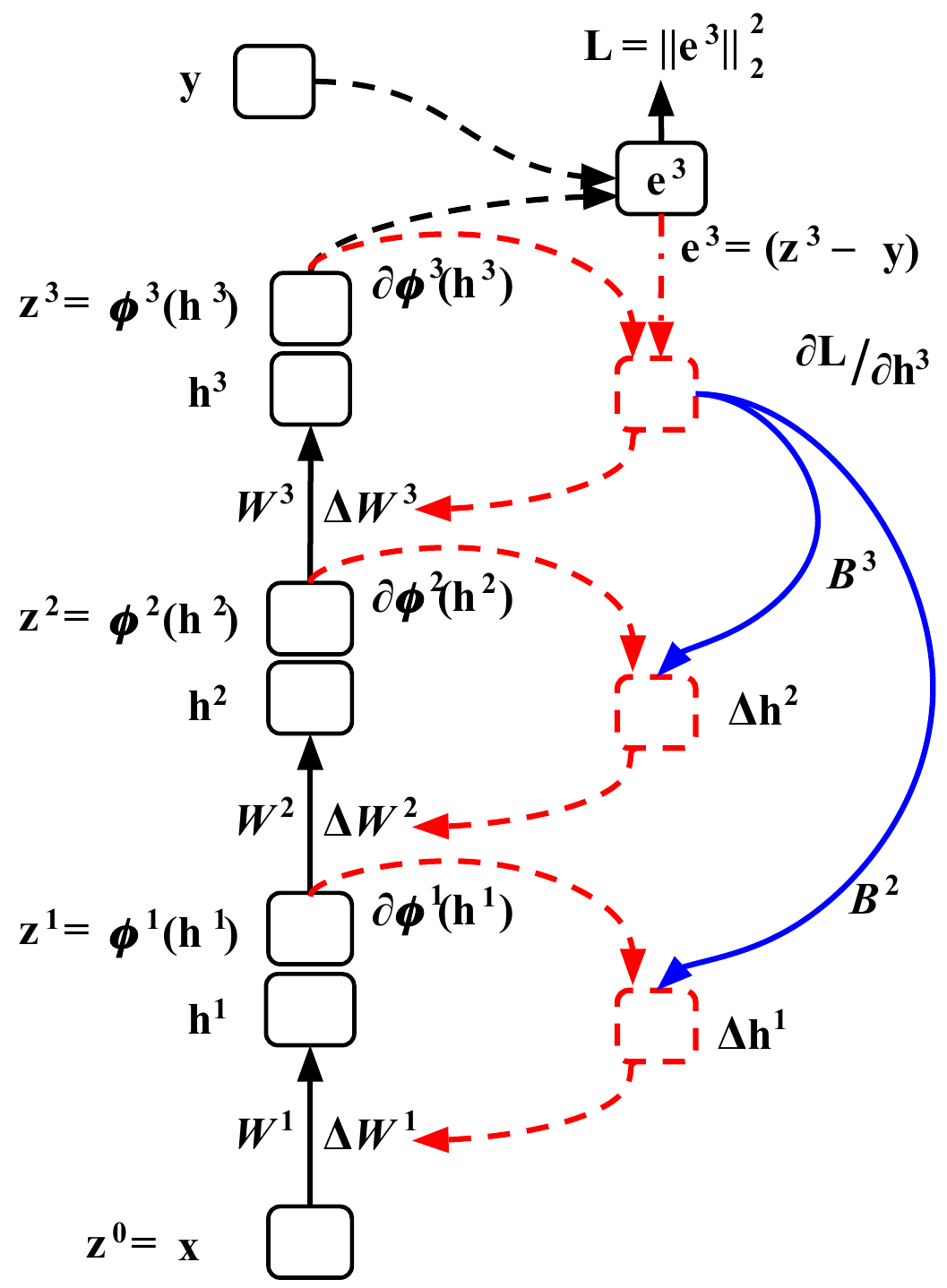}
\caption{\small{\textbf{Credit Assignment through Principles of Feedback Alignment.} 
Credit assignment schemes under backprop (Left), random feedback alignment (RFA; Middle), and direct feedback alignment (DFA; Right), applied to a three-layer MLP. Blue lines indicate the feedback pathway taken to produce teaching signals, or  $\Delta \mathbf{h}^\ell$ for each layer (which drives synaptic adjustment); solid lines indicate synaptic pathways while dash-dotted ones indicate ``virtual'' ones that depend on feedforward connections. In backprop, the (global) feedback pathway used to backwards propagate adjustment (derivative) information is made up of the transposes of the feedforward parameters. In contrast, RFA and DFA (on the same model) use separate, explicit parameter matrices to build either a feedback pathway similar to that of backprop (as in RFA) or a shorter, more direct one (as in DFA).
}}
\label{fig:global_feedback}
\vspace{-0.5cm}
\end{figure}

In essence, RFA is effectively backprop but with the key removal of its symmetric weight constraint for transmitting error signals back along the network; the updates produced by RFA look similar to those in Equation \ref{eqn:backprop_out_delta} (backprop) with the crucial difference that a teaching signal $\delta^\ell$ is produced by a synaptic pathway. Concretely, this means that the synaptic adjustment under RFA for any hidden layer is carried out as follows:
\begin{align}
    \Delta \mathbf{W}^\ell = \bigg( \Big( \mathbf{B}^\ell \cdot \frac{\partial \mathcal{L}}{\partial \mathbf{h}^{\ell+1}} \Big) \odot {\partial\phi}^\ell(\mathbf{h}^\ell) \bigg) \cdot (\mathbf{z}^{\ell-1})^{\mathsf{T}} = \Big(\widehat{\delta}^\ell \odot {\partial\phi}^\ell(\mathbf{h}^\ell) \Big) \cdot (\mathbf{z}^{\ell-1})^{\mathsf{T}}\,, \label{eqn:rfa_hidden_delta}
\end{align}
where the RFA teaching signal $\widehat{\delta}^\ell$ is produced via a product of a fixed (i.e., it is never adjusted), randomly initialized feedback matrix $\mathbf{B}^\ell \in \mathbb{R}^{\mathcal{J}_{\ell-1} \times \mathcal{J}_{\ell}}$ (the same shape as $(\mathbf{W}^\ell)^{\mathsf{T}}$. 
In the case of the direct feedback alignment (DFA) algorithm \cite{nokland2016direct}, the update for any hidden layer $\ell$ synaptic matrix $\mathbf{W}^\ell$ is instead:
\begin{align}
\Delta \mathbf{W}^\ell = \bigg ( \bigg ( \mathbf{B}^\ell \cdot \frac{\partial \mathcal{L}}{\partial \mathbf{h}^{L}} \bigg ) \odot {\partial\phi}^\ell(\mathbf{h}^\ell) \bigg ) \cdot (\mathbf{z}^{\ell-1})^{\mathsf{T}} \label{eqn:dfa_hidden_delta} \mbox{.}
\end{align}
Notice that, in DFA, while fixed random feedback synapses are also employed, they take on a very different form than those in RFA: feedback synapses are directly wired to skip from the output layer to each layer directly, rather than traversing backwards across each layer of activities as in RFA. In either RFA or DFA, during the learning step, the forward weights are adapted to bring the network to a regime where the random backward weights are able to carry information that roughly approximates the gradient produced by backprop. Notably, this form of ``random backpropagation'' \cite{baldi2018learning,refinetti2021align} has also been used to develop an event-driven variation of the learning rule suitable for spiking neural networks \cite{neftci2017event,samadi2017deep}. 
In addition, empirical results have been obtained to show that schemes based direct feedback alignment (DFA) can scale to large-scale datasets  \cite{moskovitz2018feedback,launay2020direct}; however, it has also been observed that such schemes fail to learn efficiently \cite{bartunov2018assessing,moskovitz2018feedback,refinetti2021align}, particularly in the case of convolutional networks, which stems from the inherent challenge of aligning the (very sparse) Toeplitz matrix formulation of convolution towards the randomized feedback parameters of alignment frameworks. 

An interesting variation of RFA/DFA was furthermore proposed, labeled as indirect feedback alignment (IFA), which also employed skip-layer fixed feedback pathways but in tandem with the feed-forward inference machinery:
\begin{align}
\Delta \mathbf{W}^\ell = \bigg ( \widehat{\delta}^\ell \odot {\partial\phi}^\ell(\mathbf{h}^\ell) \bigg ) \cdot (\mathbf{z}^{\ell-1})^{\mathsf{T}} \label{eqn:ifa_hidden_delta}, \; \text{where } \; 
\widehat{\delta}^\ell = 
\begin{cases} 
  \mathbf{B} \cdot \frac{\partial \mathcal{L}}{\partial \mathbf{h}^{L}} & \mbox{if} \; \ell = 1
  \\
  \mathbf{W}^\ell \cdot \Big( \widehat{\delta}^{\ell-1} \odot {\partial\phi}^{\ell-1}(\mathbf{h}^{\ell-1}) \Big)  & \mbox{otherwise.} 
\end{cases}
\mbox{.}
\end{align}
We highlight IFA given that it: 
\textbf{1)} introduces a short feedback pathway using only a single matrix of synapses, and 
\textbf{2)} it reuses the feedforward pathway (unmodified) to produce its teaching signals. The latter of these two useful properties serves as a useful basis for some algorithms that fall under a more recent family of credit assignment that we will later review -- forward-only learning (see end of Section \ref{sec:synergistic_local_explicit_algos}). % a useful property for implementations on photonic chips \cite{launay2020hardware,filipovich2022silicon}

The underlying premise of random backpropagation has spurned a chain of developments over the years \cite{baldi2018learning,crafton2019direct,han2020extension,chu2020training}. Some variants of feedback alignment change what is propagated along the fixed random pathways \cite{frenkel2019learning} (such as using the label directly instead of the first derivative of the cost function) while others address the problem of using fixed random feedback parameters; alignment approaches based on underlying ideas of the Kolen-Pollack (KP) method \cite{kolen1990back} adjust the feedback connections using the (transpose of the) same update produced for the feedforward ones in tandem with a synaptic decay (regularization) term. Schemes that introduce a form of this complementary feedback adjustment, e.g., `weight mirrors' (WM) \cite{akrout2019deep}, rest on the theoretical premise that, given enough adjustments over time, the feedback synapses will approximately converge to the transpose of the forward ones (which, consequentially, means a weight-mirrored RFA recovers backprop in the limit). Alignment schemes such as weight mirrors have yielded promising results on large extensive benchmarks, all without requiring weight transport and even when incorporated into convolutional frameworks. Very recent credit assignment schemes have combined the notion of weight mirroring in the context of complementary (global) forward and backward pathways \cite{cheung2018many,toosi2023brain} (this scheme combines RFA with concepts of target propagation \cite{bengio2014auto}, reviewed later). Other variant schemes take the sign of the forward activities instead, such as the `sign symmetry' method \cite{xiao2018biologically}, where the feedback synapses that generate the requisite teaching signals share only the sign but not the magnitude of the feedforward ones \cite{liao2016important}. It is worth pointing out that sign symmetry-based approaches lack theoretical justification and such schemes do not accurately propagate either the magnitude or sign of the error gradient. Nevertheless, their performance has been empirically shown to be only slightly worse than backprop (even when using tensor connections as in convolutional  architectures \cite{xiao2018biologically}). % note: xiao's method takes the sign of the forward activities

\noindent
\textbf{Neuromodulatory Approaches.} Another set of approaches within this family are those that, instead of implementing an explicit transmission pathway in terms of (separate) synapses to place credit/blame on individual neuronal units, produce adjustments by broadcasting a `modulatory' signal $M$ that will drive (typically locally-generated) synaptic adjustments. In the brain, neuromodulators, e.g., dopamine, norepinephrine, or acetylcholine, represent viable biological candidates for broadcasting success/failure (or uncertainty/surprisal) signals across neurons  \cite{schultz1997neural,angela2005uncertainty,schultz2010dopamine}.
As such, this sub-family as labeled `neuromodulatory' (or dopamine-modulated) credit assignment, largely from the fact that many approaches within it draw inspiration from the concept of `neurotransmittion'.\footnote{In neurobiology, a neurotransmitter is a chemical messenger that strives to carry/move a bio-chemical signal from one neuron (such as a nerve cell) to another.} 

%%%%%%%%%%%%%%%%%%%%%%%%%%%%%%%%%%%%%%%
\begin{figure}[!t]
\centering     %%% not \center
\begin{subfigure}{0.495\textwidth}
    \centering
    %\vspace{-5mm}
     \includegraphics[width=0.495\linewidth]{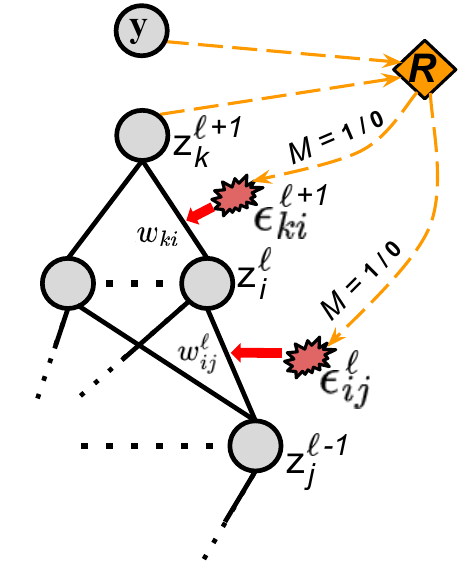} 
    % \caption{Modulated random perturbation-based scheme.}
    % \label{fig:neuromod_rand}
\end{subfigure}
\begin{subfigure}{0.495\textwidth}
    \centering
    %\vspace{-5mm}
     \includegraphics[width=0.495\linewidth]{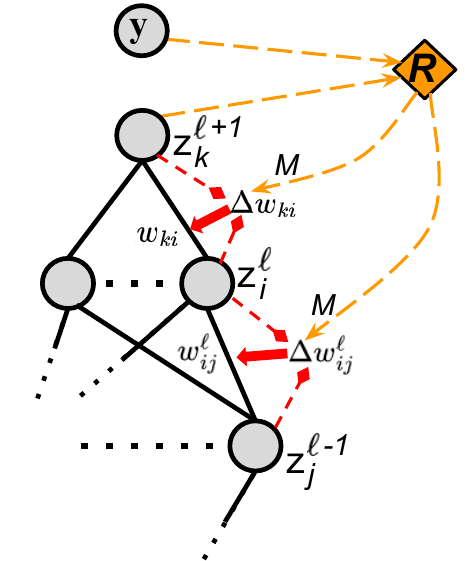} 
    % \caption{Dopamine-modulated Hebbian (three-factor) scheme.}
    % \label{fig:neuromod_hebb}
\end{subfigure}
%\vspace{-0.3cm}
\caption{\small{\textbf{Neuromodulatory Approaches to Credit Assignment.} 
Depicted are forms of neuromodulation for adjusting synapses in a neural circuit (in a globally-synchronized, parallel fashion). $R$ is a dopamine (reward) functional that encodes behavioral goals and is used to produce a modulatory signal $M$ that is either: (Left) a (binary) gating variable for accepting/rejecting a synaptic adjustment to be made, or (Right) a value to scale the update yielded by local Hebbian plasticity.
}}
\label{fig:neuromod}
\vspace{-0.5cm}
\end{figure}
%%%%%%%%%%%%%%%%%%%%%%%%%%%%%%%%%%%%%%%

%% the reason this falls here is b/c we still use a globally available cost function to direct the optimization process (even with temperature scheduling)
How this globally-modulated form of adjustment occurs varies greatly across schemes. For example, a rather simple way to view  neuromodulation is through an implementation of a form of stochastic hill-climbing, where an ensemble of noisy perturbations is directed by the (possibly task-specific) cost function(s) $\mathcal{L}$; see Figure \ref{fig:neuromod} (Left). In essence, a random process is leveraged to ``confabulate'' a set of possible directions that synaptic weights could be nudged towards and, so long as these directions are related back to an objective (with which the current state of the network can be directly correlated to), e.g., a dopamine/reward functional $R$ \cite{barto1992gradient} that could encode a typically cost $\mathcal{L}$ and target values $\mathbf{y}$ (as shown in Figure \ref{fig:neuromod}), the neural system could be gradually moved to more promising configurations. Credit assignment schemes that operate like this are, effectively, evolving synapses by largely disregarding the nature by which information flows through a neural system (to produce its internal distributed representations) and instead move along the cost function (e.g., error) space via random perturbations of synaptic connections, checking the global cost function  along the way \cite{takefuji1991artificial,chalup1999study}. A simplified variation of such an update, per synapse, would be: $\Delta w^\ell_{ij} = M \epsilon^\ell_{ij}$, where $M = \{0,1\}$ is a binary variable produced by reward functional (for instance, $M = 1$ if $R_t - R_{t-1} > 0$ in the case of reward maximization). 
Note that variations of stochastic hill-climbing that fall under this family do not require gradient information nor any functional aspect of the system to be differentiable (including the cost function itself) -- these approaches generally correlate the current state/configuration of a network, under random fluctuation(s), to a desired objective, deciding if a particular nudge/move that could be made results in performance improvement. Variations of this theme  incorporate synaptic feedback pathways, much akin to those studied in the previous section (e.g., RFA/DFA pathways), and have even shown that noise-based (feedback) modulation cycles can approximate the gradients produced by backprop \cite{lansdell2019learning}.  %An reinforcement perturbation scheme constructs a noise-based feedback modulation cycle \cite{lansdell2019learning} was shown to approximate the gradients produced by backprop.

An improved format for conducting the above cost/reward-driven perturbation style of credit assignment is simulated annealing (SA) \cite{pincus1970monte,kirkpatrick1983optimization}, which is a generalization of stochastic hill-climbing that decides on which random moves to take based on a probability of acceptance, guided by a temperature/settling schedule and a comparison of the current state of a cost functional with its previous value. Notably, SA has been used to train ANNs before \cite{chen1995chaotic}, including the well-known, classical Boltzmann machine \cite{hinton1986learning,aarts1988simulated}, despite its origins as a metaheuristic inspired by the heating/cooling of materials to change their fundamental properties in metallurgy \cite{pincus1970monte,kirkpatrick1983optimization}. Although it is unlikely that longer-term plasticity in neuronal networks could be likened to purely random walk-like processes (which renders random hill-climbing less plausible than other schemes, e.g., Hebbian ones), a cost-modulated form of perturbation offers a massively parallel scheme for synaptic change -- given that only random noise values, the cost function, and a synapse are required at any given time, parameter adjustment is local while behaviorally driven/guided by a task-centric signal, i.e., the cost functional. Neuro-evolutionary frameworks and strategies represent an ever-growing set of metaheuristic efforts (which greatly improve upon the premise of modulated stochastic adjustments) \cite{stanley2002evolving,igel2003neuroevolution,salimans2017evolution} % ororbia2019investigating}
for searching for optimal synaptic values.\footnote{We remark that, given this paper's focus is on single-agent systems, rather than multi-agent or population-based ones, reviewing this branch of metaheuristic optimization further is beyond the particular scope of this survey. For relevant reviews on this topic, we refer the reader to \cite{floreano2008neuroevolution,galvan2021neuroevolution}.} 

Other approaches build on the premise of global modulation but do not resort to exclusively random perturbation(s); these schemes utilize forms of reinforcement to correlate the activities of individual neurons with a global objective in the form of reward function \cite{mazzoni1991more,barto1992gradient,williams1992simple}; note that some variations combine noise with reward modulation \cite{fiete2006gradient}. The assumption is that neurons, which behave stochastically, receive a signal that describes/encapsulates the error at the output, i.e., the ``neuromodulator'' in dopamine/reward-drive adaptation. When an improvement is observed in an objective or reward functional, the connection weights are changed so as to make the produced activity associated with the positive change more likely. The AuGMEnT framework, which produces adjustments labeled as `synaptic tags', represents one such neuromodulatory-centered credit assignment scheme \cite{roelfsema2010perceptual,rombouts2015attention,rombouts2015learning,pozzi2018biologically} and one that is generally combined with other neurocognitively-plausible mechanisms such as models of working memory \cite{kruijne2021flexible,yoo2022working}. Other relevant modulation approaches center around the error (or first derivative) associated with the output cost function and craft mechanisms for broadcasting a value correlated to this error -- as in error-vector broadcasting \cite{clark2021credit} -- while other schemes broadcast a context value (label/regression target) -- as in the dendritic gating framework \cite{budden2020gaussian,sezener2021rapid}, e.g., Gaussian linear-gated neural networks.  %and contrastive variations \cite{illing2021local} (which further mixes a local prediction of neural activity to produce a modulatory signal per neuron). %(globally-broadcast terms multiplied by a simple top-down/local prediction of neural activity yields the third term)
Notably, it is in the context of neuromodulation that three-factor Hebbian plasticity emerges (Hebbian 3F) \cite{porr2007learning,pfeiffer2010reward,kusmierz2017learning,gerstner2018eligibility}, as visually depicted in Figure \ref{fig:neuromod} (Right). For instance, if we focus on three-factor credit assignment using dopamine/reward modulation \cite{reynolds2002dopamine,loewenstein2008robustness,pfeiffer2010reward}, then we can replace the perturbation factors described earlier, e.g., the $\epsilon_{ij}$ as in Figure \ref{fig:neuromod} (Left), with synaptic adjustments generated by a two-factor Hebbian rule, such as one present in Section \ref{sec:implicit_algos}, to obtain: $\Delta w_{ij} = M(t) \Big( W_{ij} ( z^\ell_i z^{\ell-1}_j ) \Big)$ or, more generally, $\Delta \mathbf{W}^\ell = M \Big( \mathbf{W}^\ell \odot (\mathbf{z}^\ell \cdot (\mathbf{z}^{\ell-1})^{\mathsf{T}}) \Big)$. Note that the modulator $M$ can be a function of time $M(t)$ as well as the dopamine/reward functional, having taken on different forms in the literature; an oft-chosen form is the temporal difference error: $M(t) = r(t) - \mathbb{E}\big[ r(t) \big]$, where $\mathbb{E}\big[ r(t) \big]$ is an estimate of the expected reward, i.e., typically a dynamically calculated running average. This reward-modulated multi-factor setup is further modified to incorporate what is known as an \emph{eligibility trace}\footnote{Mechanistically, such a trace offers a temporary record or memory of an event's occurrence, e.g., taking a specific action, and provides a marker or ``tag'' of this event as eligible for inducing a form of adaptation (or learning). When a modulator triggers synaptic adjustment, only events/states (and associated actions) that are deemed eligible are assigned credit/blame in association with the (global) error, i.e., eligibility tracing facilitates a basic form of temporal credit assignment. } $\mathbf{E}^\ell(t)$; this yields a set of ordinary differential equations that characterizes a generalized Hebbian plasticity scheme via the following related differential equations:
\begin{align}
    \tau_e \frac{\partial \mathbf{E}^\ell(t)}{\partial t} &= -\mathbf{E}^\ell(t) + \Big( \mathbf{W}^\ell \odot (\mathbf{z}^\ell \cdot (\mathbf{z}^{\ell-1})^{\mathsf{T}} \Big) \label{eqn:elg_trace} \\ 
    \Delta \mathbf{W}^\ell \propto \tau_w \frac{\partial \mathbf{W}^\ell(t)}{\partial t} &= M(t) \Big( \mathbf{E}^\ell(t) \odot \Big( \mathbf{W}^\ell \odot (\mathbf{z}^\ell \cdot (\mathbf{z}^{\ell-1})^{\mathsf{T}} \Big) \Big) \label{eqn:reward_hebb}
\end{align}
where $\tau_e$ and $\tau_w$ are the eligibility and synaptic plasticity time constants, respectively (both generally set to be on the order of milliseconds). An eligibility trace, which has support in neuroscience \cite{froemke2007synaptic}, is required particularly in light of action-oriented/behavioral tasks; such as those that characterize reinforcement learning \cite{sutton2018reinforcement}. Given that success/failure identification can come after a delayed period time once an action has been taken, neuronal networks will, in some way, need to store a short-term memory of past actions taken and, neurobiologically, one suitable location for such a memory is at the synapses themselves.\footnote{Biologically, it is postulated that co-activation of pre- and post-synaptic neuronal cells provides/sets a ``flag'' (at the connecting synapse) that is used in a later weight adjustment in the presence of a trigger; such a trigger could be reward/punishment, novelty, or surprise, which could be implemented by either particular special neuronal signaling events or the phasic activity of neuromodulators. If dopamine receptors are blocked, then synaptic change (long-term potentiation induced via STDP) occurs.} Notably, three-factor Hebbian plasticity (sometimes referred to as ``neoHebbian plasticity'') has proven to be invaluable for adapting the parameters of spiking neuronal systems \cite{fremaux2016neuromodulated}, resulting in what is known as reward-modulated spike-timing-dependent plasticity (R-STDP) \cite{izhikevich2007solving,legenstein2008learning,fremaux2010functional}. Note that the process of R-STDP enjoys considerable experimental validation and support in neurobiological studies \cite{chase2009functional,pawlak2010timing,brzosko2017sequential,gerstner2018eligibility}.

%% The Bottleneck Approach -- this is kind of like a neuro-mod type of approach (signal perturbations applied to all neurons/layer in system at once) == violates external control mechanism somewhat
Another important approach that falls under this family takes its motivation from information theory -- the Hilbert-Schmidt independence criterion (HSIC) bottleneck algorithm \cite{ma2019hsic}. HSIC credit assignment, which is based on the Information Bottleneck (IB) principle \cite{tishby2000information}, %performs credit assignment locally and layer-wise
operates with the aim of seeking (internal/hidden) representations that exhibit high mutuality with target value(s) and less mutuality with input patterns (or neural activities) presented to each layer. In other words, this approach is not driven by a dependency on the information flow (propagated layer-by-layer) through the neural model. Notably, some work has demonstrated that a kernelized version of HISC can recover three-factor Hebbian rules \cite{pogodin2020kernelized}.

%% issue with neuromod/mod approaches
While approaches within this family exhibit some very appealing properties from a biological standpoint, it remains a challeng to scale them to complex, high-dimensional tasks \cite{werfel2004learning,hiratani2022stability}.  For instance, some of these approaches (such as SA) exhibit slow convergence, requiring many iterations in order to find optimal solutions. 
%%%%

%%Family # 3: 
\subsection{Non-Synergistic Local Explicit Signal Algorithms}
\label{sec:explicit_local_algos}

%Talk about algorithms use disconnected yet parallelizable modules for producing the driving/target signals that will trigger/guide plasticity
As we have seen in the previous few sections, signals for driving changes in plasticity can either be implicit -- there is no particular target signal and  statistics/correlational properties purely local to any given synapse are used -- or single explicit yet global -- a signal, such as an output error or a dopamine functional, is directly broadcasted to any given synapse. In contrast to these types of types of driving signals are what we refer to within our credit assignment taxonomy as \emph{explicit local target signals}. 
These type of signals are further broken down in this work to be either \emph{non-synergistic} or \emph{synergistic}. For non-synergistic local signals, a credit assignment scheme would utilize (neural) machinery (such as a local classifier or a inversion model/decoder), which is itself situated within the immediate vicinity of synapses to be adapted. This local mechanism produces driving signals that operate with information only available to a particular block/region of the network (or its ``computational subgraph''), typically focusing on a pair of inter-connected neuronal layers. In contrast, as discussed later in Section \ref{sec:synergistic_local_explicit_algos}, synergistic local signals are those that are iteratively produced by neural machinery that often takes the form of a settling process or message passing schemes. It is important to note that, in a non-synergistic local credit assignment algorithm, the global link between the updates across regions/layers of neurons is through the flow of information across the whole network via inference (e.g., a feedforward pass in our MLP example).
% non-synergistic  rules can even take on different polynomial forms, such as first, second, and third order \cite{baldi2018learning}. 

% globally synergistic by feedback among update models/predictors
\noindent
\textbf{Synthetic Local Updates (SLU).} An interesting alternative to backprop is a hybrid algorithmic and architectural approach referred to as error critic(s) \cite{werbos1992approximate}, local gradient estimators \cite{schmidhuber1990networks,schmidhuber1990recurrent} and meta-learning (local) backprop \cite{kirsch2021meta}, or decoupled neural interfaces \cite{jaderberg2016decoupled,czarnecki2017understanding,belilovsky2021decoupled,zhuang2021fully}, which, at least originally, was not methodology directly motivated by biological considerations though later connections to neurobiology have been made \cite{pemberton2021cortico}. %% deeply nested systems \cite{miguel2014nested}; ADMM \cite{taylor2016training}; block-coordinate learning \cite{zhang2017convergent}
Other complementary schemes related to the goal of decoupling include the method of alternating direction method of multipliers (ADMM) \cite{taylor2016training} and block-coordinate descent-centric training \cite{zhang2017convergent}. % aim at better utilization of highly paralleized computing resources / improve the parallelism in training DNNs by decoupling the layer dependencies
The primary motivation behind this work, which we will broadly refer to as `synthetic local updates', came from considering the three practical computational flow problems that plague neural systems described in Section \ref{sec:backprop_problems}; forward locking, update locking, and backwards locking. % -- 1) forward locking, where no layer can process incoming data before the layers that precede it have activated, 2) update locking, where no set of weights for a layer $\ell$ can be updated before all the other dependent layers have been activated, and 3) backwards locking, where no layer's parameters can be updated before all dependent layers have been executed in both forward and backward modes. 
These three locking problems constrain the computation in a neural system context to be purely linear, hindering how much benefit one can take from massively parallel computing systems.

In terms of our survey's MLP architectural running example, the basic idea is to augment each layer with an additional parametric model, one that learns to \emph{predict weight updates}, or gradients in the context of differentiable systems. Assuming the simplest possible `update synthesizer', a linear predictor, we would only need to introduce one additional synaptic weight matrix $\mathbf{K}^\ell$ (similar in spirit to the random local classifier approach described above). This would mean that the computation of a synaptic adjustment within the MLP would be:
\begin{align}
\Delta \mathbf{W}^\ell &= \Big( \Delta \mathbf{z}^\ell \odot \partial\phi^\ell(\mathbf{h}^\ell) \Big) \cdot (\mathbf{z}^{\ell-1})^{\mathsf{T}} \\
&\text{where } \Delta \mathbf{z}^\ell  = \phi_p(\mathbf{K}^\ell \cdot \mathbf{z}^\ell). \nonumber
\end{align}
$\phi_p(\circ)$ is the post-activation of the gradient/update predictor, which can be set to the identity function to recover the setting of \cite{jaderberg2016decoupled}. 
In the equation above, we see that during feedforward computation, layer by layer, changes in plasticity may be readily computed along the way. Note that each $\Delta \mathbf{z}^\ell$ is an approximation of the true update/gradient, or rather, $\frac{\partial \mathcal{L}(\mathbf{y},\mathbf{z}^L)}{\partial \mathbf{z}^\ell} = \delta^\ell \approx \Delta \mathbf{z}^\ell$. To train the weight update prediction models for any layer $\ell$, we simply use the proxy update from the layer above $\ell+1$, and compare the layerwise predicted update with the one determined above via backward transmission. Assuming a local cost, e.g., mean squared error, measuring the difference between the top-down generated gradient and the predicted gradient, $\mathcal{L}_d(\delta^\ell, \Delta \mathbf{z}^\ell)$, the updates to the update synthesizer are:
\begin{align}
\delta^L &= \frac{\partial \mathcal{L}(\mathbf{y}, \mathbf{z}^L)}{\partial \mathbf{z}^L}, \; \Delta \mathbf{K}^L \propto \frac{\partial \mathcal{L}_d(\delta^L, \Delta \mathbf{z}^L)}{\partial \mathbf{K}^L} =  \Big( (\Delta \mathbf{z}^L - \delta^L) \odot \partial\phi_p(\mathbf{K}^L \cdot \mathbf{z}^L) \Big) \cdot (\mathbf{z}^{L-1})^{\mathsf{T}} \\
\delta^\ell &= (\mathbf{W}^{\ell+1})^{\mathsf{T}} \cdot \Delta \mathbf{z}^{\ell+1} , \; \Delta \mathbf{K}^\ell \propto \frac{\partial \mathcal{L}_d(\delta^\ell, \Delta \mathbf{z}^\ell)}{\partial \mathbf{K}^\ell} =  \Big( (\Delta \mathbf{z}^\ell - \delta^\ell) \odot \partial\phi_p(\mathbf{K}^\ell \cdot \mathbf{z}^\ell) \Big) \cdot (\mathbf{z}^{\ell-1})^{\mathsf{T}} \mbox{.}
\end{align}
The above local generator approach, which resolves the update (and backward) locking problem, can be taken even further and used to generate the activity values themselves, which completely decouples the entire MLP network and resolves the forward locking problem (see \cite{jaderberg2016decoupled} for details). The interpretation that we will take of a local gradient/update synthesizer scheme is that each layerwise predictor serves as a form of (hetero-associative) memory; in other words, we may instead view each $\mathbf{K}^\ell$ as a memory matrix that aims to recall error gradient patterns generated for a given target layer. These locally-integrated memory modules effectively decouple the computations that occurs across the layers in a deep neural system -- each gradient (and/or activation) predictor iteratively and progressively learns to become estimators of the kinds of synaptic/ativation updates that result in better latent representations of the input/output data pairing. %links together, or rather, coordinates, the various layers of the network.

\begin{figure}[!t]
\centering     %%% not \center
\begin{subfigure}{0.495\textwidth}
    \centering
    %\vspace{-5mm}
     \includegraphics[width=0.755\linewidth]{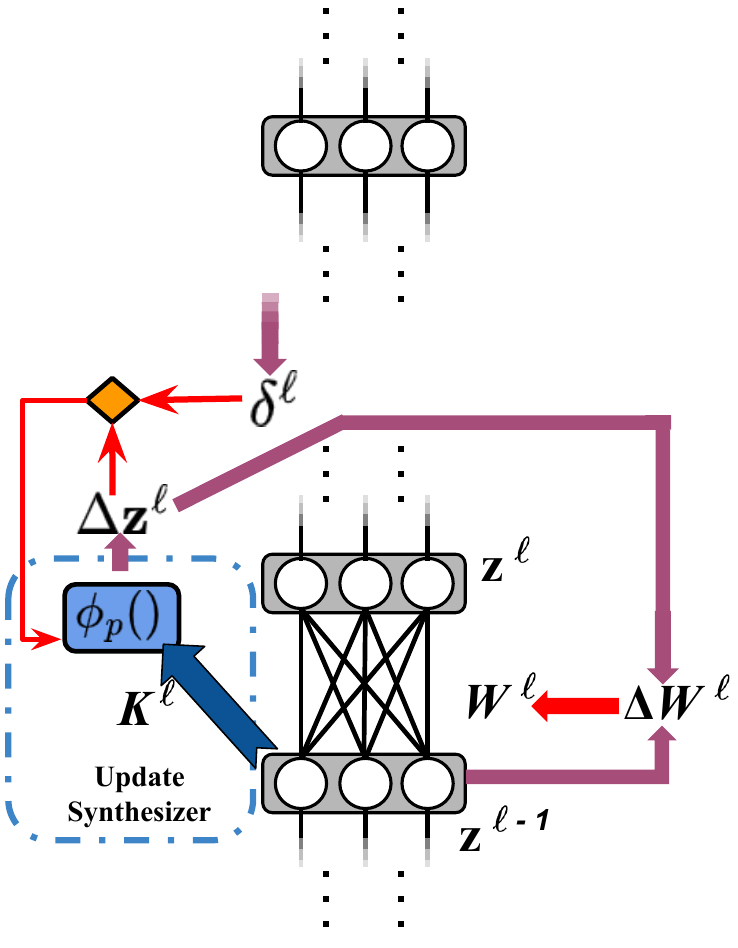}
    % \caption{Synthetic local updates.}
    % \label{fig:synth_grad}
\end{subfigure}
\begin{subfigure}{0.495\textwidth}
    \centering
    %\vspace{-5mm}
     \includegraphics[width=0.675\linewidth]{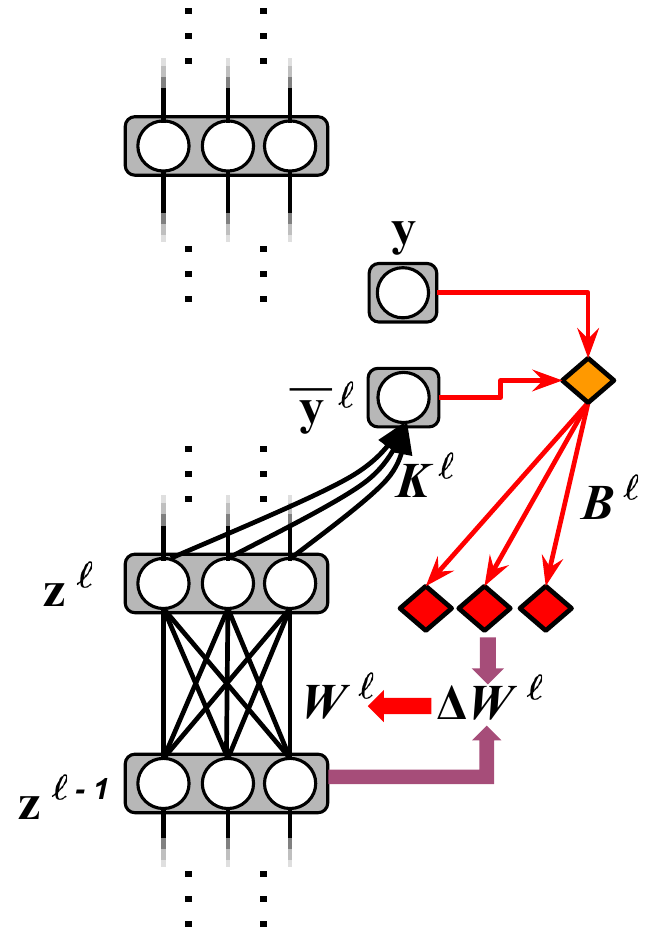} 
    % \caption{Local signalers.}
    % \label{fig:local_signaling}
\end{subfigure}
\vspace{-0.25cm}
\caption{\small{\textbf{Non-synergistic Forms of Credit Assignment.} Orange diamonds represent error/mismatch. 
(Left) In a synthetic local update scheme, an extra set of neurons synthesizes an update $\Delta \mathbf{z}^\ell$ to be used as part of a plasticity update $\Delta \mathbf{W}^\ell$. The synthesizing module is itself updated with a value of the real local (layer) gradient $\delta^\ell$. 
(Right) In a local signaling process, the neural sub-module attached to layer $\ell$, aimed at predicting regression target $\mathbf{y}$ using fixed random parameters $\mathbf{K}^\ell$, provides a local error message that is transmitted back along fixed random synapses $\mathbf{B}^\ell$ to provide a post-synaptic term that triggers a plasticity adjustment $\Delta \mathbf{W}^\ell$ to synapses $\mathbf{W}^\ell$ that connect neuronal units in layers $\ell-1$ (i.e., $\mathbf{z}^{\ell-1}$) to those in $\ell$ layers (i.e., $\mathbf{z}^{\ell}$).
}}
\label{fig:local_schemes}
\vspace{-0.6cm}
\end{figure}

\noindent
\textbf{Local Signalers (Predictors).} One scheme that has been shown to be effective for conducting credit assignment in neural systems, based on non-synergistic signals, is through the integration of a set of layer-wise, often non-learnable ``signalers'', i.e., predictors or estimators for classification or regression, in conjunction with the idea of random fixed feedback from feedback alignment. This characterizes the central idea proposed in \cite{mostafa2018deep}. % and we generalize the algorithm and present it in full detail in Algorithm \ref{alg:random_local}. 
Specifically, for each (hidden) layer $\ell$ of the model, e.g., our MLP, a neural classification module (or possibly a regression module, depending on the type of supervised task being tackled) parameterized by fixed, random synaptic weights $\mathbf{K}^\ell$ that project the activity values of layer $\mathbf{z}^\ell$ and are followed by a normalized nonlinearity so as to produce a set of class scores $\mathbf{\bar{y}}^\ell$. As a result, calculating the update to any layer's set of connection weights $\mathbf{W}^\ell$, except for the top-level ones, amounts to calculating the partial derivative $\frac{\partial \mathcal{L}(\mathbf{y},\mathbf{\bar{y}}^\ell)}{\partial \mathbf{W}^\ell}$. A further modification to computing this gradient is that the transpose of the local classifier's weight matrix $(\mathbf{K}^\ell)^{\mathsf{T}}$ is replaced with a fixed random matrix $\mathbf{B}^\ell$ of the same shape. Interestingly enough, it is not necessary to learn the parameters of this random classifier, which means that, in reality, no extra parameters have been introduced to the model from the perspective of credit assignment; it is further relatively simple to align each pair of classification $\mathbf{K}^\ell$ and feedback $\mathbf{B}^\ell$ weights at initialization. This yields the following synaptic update for any layer $\ell$ in our MLP (except for the topmost layer $\ell = L$):
\begin{align}
    \mathbf{\bar{y}}^\ell &= \phi_y(\mathbf{\bar{h}}^\ell), \; \text{and } \; \mathbf{\bar{h}^\ell} = \mathbf{K}^\ell \cdot \mathbf{z}^\ell \\
    \Delta \mathbf{W}^\ell &= \bigg( \Big( \mathbf{B}^\ell \cdot \frac{\partial \mathcal{L}^\ell(\mathbf{y}, \mathbf{\bar{y}}^\ell)}{\partial \mathbf{\bar{h}^\ell}} \Big) \odot \partial\phi^\ell(\mathbf{h}^\ell) \bigg) \cdot \Big( \mathbf{z}^{\ell-1} \Big)^{\mathsf{T}} \label{eqn:local_signaler}
\end{align}
where $\phi_y()$ is a normalized layer-wise local prediction activation function and $\mathbf{B}^\ell \in \mathbb{R}^{\mathbf{J}_\ell \times C}$ and $\mathbf{K}^\ell \in \mathbb{R}^{C \times \mathbf{J}_\ell}$. For classification, a variation of Equation \ref{eqn:local_signaler} entails setting $\phi_y(\mathbf{\bar{h}}^\ell) = \mathbf{\bar{h}}^\ell$ (the identity function) and eschewing $\partial \phi^\ell(\mathbf{h}^\ell)$, resulting in an update rule that is derivative-free. $\mathbf{\bar{h}}^\ell$ contains the pre-transformation score values for the local classifier at $\ell$ while $\mathbf{\bar{y}}^\ell$ is its local prediction of $\mathbf{y}$ (the class label or regression target). %Note that the above framing, even though presented in terms of classification, would in principle work for prediction/estimation tasks such as regression. 
The key is that each layer of neurons is coupled to a local prediction model (with typically fixed random parameters) with a local cost function related to the overall network's central task or possibly a related auxiliary task -- the error associated with each local predictor signals (triggers) and modulates the synaptic adjustment for layer $\ell$.

It was observed in \cite{mostafa2018deep} that the classification performance of each layer of the neural model would improve across parameter updates/learning iterations, even though the local classifiers themselves remained static. Furthermore, the predictions of each layer's random classifier can also be utilized at test-time inference. More importantly, the value of this local learning framework is that the activities and error signals for each layer $\mathbf{z}^\ell$ are immediately available upon the feedforward pass of the MLP, meaning that for any layer weights $\mathbf{W}^\ell$, the update may be readily calculated as soon as the local classifier at $\ell$ has made its prediction -- only the activities $\{\mathbf{z}^0,\mathbf{z}^1,...\mathbf{z}^{\ell-1}\}$ are required/need to exist for layer $\ell$'s update. This has been argued to be more hardware-friendly \cite{mostafa2018deep} and more biologically realistic, notably resolving the update-locking and backwards-locking problem presented in Section \ref{sec:backprop_problems}. Furthermore, this approach to local learning is similar in spirit to the older plasticity process proposed in \cite{ororbia_deep_hybrid_2015b} (specifically the \emph{Bottom-Up} algorithm), which was a scheme that for pseudo-jointly learned a stack of dual-winged harmoniums and usefully did not require any supervisory signals; however, this historical scheme relied on a local form of contrastive divergence (discussed later) to train its per-layer parameters, which faces difficulties related to the mixing of the Markov chain used for sampling. 
%The advantage of this older biological credit assignment algorithm was that it did not require a supervisory signal (it was purely unsupervised unlike the scheme of \cite{mostafa2018deep}); however, this older scheme relied on a local form of contrastive divergence (discussed later) to train its per-layer parameters, which faces difficulties related to the mixing of the Markov chain used for sampling. %and further required its local forward and backward synapses to be symmetric (unlike the scheme of \cite{mostafa2018deep}). 
Beyond its use as a biologically-plausible learning scheme, it is worth pointing out that even backprop-based approaches have been benefited from some use of (greedy) layer-wise local learning \cite{lee_deeply_supervised_2014,rueda2015supervised,belilovsky2019greedy} (though these schemes sometimes ``mix'' the signals produced by the local classifiers with those produced by backprop's global feedback pathway) -- this is sometimes referred to as ``early exiting'' \cite{teerapittayanon2016branchynet,matsubara2022split} or as introducing local classification ``branches'' as in the famous GoogleNet \cite{szegedy2015going}. %that defines backprop to coordinate the system parameter updates.

In general, the ideas underpinning local predictors have gradually become a more explored in the context of conducting credit assignment in deep neural systems  \cite{scholkopf2001generalized,kaiser2020synaptic,belilovsky2020decoupled,bahroun2023duality}. %(\cite{bahroun2023duality} presented a similarity matching variation and made a useful theoretical connection to the representer theorem \cite{scholkopf2001generalized} justification for kernel-based machine learning). 
Other complementary local predictor schemes introduce different local loss functionals \cite{linnainmaa1970representation,werbos1982applications,schmidhuber1990networks,balduzzi2015kickback,guerguiev2017towards,nokland2019training,grinberg2019local,obeid2019structured,teng2020layer}, each with different beneficial properties, while other schemes craft layerwise learning approaches based on local information propagation \cite{wang2021revisiting}. %(or Gaussian linear-gated networks \cite{budden2020gaussian} to a degree). 
Desirably, the local signalers scheme has been demonstrated to work in training spiking neural systems \cite{ma2022deep}.

\noindent
\textbf{Recirculation.} In contrast to the local signaler scheme presented above, recirculation takes on an unsupervised format. Historically, recirculation \cite{hinton1988learning,oreilly_biologically_1996}  was originally proposed as a simpler alternative to training auto-associators (or autoencoders) with backprop. The core procedure works by essentially using two forward passes through the encoder to create signals for learning. The intuition is that the second forward pass (through the encoder) is used to inform the higher of two layers about the effect that it had on the lower-level layer. Training an autoencoder block via recirculation then amounts to: 
\textbf{1)} using the input $\mathbf{z}^{\ell-1}$ to the encoder as the target for the decoder's reconstruction $\mathbf{\bar{z}}^{\ell-1}$, and 
\textbf{2)} using the encoder's output (the initial encoded value from the first forward pass) $\mathbf{z}^\ell$ as the target for the second forward pass through the encoder, or $\mathbf{\bar{z}}^\ell$. While the original algorithm of \cite{hinton1988learning} was crafted for a single hidden-layer autoencoder and contained a brief description at the end of the original paper as to how this might apply to models with multiple hidden layers, we formulate and treat the core principles of recirculation in the context of an arbitrarily deep neural system, much as a far later effort did \cite{tavakoli2021tourbillon}. Concretely, the local statistics needed for computing synaptic updates, as produced via recirculant autoencoding machinery, formulated for any layer of our running example MLP would be as follows:
\begin{align}
    \mathbf{\bar{z}}^{\ell-1} = \beta_d \mathbf{z}^{\ell-1} + (1 - \beta_d) \phi^{\ell-1}_d(\mathbf{V}^\ell \cdot \mathbf{z}^\ell) 
    \; \text{and, } \;
    \mathbf{\bar{z}}^\ell = \phi^\ell(\mathbf{W}^\ell \cdot \mathbf{\bar{z}}^{\ell-1}) \label{eqn:recirc_inf}
\end{align}
where $\mathbf{V}^\ell \in \mathbb{R}^{\mathcal{J}_{\ell-1} \times \mathcal{J}_\ell}$ contains the synaptic weights associated with the local decoder for layer $\ell$. In the above, further observe that another important idea is introduced, through the use of the coefficient $\beta_d$, from \cite{hinton1988learning} -- the notion  
%%%%%%%%%%%%%%%%%%%%%%%%%%%%%%%%%%%%%%%%%%%%
%% recirculation schema
\begin{wrapfigure}{r}{0.545\textwidth}
\vspace{-0.455cm}
  \begin{center}
    \includegraphics[width=0.5425\textwidth]{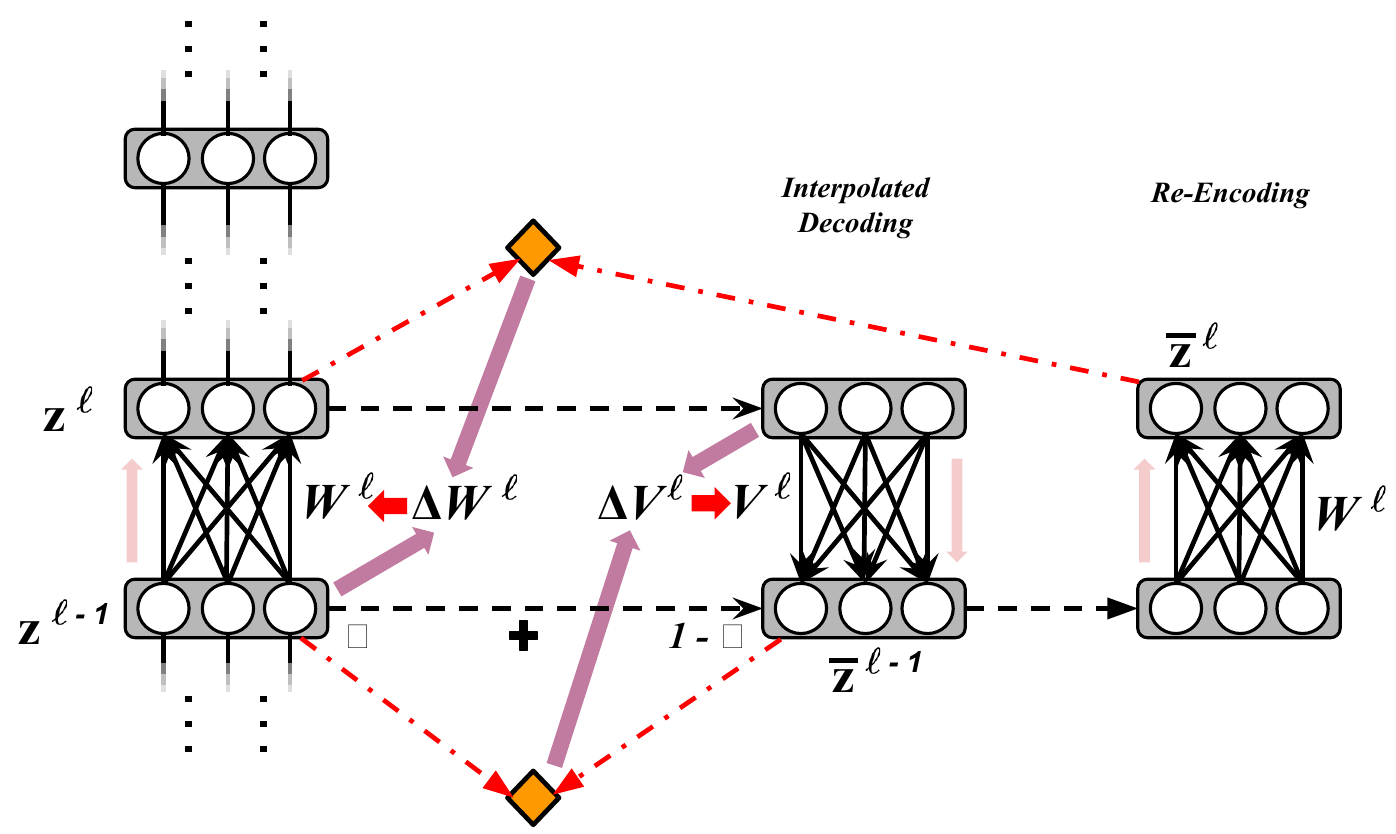}
  \end{center}
  \vspace{-0.255cm}
  \caption{\small{
  The recirculation process used to locally modify the synapses connecting neurons in $\ell-1$ to $\ell$: 1) an interpolated decoding step yields a reconstruction of $\mathbf{z}^{\ell-1}$ which is then fed into the next step, 2) a re-encoding of the interpolated reconstruction of layer $\ell-1$'s activities is performed. The reconstruction and re-encoded values are used, in tandem with the original activities of $\ell-1$ and $\ell$, to produce updates for both $\mathbf{W}^\ell$ and paired decoding synapses $\mathbf{V}^\ell$. Orange diamonds represent error/mismatch. 
  }
  }
  \label{fig:recirculation}
  \vspace{-0.7cm}
\end{wrapfigure}
%%%%%%%%%%%%%%%%%%%%%%%%%%%%%%%%%%%%%%%%%%%%
of ``regression'' or, in other words, interpolation (so as to avoid confusion with the act of regression in statistical learning).  %$\mathbf{\widehat{x}} = \lambda \odot \mathbf{x} + (1 - \lambda)
Specifically, in \cite{hinton1988learning}, it was argued that the coefficient $\beta_d$ ensured that the original input values were not simply replaced by the reconstructed values; this convex combination of the original input to the encoder and the decoder's reconstruction, with a value of $\beta_d$ set to a value closer to one, would ensure that the new input to the encoder (for the second pass) would not be too different from the initial input value. This interpolation further ensures that the upper layer ``measures'' the effect that a small change to the input would have. Once the statistics from Equation \ref{eqn:recirc_inf} are produced, a synaptic adjustment is triggered as follows:
\begin{align}
    \Delta \mathbf{W}^\ell &= -\Big(\mathbf{z}^\ell - \mathbf{\bar{z}}^\ell \Big) \cdot \Big(\mathbf{z}^{\ell-1}\Big)^{\mathsf{T}} \label{eqn:recirc_enc_update} \\ 
    \Delta \mathbf{V}^\ell &= -\Big(\mathbf{z}^{\ell-1} - \mathbf{\bar{z}}^{\ell-1} \Big) \cdot \Big(\mathbf{z}^\ell\Big)^{\mathsf{T}} \label{eqn:recirc_dec_update}
\end{align}
where we notice that the above particular updates, akin to delta rules, do not require the derivative of the activation functions of neither the encoder nor decoder output units. Furthermore, the adjustment equations only use the information produced by recirculated flow of information through the locally coupled encoding and decoding neural modules -- the plasticity that occurs at any layer $\ell$ within the MLP is only concerned with perturbations related to the encoder and decoder pair associated with it. Although recirculation requires a decoder to be paired with the encoder in order to work properly, the learning process is self-aligning, and as a result, unmatched encoder/decoder weights would ultimately tend towards symmetry.

The notions underpinning recirculation have been generalized to construct larger-scale networks \cite{tavakoli2021tourbillon}, even those of residual form, as in \cite{gomez2017reversible}. Other schemes operate much like the local signalers scheme described earlier but in an unsupervised fashion, incorporating principles similar in spirit to those underpinning recirculation, e.g., local feature alignment \cite{zhang2018local}. The generalized recirculation procedure \cite{oreilly_biologically_1996}, or Gene-Rec, which was inspired by the work of \cite{hinton1988learning} (and originally utilized the moniker `recirculation'), operates a bit differently than the recirculation scheme characterized above by Equations \ref{eqn:recirc_inf}, \ref{eqn:recirc_enc_update}, and \ref{eqn:recirc_dec_update} and utilizes a top-down feedback synapses in a relaxation process more akin to contrastive Hebbian learning. This makes Gene-Rec more of a synergistic local explicit signal algorithm and we will treat it as a scheme under the family of energy-based / phased-based schemes in Section \ref{sec:synergistic_local_explicit_algos}.

\subsection{Synergistic Local Explicit Algorithms}
\label{sec:synergistic_local_explicit_algos}

In contrast to non-synergistic local explicit signal schemes, where layers rely on exclusively local machinery to produce signals that drive plasticity, synergistic local credit assignment introduces a degree of (indirect) coordination across the layers in a neural system context, representing a different branch within our taxonomy. In contrast to backprop, which employs a global feedback pathway to transmit error gradients to internal neurons and adjust their relevant synaptic connections, algorithms within this organizational partition leverage either pathways based on functional inversion, error message transmission pathways, or cross-layer feedback synapses that induce a form of message passing based on the combination of both top-down and bottom-up effects/influences of neural activities. 
As such, our taxonomic categorization decomposes synergistic local learning into three emergent sub-families of synaptic adjustment: discrepancy reduction (through message passing), energy-based (through settling phases), and forward-only (through inference alone) mechanics. The first comprises a set of procedures that, at minimum, operate with under two general computational processes -- a target generation process, e.g., using a hierarchical transmission pathway of functional inversion targets, and a subsequent (semi-)local synaptic update process. Alternatively, as we shall see, updates to neural activities and relevant synapses could be produced according to a predictive generative process instead, similar synthetic local updating schemes.
%which could be done either through a hierarchical form of encoding-decoding and noise injection as in target propagation or through the introduction of an additional neuron type, the error unit, which characterizes local representation alignment and its variants that are more further faithful to its origin theory--predictive coding.  Alternatively, the updates to weights and activities themselves could be estimated using synergistic local information instead, as is the case for local gradient prediction. 
The second comprises an even more vast set of algorithms, often starting from the formation of an energy functional that globally relates every neuron to each other and then deriving a set of procedures (which takes into account complex interactions) for creating the target signals that will drive synaptic plasticity through two computation phases of relaxation that travel down this energy function. The final and third family focuses on strictly re-using the information flow offered by the forward inference scheme alone, generally posed in a self-supervised contrastive context.

\subsubsection{Discrepancy Reduction Approaches}
\label{sec:discrep_reduction}

This class of credit assignment, which share some similarities with the previously studied families, is primarily characterized by a form of coordination across neuronal units and layers but not through a single global (synchronizing) signal as in the schemes of Section \ref{sec:global_explicit_algos}. We name this particular family ``discrepancy reduction'' given that the procedures studied here typically work to minimize local errors or difference signals based on target activity vectors produced by a complementary neural process(es) that might span multiple layers, e.g., a message passing synaptic structure or inversion pathway, in service of learning how to better internally represent the neural system's environment. Broadly, these credit assignment schemes typically instantiate computational mechanisms for at least two key steps:
\begin{enumerate}[noitemsep,nolistsep]
	\item A ``search'' for latent representations that better explain the input/output, also known as target representations (this act we refer to as ``local target creation''). This creates the need for local, multi-level objectives that will guide current neural activity representations towards better ones and might take the form of synaptic pathways that propagate information across the entire system in some way. 
    \item Once (suitable) representations are found, updates to parameters are triggered so as to, as much as possible, reduce the mismatch between a model's currently ``formed'' representations and the produced target representations. The sum of the internal, local losses have been likened to the `total discrepancy' in a system, which can be thought of as a sort of approximate free energy functional \cite{millidge2022theoretical,friston2010free}.
\end{enumerate}
%Most of the algorithms under this family do not require post-activation derivatives here either -- or if they do, as some cases in predictive coding or target propagation, they may be dismissed (in predictive coding) or  carefully placed into the right spots of the network to avoid the derivative requirement (target propagation). 
Many of these algorithms offer computations and synaptic adjustments that are effectively local at particular points in time, and, in some cases, such as predictive coding, offer completely unlocked neural systems (no forward, backwards, or updating locking) but at the price of entangling the inference machinery with the learning process. However, many of these approaches, though far more biologically plausible than many of their predecessors, still require training and test time computations to be a bit different; the notable exception is forward-only learning, which uses the exact same information propagation computations in both its inference and plasticity processes.

\noindent
\textbf{Target Propagation.} Target propagation \cite{bengio2014auto,lee2015difference,meulemans2020theoretical,ernoult2022towards} (sometimes abbreviated to `targetprop' or `TP') builds on a premise similar to that of recirculation  -- a learning signal could be created by computing targets through an encoder/decoder pairing. With the general framework of target propagation, this notion is specifically extended to leveraging a backwards pathway through the network that transmits target vector values rather than error gradients (as in backprop); the global feedback pathway is replaced with a global (approximate) `inversion' pathway. 
%This allows for the development of plasticity rules that also do not depend on a global error pathway to carry backwards error derivative information (which would notably side-step the vanishing gradient problem).
In essence, targetprop revolves around the concept of the function inverse; if we had access to the inverse of the network of forward propagations (e.g., an MLP), we could compute which input values at the lower levels of the network might result in better values at the top that would ``please'' the global/output cost. This is much akin to the self-alignment inherent to recirculation: the feedback weights of a decoding module should be trained to (approximately) learn the inverse of the feedforward mapping represented by the encoder. 

In contrast to recirculation, targetprop produces the target values for each layer within a neural network by propagating reconstruction values through the approximate `inversion pathway' of the network. Concretely, under target prop \cite{le1986learning}, this means that for our MLP example, starting with $\mathbf{\widehat{z}}^L = \mathbf{y}$ (or a random encoding of this task-level target), a set of layer-wise target values is produced in the following manner:
\begin{align}
    \mathbf{\widehat{z}}^{\ell-1} = \phi^{\ell-1}_d \Big(\mathbf{V}^\ell \cdot \mathbf{\widehat{z}}^{\ell} \Big) \; \text{for } \; \ell = L,L-1,...,2. \label{eqn:tp_target}
\end{align} 
In the above equation, we see that inversion of the forward flow of information through the network is approximated via a backwards propagation of target signal values through each layer's complementary decoders. The triggered synaptic update for any layer then becomes:
\begin{align}
    \Delta \mathbf{W}^\ell &= \frac{\partial ||\mathbf{z}^\ell - \mathbf{\widehat{z}}^\ell||^2_2}{\partial \mathbf{z}^\ell} \cdot \Big( \mathbf{z}^{\ell-1} \Big)^{\mathsf{T}} \label{eqn:tp_encoder_update}\\ 
    \Delta \mathbf{V}^\ell &= \frac{\partial || (\mathbf{z}^{\ell-1} - \epsilon^\ell) - \mathbf{\bar{z}}^{\ell-1}||^2_2}{\partial \mathbf{\bar{z}}^{\ell-1}} \cdot \Big( \mathbf{\bar{z}}^\ell \Big)^{\mathsf{T}}, \label{eqn:tp_decoder_update} \\ 
    &\text{where } \; \mathbf{\bar{z}}^{\ell-1} = \phi^{\ell-1}_d(\mathbf{V}^\ell \cdot \phi^\ell(\mathbf{W}^\ell \cdot (\mathbf{z}^{\ell-1} + \epsilon^\ell)) ). \nonumber
\end{align}
Note that $\phi^\ell_d()$ is the decoder output activation function (much as in recirculation) while $\epsilon^\ell \sim \mathcal{N}(0,\sigma^\ell)$; note that the zero-mean Gaussian noise can have different values and even be a function of local functionals, as in \cite{ororbia2019biologically} (in its noisy version of difference targetprop, i.e., DTP-$\sigma$).\footnote{Note that noise injection, which can be omitted as in some instantiations of target prop \cite{bartunov2018assessing}, is integrated into inversion/decoder loss (and update) so as to ensure that the encoder, i.e., $\mathbb{R}^{\mathcal{J}_{\ell-1}} \mapsto \mathbb{R}^{\mathcal{J}_\ell}$, and decoder, i.e., $\mathbb{R}^{\mathcal{J}_\ell} \mapsto \mathbb{R}^{\mathcal{J}_{\ell-1}}$, become approximate inverse of one another at the point of not only their respective input vectors ($\mathbf{z}^{\ell-1}$ for the encoder and $\mathbf{z}^{\ell}$ for the decoder) but also the neighborhoods around those points, i.e., a Gaussian ball with radius $\sigma^\ell$). % the resultant targets might prove useful in helping the encoder learn separate information from noise as it tries to disentangle the factors of variation at each layer to help the layers above. 
} 
In difference target propagation (DTP; \cite{lee2015difference,bartunov2018assessing}), the inversion target signals are produced using an alternative to Equation \ref{eqn:tp_target}:
\begin{align}
    \mathbf{\widehat{z}}^{\ell-1} = \phi^{\ell-1}_d (\mathbf{V}^\ell \cdot \mathbf{\widehat{z}}^{\ell} ) + \Big(\mathbf{z}^\ell - \phi^{\ell-1}_d (\mathbf{V}^\ell \cdot \mathbf{z}^{\ell} ) \Big) \; \text{for } \; \ell = L,L-1,...,2,  \label{eqn:dtp_target}
\end{align}
where we see that DTP includes a second term that serves as a correction factor that compensates for the imperfectness of the decoding process (which can obstruct learning) of the original targetprop \cite{lee2015difference}; this term measures the reconstruction error (between a local decoding from $\mathbf{z}^\ell$ to $\mathbf{z}^{\ell-1}$ and the original activity values of $\mathbf{z}^\ell$) which provides a linear correction to account for the use of learned, imprecise inversion functions. In any variant of targetprop, the process of credit assignment simply entails using the global inversion pathway -- the chain of approximate inversions to produce target values -- and then updating each layer to move closer to its target value. Like recirculation, targetprop also works with non-differentiable nonlinear activation functions, including those that are discrete-valued (so long as it is possible to approximate its inverse). Under simple conditions, when all the layer objectives are combined, targetprop can yield updates with the same sign as the updates obtained by backprop \cite{le1986learning,ahmad2020gait} and, under other theoretical conditions -- assuming that underlying function to be approximated is Lipschitz continuous and that the difference between backward and forward activities is less than some small value $\epsilon$ -- the resultant learned ANN can be shown to converge to an optimal point (and loosely approximate a predictive coding scheme \cite{salvatori2023brain}). % (and, furthermore, that target prop can be seen as an approximation of predictive coding, a credit assignment we review later). 
Various modern incarnations of targetprop exist, such as those that choose to keep the inversion/decoding synapses random and fixed \cite{shibuya2023fixed}, as well as algorithms that embody its central principles \cite{podlaski2020biological} such as function inversion.
%Through a layerwise training approach, target propagation tries to ensure that each layer is individually a useful contractive autoencoder

% spike nets and rnns
Targetprop has also been applied to training recurrent networks that process sequential data points \cite{wiseman2017training,manchev2020target,mali2021investigating,roulet2021target}, e.g., natural language modeling, but still requires unrolling the computation graph along the length of the sequence. Furthermore, some connections have been made between targetprop and spike-time-dependent plasticity \cite{andrew2003spiking}, although targetprop has not, at the time of this article's writing, been formulated for spiking neural networks. Empirically, however, it has been demonstrated that, at times, targetprop can result in an unstable learning process \cite{bartunov2018assessing,ororbia2019biologically} (unless an intelligent noise scheme is used when the underlying network is deep) and, furthermore, its performance is not quite comparable to that of backprop's. A theoretical study attempted to provide an explanation of this gap by establishing an equivalency between targetprop and Gaussian-Newton optimization \cite{meulemans20}, and further proposed an algorithmic reformulation that can be viewed as a hybrid between Gauss-Newton optimization and gradient descent. The biological plausibility and generalization performance of this reformulation was improved through the introduction of a prior % over targets (in terms of a loss function) 
that nudged the Jacobian of the feedback/inversion module towards that of the corresponding feedforward parameters \cite{ernoult2022towards}. 

%%%%%%%%%%%%%%%%%%%%%%%%%%%%%%%%%%%%%%%
\begin{figure}[!t]
\centering     %%% not \center
\begin{subfigure}{0.495\textwidth}
    \centering
    %\vspace{-5mm}
     \includegraphics[width=0.875\linewidth]{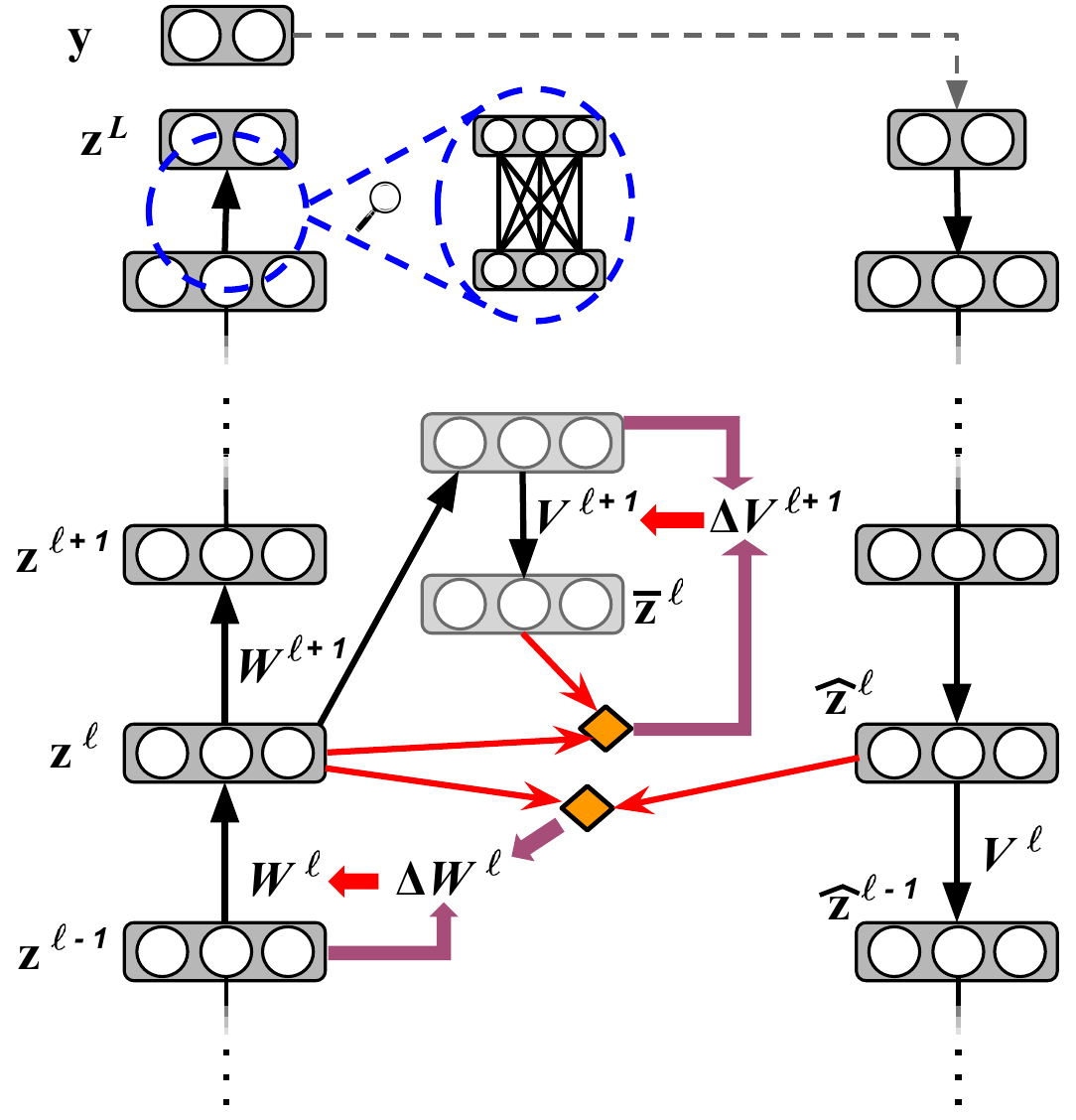} 
    % \caption{Target propagation (noise-free).}
    % \label{fig:tprop}
\end{subfigure}
\begin{subfigure}{0.495\textwidth}
    \centering
    %\vspace{-5mm}
     \includegraphics[width=0.7\linewidth]{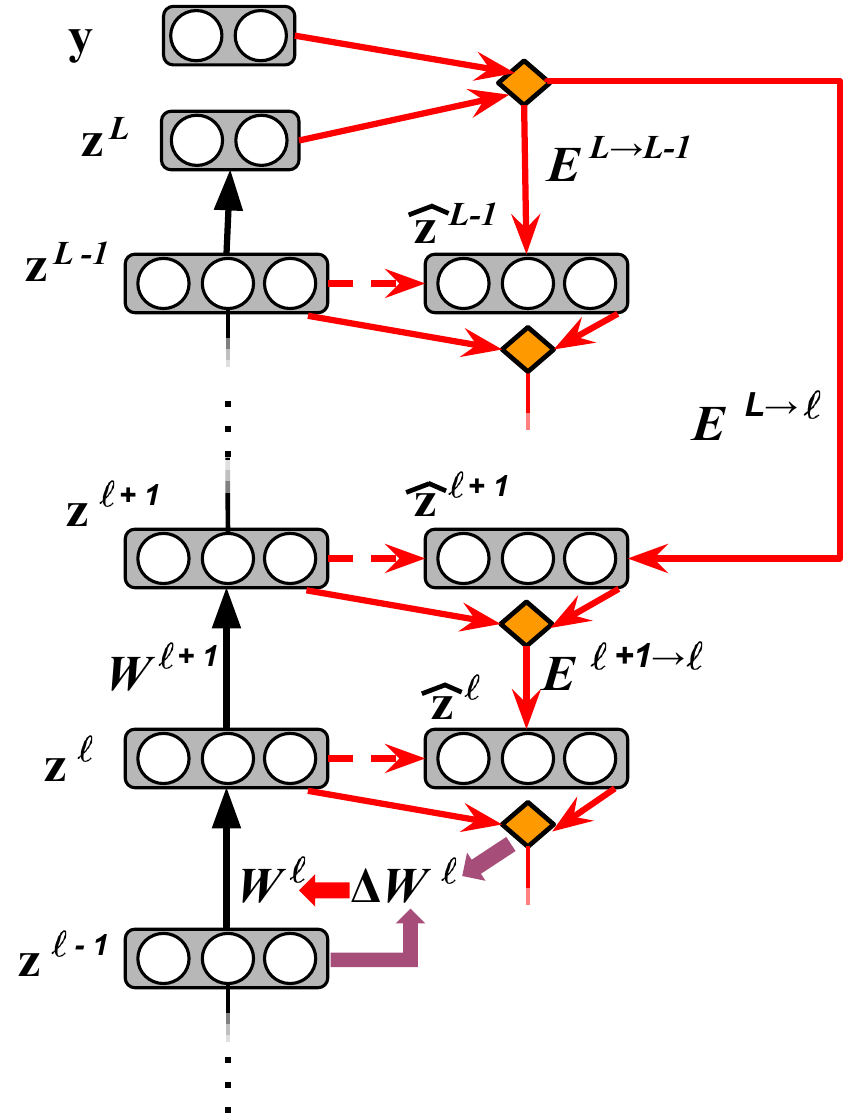} 
    % \caption{Recursive local representation alignment.}
    % \label{fig:lra}
\end{subfigure}
%\vspace{-0.3cm}
\caption{\small{\textbf{Credit Assignment via Discrepancy Reduction.} Orange diamonds represent error/mismatch. 
(Left) A target propagation scheme (with no noise injected) producing a target $\mathbf{\widehat{z}}^\ell$ that triggers a synaptic update for $\mathbf{W}^\ell$ (also shown is adjusting for the corresponding inversion synapses $\mathbf{V}^{\ell+1}$). (Right) A recursive local representation alignment scheme producing target $\mathbf{\widehat{z}}^\ell$ so as to trigger a synaptic update for $\mathbf{W}^\ell$.
}}
\label{fig:prop_circuits}
\vspace{-0.6cm}
\end{figure}
%%%%%%%%%%%%%%%%%%%%%%%%%%%%%%%%%%%%%%%

\noindent
\textbf{Local Representation Alignment.} %% a fast, noisy approximation of predictive coding
Credit assignment within the framework of local representation alignment (LRA) \cite{ororbia2019biologically} can viewed as a blending of principles underlying targetprop, recirculation, and predictive processing theory \cite{rao1999predictive,clark2015surfing,salvatori2023brain}. However, as opposed to formulating each layer of neurons as a pairing of an encoder and decoder in the context of functional inversion, LRA crafts a form of synergistic learning that seeks to minimize a (pseudo-)free energy functional \cite{ororbia2017learning,ororbia2019biologically} -- effectively a form of what is known in predictive coding as the variational free energy \cite{friston2010free} -- which facilitates the matching or alignment of a neural model's current means of representing input-output pairings, e.g., a sensory input $\mathbf{x}$ and its associated context vector $\mathbf{y}$, to a corresponding set of desired `target representations'. Much as targetprop schemes \cite{miguel2014nested,bengio2014auto,lee2015difference} do, LRA posits that these target values must be created by a complementary process (in the form of a neural structure), but instead of constructing a backwards chain of approximate decoders/inversions, LRA entails constructing a synaptic message passing structure, where mismatch values (or local `discrepancies') are communicated along error connections. Furthermore, LRA-based schemes \cite{ororbia2018conducting,ororbia2019biologically,ororbia2023backpropagation,flugel2023feed,kappel2023block} (including variants such as feedback control \cite{meulemans2021credit,meulemans2022minimizing}) focus on decomposing the larger credit assignment problem into smaller, easier-to-solve sub-problems; viewing a full neural architecture's underlying computational graph as a composition of connected ``computational subgraphs'' (the boundaries of each subgraph are defined as the set of all pre-synaptic neural units/variables and all post-synaptic neural units/outputs). Ultimately, learning under LRA will first require that, for any given input/output pattern pairing, target representations are computed and then, given the acquired targets, synaptic parameters are adjusted to better recall these target representations for similar sensory input/output pairings. 

Concretely, each neuronal layer in LRA will have a target (activity) vector associated with it such that adjusting the associated synaptic weights will help move that layer's activity values towards the better, matched target values in the future. LRA ensures that useful changes in plasticity are produced within LRA by effectively choosing targets that are ``within'' the possible representation of a network's associated layers, i.e. layers are not forced to try to match a target that is impossible to achieve \cite{ororbia2018conducting} (for exampling, the layer's activation function imposes bounds on what values it outputs). Formally, under LRA, we define the mismatch signals at any layer $\ell$ to be computed as the first derivative of a local distance function $\mathbf{e}^\ell = \frac{\partial ||\mathbf{z}^\ell - \mathbf{\widehat{z}}^\ell||^q_p}{\partial \mathbf{z}^\ell}$, e.g., if $p=2$ then $\mathbf{e}^\ell = -2(\mathbf{z}^\ell - \mathbf{\widehat{z}}^\ell)$, and the target representation to be produced for any layer $\mathbf{\widehat{z}}^\ell$ is done according to the following:
\begin{align}
    \mathbf{\widehat{z}}^\ell &= \phi^\ell\Big( \mathbf{h}^\ell - \beta \Big( \sum_{b \in \mathcal{B}_\ell} \mathbf{d}^\ell_b \Big) \Big), \text{where } \mathbf{d}^\ell_b = \mathbf{E}^{b \rightarrow \ell} \cdot \mathbf{e}^b \label{eqn:lra_target}
\end{align}
where $\beta$ is a the pre-transformation perturbation coefficient and $\mathcal{B}_\ell$ is the set of integer indices where each index corresponds to the $b$th layer ($b \neq \ell$) in the neural system that will pass/transmit its own mismatch message/signal to layer $\ell$. Note that $\mathbf{E}^{b \rightarrow \ell}$ indicates a synaptic error message passing pathway between layers $b$ and $\ell$. 
In effect, according to Equation \ref{eqn:lra_target}, an adjustment, i.e., the accumulation of all transmitted error messages of layers indexed in set $\mathcal{B}_\ell$, is applied to the (pre-transformed) neural activity itself to produce a target representation vector $\mathbf{\widehat{z}}^\ell$ (in effect, this could be viewed as a single step of short-term plasticity). Once a target is produced for layer $\ell$, synaptic plasticity proceeds according to the following local rules:
\begin{align}
    \Delta \mathbf{W}^\ell = \mathbf{e}^\ell \cdot (\mathbf{z}^{\ell-1})^{\mathsf{T}}, \; \text{and, } \; \Delta \mathbf{E}^{b \rightarrow \ell} = \gamma \Big( \mathbf{e}^\ell \cdot (\mathbf{z}^{b})^{\mathsf{T}} \Big)
\end{align}
where the above prescribes an adjustment to both the connections between layers $\ell-1$ and $\ell$ but also for any message passing pathway $\mathbf{E}^{b \rightarrow \ell}$; $0 \leq \gamma < 1$ is a temporal scaling constant that slows down the evolution of message passing synapses and has been shown to ensure stability in a neural system's learning \cite{ororbia2019biologically,ororbia2023backpropagation}. Note that the error synapses could also be set to fixed random values as was done in LRA's earlier incarnations \cite{ororbia2018conducting}).

% rec-LRA breaks free of backwards/update locking problems 
Note that the notion that every single layer must be associated with a target can be relaxed, as was shown in the scheme known as recursive-LRA (rec-LRA; \cite{ororbia2023backpropagation}), which circumvents the problems of backwards and update locking. This stems from the very fact that LRA schemes do \emph{not} require the message passing structure to be designed to mirror the information flow through a neural model's various layers, i.e., its inference scheme; this permits ``skip error feedback connections'' or (error) synaptic transmission pathways that skip across layers (similar in spirit to, but more general than, the wiring patterns of direct feedback alignment \cite{nokland2016direct}). Underwriting the setup of rec-LRA is the idea that subgraphs -- portions of a neural architecture such as the stack of operators (represented by hierarchically arranged neurons) that characterizes a `residual block' in a deep residual network -- can be paired with one target, which has also been referred to as an intermediate or ``supporting representation'' \cite{ororbia2023backpropagation}. Once the message passing structure produces a target representation vector for a given subgraph, the synaptic updates made to the internal layers of that subgraph can be made with a small skip-layer transmission pathway -- in other words, a recursive application of rec-LRA to the block itself -- or. alternatively. with direct feedback alignment or Hebbian plasticity rules, independently of and in parallel to the other subgraphs that compose the entire neural system.

% cite timothy work + rec-LRA as means of introducing conv/pconv
% general content
Notably, under some mild assumptions, it can also be shown that LRA approximates backprop \cite{ororbia2018conducting} as well as predictive coding \cite{salvatori2023brain}. LRA-based approaches have been mainly tested on classification tasks focused on the domain of computer vision and have been shown to generalize well on natural images \cite{zee2022robust}, e.g., CIFAR-10 or SVHN, as well as large-scale (image) benchmarks such as ImageNet \cite{ororbia2023backpropagation}. Furthermore, evidence has been found showing that LRA is robust to poor choices of initialization and can train deep networks (including very thin and deep ones) from initializations that would cause backprop, as well as some other biologically-inspired algorithms, to fail \cite{ororbia2018conducting,ororbia2019biologically}. Furthermore, LRA, like targetprop, does not require differentiable neural activities and has been shown to robustly handle stochastic and discrete-valued functions \cite{ororbia2018conducting,ororbia2019biologically,ororbia2023backpropagation,zee2022robust}. In addition, elements of LRA have been been integrated into temporal learning architectures that process sequential data online \cite{ororbia2018continual} and interesting generalizations of it entail treating the synergy afforded by feedback as a control problem \cite{meulemans2021credit,meulemans2022minimizing}.
%which connects directly to the next set of procedures referred to as neural predictive coding algorithms. % a transition to next sub-section PC

\noindent
\textbf{Predictive Coding.} While some neurobiological algorithms and schemes borrow ideas from predictive coding/processing \cite{clark2013whatever,marino2022predictive}, such as the aforementioned LRA or targetprop, a particularly noteworthy member of the discrepancy reduction family in our credit assignment taxonomy is predictive coding itself, or rather, computational (sparse) predictive coding \cite{rao1999predictive,olshausen1997sparse,rozell2008sparse,friston2005theory,bastos2012canonical,panichello2013predictive,chalasani2013deep,spratling2017hierarchical,ororbia2022ngc,salvatori2022reverse}. Crucially, predictive coding centers around the fundamental premise that the brain is a probabilistic generative model; it is a stochastic engine that continuously makes predictions about its environment and updates its internal hypotheses (that create those predictions) based on the sensory evidence it gathers. Theoretically, predictive coding entails a commitment to the Bayesian brain or predictive processing hypothesis \cite{lee2003hierarchical,doya2007bayesian,clark2013whatever}. Mechanistically, a core distinguishing element of predictive coding is that of the \emph{error neuron} -- a specific neuronal cell (modeling aspects of superficial pyramidal cells) whose dynamics embody local mismatch signals -- and the \emph{state neuron} -- a stateful neuronal cell whose dynamics (that model aspects of deep excitatory pyramidal cells) capture statistical regularities and, depending on their design, temporal dependencies inherent to sensory streams. Error neurons effectively encode the difference between the activity at a given layer of the network and that predicted/generated by a higher-level; these mismatch signals are then propagated throughout the network \cite{rao1999predictive}. In neurophysiological studies, evidence of prediction errors in neural activity during perceptual decision tasks has been found \cite{summerfield2006predictive,summerfield2008neural}. Furthermore, neurons in the primary visual cortex have been reported to respond to mismatches between predicted and actual visual input \cite{zmarz2016mismatch,fiser2016experience}.
%implying that synapses evolve at a slower time-scale than neural activities.

Formally, predictive coding (PC) starts from the perspective that any one of the brain's underlying neural circuits (a network of neuronal units) can be cast a generative model and expressed in terms of a quantity that it optimizes at any single step in time -- a functional known as the variational free energy (VFE; \cite{friston2010free}). VFE effectively states that two key terms must be balanced: one that encourages improvement in the accuracy the generative component and another that penalizes model complexity (which aligning with the notion of Occam's razor) by encouraging the underlying model to ensure that the Kullback-Leibler (KL) divergence \cite{kullback1997information} between its recognition model and its prior is as small as possible. Formally, VFE can be stated in the following manner:
\begin{equation}
    E(p,q,\mathbf{x}) = \underbrace{\sum_{\mathbf{z}}q(\mathbf{z}) \text{log}\bigg( \frac{q(\mathbf{z})}{p(\mathbf{z})} \bigg)}_{\text{Complexity Term}} \ + \ \underbrace{\sum_{\mathbf{z}}q(\mathbf{z})\text{log}\bigg(\frac{1}{p(\mathbf{x} \mid \mathbf{z})}\bigg)}_{\text{Accuracy Term}}  \label{eqn:vfe}
\end{equation}
where $p(\mathbf{x}|\mathbf{z})$ is the underlying directed generative model (likelihood) that a predictive coding circuit embodies while $p(\mathbf{z})$ is the prior over its (latent) variables, cast in terms of neural activity values. $q(\mathbf{s})$ is an approximate posterior distribution over the latent activities $\mathbf{z}$ (given an observation); the choice of this posterior often depends on the structure of the generative model and the parameters to be optimized. The marginal probability of a predictive coding neural circuit exhibits the following dependencies:\footnote{Note that, in what follows, we have, for consistency within this survey, presented predictive coding in a ``bottom-up'' depiction whereas, in the literature, this kind of model is presented in a ``top-down'' fashion, i.e., the generative model is typically formulated as $p(\mathbf{z}^0,...\mathbf{z}^L) = p(\mathbf{z}^L \prod^{L-1}_{\ell=0} p(\mathbf{z}^\ell | \mathbf{z}^{\ell+1})$; see \cite{salvatori2023brain} for further details.}
\begin{align}
    p(\mathbf{z}^0, \dots, \mathbf{z}^L) = p(\mathbf{z}^0) \prod_{\ell=1}^{L} p(\mathbf{z}^\ell \mid \mathbf{z}^{\ell-1})
\end{align}
and, if each intermediate distribution is chosen to be multivariate Gaussian, i.e., $p(\mathbf{z}^\ell \mid \mathbf{z}^{\ell-1})$ with mean produced by {$\mathbf{\bar{z}}^\ell = \mathbf{W}^\ell(t) \cdot \phi^{\ell-1}\big( \mathbf{z}^{\ell-1}(t) \big)$}, a transformation of the latent neural states in the layer above, we obtain:% the following:
\begin{align}
    p(\mathbf{z}^0) = \mathcal N(\mathbf{z}^0; \mathbf{\bar{z}}^0, \mathbf{\Sigma}^0), \; \text{and, } \; p(\mathbf{z}^\ell \mid \mathbf{z}^{\ell-1}) = \mathcal N(\mathbf{z}^\ell; \mathbf{\bar{z}}^\ell, \Sigma^\ell), 
\end{align}
meaning that we treat a neural circuit as a hierarchical latent variable Gaussian generative model. To optimize this generative neural circuit, we can write down a concrete form of the VFE in Equation \ref{eqn:vfe}, after assuming a mean-field approximation that enforces independencies among the neurons within the circuit (see \cite{friston2010free,ororbia2019lifelong,ororbia2022ngc,salvatori2023brain} for details), in the following manner:
\begin{align}
\mathcal{F}(\Theta) = \sum^L_{\ell=0} \frac{1}{2\Sigma^\ell} \sum^{\mathcal{J}_\ell}_{i=1} \big( \mathbf{z}^\ell_i(t) -  \mathbf{\bar{z}}^\ell_i \big)^2\label{eqn:pc_vfe}
\end{align}
where $\Theta = \{\mathbf{W}^\ell(t),\mathbf{E}^\ell(t)\}^L_{\ell=1}$, the construct that houses all of the synapses that make up the predictive coding circuit; $\mathbf{W}^\ell(t)$ are the generative/predictive synapses while $\mathbf{E}^\ell(t)$ are the message passing synapses. The functional of Equation  \ref{eqn:pc_vfe} highlights two key neurobiological commitments that predictive coding models make: 
\textbf{1)} there exists a kind of neuron which specializes in calculating mismatch values -- this means that each layer of the circuit contains a set of precision-weighted \emph{error neurons} as in: ${\mathbf{e}^\ell(t) = \frac{1}{\Sigma^\ell} \big( \mathbf{z}^\ell(t) -  \mathbf{\bar{z}}^\ell} \big)$\footnote{This treatment of predictive coding assumes a fixed scalar precision $\frac{1}{\Sigma^\ell}$. However, this need not be the case and $\Sigma^\ell$ could instead take the form of a learnable synaptic (covariance) matrix, which is done in some variants of predictive coding \cite{ororbia2022ngc,salvatori2023brain}.}, and 
\textbf{2)} any state neuron $i$ in one layer of the circuit (specifically inside of the vector $\mathbf{z}^{\ell+1}(t)$) is actively attempting to guess the activity value of another neuron $j$ in layer $\ell$. 
Furthermore, in PC, the activities of any layer of (state) neurons are modeled in terms of time-evolving dynamics, formally through variations of the following ordinary differential equation:
\begin{align}
    \tau_m\frac{\partial \mathbf{z}^\ell(t)}{\partial t} = -\gamma \mathbf{z}^\ell(t) + \mathbf{d}^\ell \odot \partial \phi^\ell(\mathbf{z}^\ell(t))  -\mathbf{e}^\ell(t)  \label{eqn:pc_state_update}
\end{align}
%}%
where $\tau_m$ is the cellular membrane time constant and $\mathbf{d}^\ell = \mathbf{E}^{\ell+1}(t) \cdot \mathbf{e}^{\ell+1}(t)$, i.e., the perturbations produced by error messages that are passed back along feedback synapses $\mathbf{E}^\ell(t)$. Typically, predictive coding circuitry is simulated over a window of time of length $T$ (the stimulus presentation time), in a dynamic expectation-maximization fashion \cite{dempster1977maximum,friston2008variational,friston2010generalised}, where Equation \ref{eqn:pc_state_update} is repeatedly run, via Euler integration, for $T$ discrete steps to obtain activity values for $\mathbf{z}^\ell(t)$ (this equation is applied to all layers in parallel). After values for each layer are obtained, a synaptic update is triggered (for both the generative/predictive synapses and the message passing ones), such as in the following two-factor Hebbian rules:
\begin{align}
    \tau_w \frac{\partial \mathbf{W}^\ell(t)}{\partial t} &= -\gamma_w \mathbf{W}^\ell(t) + \mathbf{e}^{\ell}(t) \cdot \Big(\mathbf{z}^{\ell-1}(t)\Big)^T \label{eqn:pc_gen_synaptic_update} \\
    \tau_e \frac{\partial \mathbf{E}^\ell(t)}{\partial t} &= -\gamma_e \mathbf{E}^\ell(t) + \mathbf{z}^{\ell-1}(t) \cdot \Big(\mathbf{e}^{\ell}(t)\Big)^T\label{eqn:pc_err_synaptic_update}
\end{align}
where $\tau_w$ and $\tau_e$ are generative and error feedback synaptic plasticity time constants, respectively, while $\gamma_w$ and $\gamma_e$ are the decay constants for the generative and error feedback synaptic updates, respectively. In essence, the above equations indicate that plasticity for synapses is based on error neuron activity and pre-synaptic state unit values. Note that, while we present a generalized form of predictive coding using separate forward and backward connections, many implementations of predictive coding simplify the process by setting $\mathbf{E}^\ell(t) = (\mathbf{W}^\ell(t))^T$, removing the need for Equation \ref{eqn:pc_err_synaptic_update} entirely.

%%%%%%%%%%%%%%%%%%%%%%%%%%%%%%%%%%%%%%%%%%%%
%% predictive coding schema
\begin{wrapfigure}{r}{0.4\textwidth}
\vspace{-0.455cm}
  \begin{center}
    \includegraphics[width=0.225\textwidth]{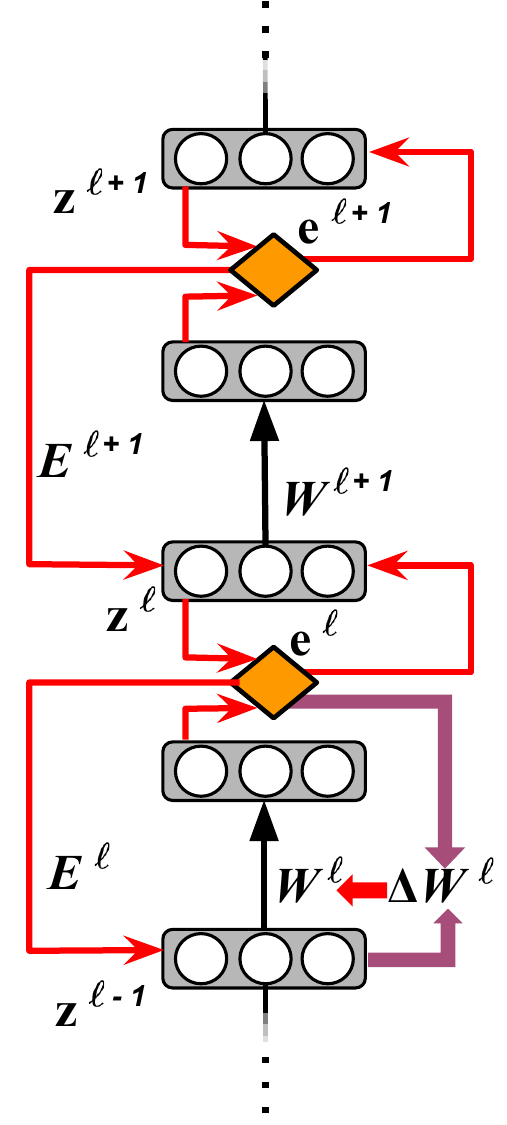}
  \end{center}
  \vspace{-0.255cm}
  \caption{\small{
  A predictive coding (sub-)circuit producing a synaptic update for $\mathbf{W}^\ell$; this change in plasticity is a function of local information, i.e., (post-synaptic) error neuron activities at $\ell$ and (pre-synaptic) state neuron activities at $\ell-1$.
  }}
  \label{fig:pred_coding}
  \vspace{-0.5cm}
\end{wrapfigure}
%%%%%%%%%%%%%%%%%%%%%%%%%%%%%%%%%%%%%%%%%%%%

PC circuitry can be used to construct a wide plethora of models, including unsupervised generative models \cite{ororbia2022ngc} and graph networks \cite{salvatori2022reverse}. To construct a supervised prediction model, like our example MLP, all that one needs to do is clamp the bottom layer of the network to an input observation $\mathbf{z}^0(t) = \mathbf{x}$ and the top layer to a target context value $\mathbf{z}^L(t) = \mathbf{y}$, initialize the values for the hidden layers by next running a feedforward pass through the network\footnote{This greatly speeds up model optimization convergence and reduces the number of steps $T$ taken for the short-term plasticity equations to be run \cite{ororbia2022cogngen}.}, and then run Equation \ref{eqn:pc_state_update} for $T$ steps\footnote{An alternative is to run Equation \ref{eqn:pc_state_update} until the neural circuit has converged to a fixed point/steady state \cite{whittington2017approximation}.} before finally calculating synaptic updates using Equations \ref{eqn:pc_gen_synaptic_update} and \ref{eqn:pc_err_synaptic_update}. Notice that this scheme can be viewed as treating the updates to state neurons as a form of short-term plasticity and the synaptic updates as a form of long-term plasticity (potentiation and depression). 

Recent efforts in machine intelligence have provided encouraging results for a wide range of architectures based on predictive coding credit assignment \cite{chalasani2013deep,santana2017exploiting,ororbia2017learning,whittington2017approximation,ororbia2018continual}, including those that have demonstrated its value on time-varying, sequential data streams \cite{ororbia2017learning,ororbia2018continual,jiang2022dynamic}, continual learning \cite{ororbia2019lifelong,yoo2022bayespcn}, reinforcement learning control \cite{ororbia2022active,gklezakos2022active,rao2022active}, and large language modeling \cite{pinchetti2022} (see \cite{salvatori2023brain} for a more thorough survey of machine learning applications of predictive coding). Furthermore, predictive coding has been generalized, in various ways, to operate at the level of spiking neuronal dynamics \cite{zhang2004single,rao2004hierarchical,ororbia2019spiking,n2023predictive}. Other complementary work has proposed augmenting the typical state-of-the-art feedforward DNNs \cite{wen2018deep} with predictive coding elements/features
(a ``side predictive network'' that can be treated as a process that must settle to good representations that feed to a final linear classifier). Others have drawn inspiration from the concept of the error unit in predictive coding and incorporated it into backprop-based frameworks for video processing \cite{lotter2016deep}. % Could better merge these two sentences since they kind of are doing the same thing for different types of problems...

The synergistic local credit assignment conducted under the framework of predictive coding satisfies many criteria for biological plausibility and is further grounded in a principled Bayesian framing of neuronal inference and learning \cite{friston2008hierarchical,friston2017graphical}, %(in certain cases, it can even be shown to approximate backprop itself \cite{whittington2017approximation}) 
further offering reasonably efficient computational processing mechanisms. The update rules are simple and local -- as presented above, they reduce to two-factor Hebbian-like adjustments \cite{rao1999predictive,whittington2017approximation,ororbia2018continual,ororbia2022ngc} -- and are triggered by the neural machinery of the system's inference, particularly its message passing scheme as constructed via error transmission pathways. Crucially, a predictive coding circuit is completely unlocked -- it resolves the forward, backwards, and update locking problems inherent to backprop. 
Furthermore, in the formulation shown above, the weight transport problem is additionally resolved (if using separate error transmission synapses \cite{ororbia2022ngc} as opposed to tied forward/backward weights as in \cite{rao1999predictive}). Furthermore, it has even been theoretically shown that, under certain conditions, predictive coding can approximate, or rather converge to, backprop \cite{whittington2017approximation,salvatori2022reverse}, specifically in the limit where the activity of the error neurons approaches zero; earlier evidence of PC-like neural model approximately emulating backprop was provided in \cite{kording2001supervised}. %Unlike backprop, predictive coding does not require a specific kind of neural structure, even capable of performing bi-directional mapping allowing for the specification of higher-level variables (possibly that akin to an independently controllable factor like that of \cite{thomas2018disentangling}) that can successfully correspond to inputs of different modalities. 
PC structures, including models such as \cite{guerguiev2017towards,sacramento2018dendritic}, although far more biologically-plausible than backprop-trained ANNs, are still not without issue -- it is argued that their typical formulation of one-to-one pairing between states neurons and error neurons is not biophysically realistic   \cite{bogacz2017tutorial,whittington2017approximation} and that error neurons cannot take on both positive and negative values as they do in the dynamics presented above, i.e., biological neurons cannot have negative activity values though it has been remarked that this could be corrected by using the more biologically plausible linear rectifier \cite{bogacz2016properties}. In general, for more focused, targeted reviews of PC that also cover its wide plethora of variants as well as task applications, we refer the reader to \cite{spratling2017review,millidge21review,salvatori2023brain}.

\subsubsection{Energy-Based / Phase-Based Approaches}
\label{sec:energy}

This credit assignment family shares many similarities with discrepancy reduction; notably that it is possible to write down a global energy functional that is being optimized by the neural system. However, a defining feature of the procedures reviewed here is that the signals required to trigger changes in synaptic values come from multiple phases of settling or relaxation; these require processes that require multiple steps of iterative computation to the dynamics of each neuronal layer. %The procedures also generate targets to guide the activities of the hidden layers to representations that better explain the data, but they do so through relaxation phases, or processes that require multiple iterations of computation, steps derived from the global energy function to create destination representations. 
In short, the underlying neural model must ``settle'' to an equilibrium point (if deterministic) or stationary point (if stochastic) to obtain both its predictions and the targets needed for parameter adaptation/adjustment -- a key advantage, as we will observe, is that effectively the same neural machinery used in inference directly plays a role in changes in synaptic plasticity.

The main motivation behind this family is that neurobiological learning/memory can be cast as manipulating points in an energy functional's landscape \cite{lecun2006tutorial}. %-- sensory evidence (data) will correspond to points that correspond to lower energy values while patterns not collected as evidence will map to higher energy values. 
This means that the hidden units of the network will gradually, as a result of adjustments made to synapses, move towards configurations that are more probable given the sensory input and the agent's internal model of the world (as represented by its parameters). %One common unifying theme of models that fall under the phase-based computing family are that they are all explicit energy-based models, i.e., they are directly derived from a specified energy function \cite{lecun2006tutorial}. 
Note that phase-based computation typically applies to a type of neural model called an `interactive activation network' \cite{mcclelland1981interactive}, a well-known instantiation being the continuous Hopfield model \cite{hopfield1984neurons} with modern incarnations \cite{ramsauer2020hopfield}. In contrast to networks such as our MLP example, interactive networks offer several interesting benefits: 
\textbf{1)} they naturally exhibit a pattern completion capability (i.e., imputation of missing features), 
\textbf{2)} their design offers a flexible treatment of units as either inputs or outputs and have a useful connection to an associated energy function that is useful for pattern recognition in general \cite{williams1991mean,lecun2006tutorial}, 
\textbf{3)} they can be used to solve soft-constraint problems (making them suitable for modeling various cognitive processes \cite{mcclelland1981interactive}), and 
\text{4)} they often do not require post-activation derivatives (unless the iterative dynamics of the neurons have been particularly specified to require them). Central drawbacks for phase-based models are that, as we will see, they generally require strict symmetry of forward and feedback weights and, more significantly, are slow to train given the lengthy relaxation phases needed to find a system's fixed-points.

\textit{Why is an energy functional useful?} The main idea is to create a set of attractor states in a nonlinear network and, as pointed out by  \cite{hopfield1983unlearning,hopfield1984neurons}, if the feedforward and feedback connections are symmetric, the network will settle down into states that are the local minima of a simple (energy) functional. New minima would be produce by Hebbian plasticity, yielding dynamics that would allow a network (with fixed weights) to conduct memory retrieval from incomplete or corrupted patterns/memories. Anti-Hebbian learning could then be used to store new memories. The energy functional, in a Hopfield network, could be used to determine the ``fast'' dynamics of neuronal activities while a different objective, one that measured the proximity of energy minima to pattern vectors that needed to be stored, could be used to determine the ``slow'' dynamics of the synaptic strengths \cite{hinton2003ups}. In addition to characterizing such a system's generalization ability \cite{hinton1986learning}, the (stochastic) dynamics of neurons could be justified as performing Bayesian inference, providing mathematical justification of simple synaptic adjustment rules; we will observe this in credit assignment schemes such as contrastive Hebbian learning and contrastive divergence. A consequence of this perspective is that the algorithms of this section must use one phase to sample interpretations of a sensory vector, in proportion to their probabilities, to lower the energies for the sampled interpretations (by increasing synaptic weight values) while another phase is used to raise the energies of the alternative interpretations. % usually these are slow

\noindent
\textbf{Contrastive Hebbian Learning. } Contrastive Hebbian learning \cite{hopfield1983unlearning,movellan1991contrastive,galland1991deterministic,baldi1991contrastive,seung2017correlation,kermiche2019contrastive,detorakis2019contrastive} (CHL), %, of which we will also refer to as contrastive Hebbian (CH) plasticitiy,
with modern variations \cite{qin2021contrastive,williams2023flexible}, is a generalization of Hebbian plasticity that ultimately results in a subtractive measurement between the cross-products of activation values collected from a clamped phase and those collected from a free running phase. CHL notably underpins the foundations of Boltzmann learning \cite{hinton1989deterministic} as well as the general adaptation of mean-field networks \cite{anderson1987mean}. Though CHL was originally proposed for a specific class of interactive networks, later work demonstrated CHL could be applied to any case of what is now known as the continuous Hopfield model \cite{hinton1989deterministic,movellan1991contrastive,vsima2003continuous}; recent incarnations are sometimes referred to as `modern' or `universal' Hopfield networks \cite{ramsauer2020hopfield,seidl2022improving,millidge2022universal}. These early results are important in that they allow us to treat even our running example MLP as an interactive system, provided that we treat its forward connections also as symmetric feedback connections (any arbitrary network can be created from the fully-connected Hopfield by simply zeroing out certain connections, an important point that was revived in equilibrium propagation \cite{scellier2017equilibrium}; reviewed later in this section).

%%%%%%%%%%%%%%%%%%%%%%%%%%%%%%%%%%%%%%%%%%%%%%%%%%%%%%%%%%%
\begin{wrapfigure}{r}{0.515\textwidth}
\vspace{-0.45cm}
  \begin{center}
    \includegraphics[width=0.51\textwidth]{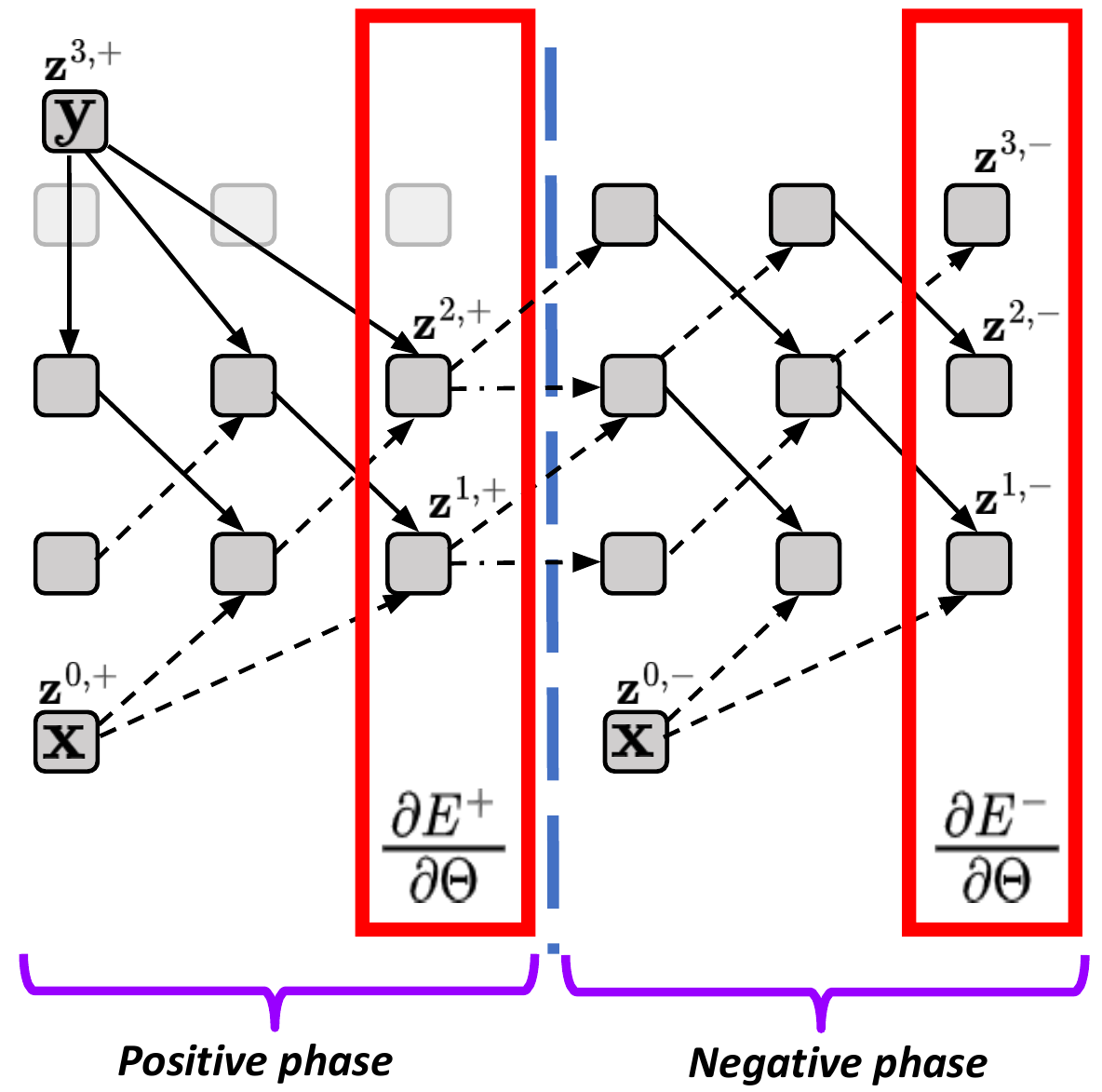}
  \end{center}
  \vspace{-0.255cm}
  \caption{\small{
  The two phases of contrastive Hebbian learning: a clamped positive phase followed by an unclamped negative phase.
  }}
  \label{fig:markov_blanket_cell}
  \vspace{-0.25cm}
\end{wrapfigure}
%%%%%%%%%%%%%%%%%%%%%%%%%%%%%%%%%%%%%%%%%%%%%%%%%%%%%%%%%%%

As mentioned earlier, given the presence of recurrent connections, if several simple conditions are met \cite{hopfield1984neurons}, it is guaranteed that a system's neural activities will settle to a fixed point. More importantly, this fixed point will lie at a minimum of an energy function. In the decomposition provided in \cite{movellan1991contrastive,oreilly_biologically_1996}, the full energy function can be broken into two terms, one that reflects the constraints imposed by the connection weights of the model which will drive neural activities to extreme values and another one that serves as a penalty function that drives neural activities towards a resting value. By evolving the activities of neurons through a differential equation (since the model is considered from the perspective of continuous time), we will find a state that is maximally harmonious with the information already encoded in the weights and one that minimizes the energy. In essence, CHL modifies synapses such that the stable states of the neuronal units match desired patterns of activations. This requires the use of a contrastive function, which has been shown to come from an even broader class of functions \cite{baldi1991contrastive}, which focuses not on the trajectories taken by the units to equilibrium but the final state found at equilibrium itself. In effect, this functional is simply $\mathcal{L}(\Theta) = E(\{\mathbf{z}^{\ell,+}\}^L_{\ell=0}) - E(\{\mathbf{z}^{\ell,-}\}^L_{\ell=0})$ (or $\mathcal{L} = E^+ - E^-$ in simplified notation), where $E(\{\mathbf{z}^{\ell,+}\}^L_{\ell=0})$ is the value of the energy function at equilibrium when output units are clamped and $E(\{\mathbf{z}^{\ell,-}\}^L_{\ell=0})$ is the equilibrium value when outputs are free. In other words, the contrastive function measures the difference of the network's Lyapunov functions \cite{drazin1992nonlinear} (or measures of stability) between clamped and free modes. After each step of learning, the difference in energy at equilibrium between clamped and free states becomes smaller. 
%We depict how contrastive Hebbian learning would be applied to the MLP example in Algorithm \ref{alg:chr}. 

Formally, in an interactive network that is trained via CHL, layer-wise activities are updated in accordance with the following neuronal dynamics:
\begin{align}
    \mathbf{z}^\ell(t + \Delta t) &= \mathbf{z}^\ell(t) + \frac{\Delta t}{\tau} \Big( -\mathbf{z}^\ell(t) + \phi^\ell \big(\mathbf{W}^\ell \cdot \mathbf{z}^{\ell-1}(t) + \gamma(\mathbf{W}^{\ell+1})^{\mathsf{T}} \cdot \mathbf{z}^{\ell+1}(t) \big) \Big) \label{eqn:chl_state_update}
\end{align}
where we notice that neural values change due to (leaky) Euler integration and the top-down and bottom-up transmitted signals are combined to produce a perturbation to the layer. Notice that the top-down recurrent feedback connections are set to be equal to the transpose of the corresponding forward ones. To produce synaptic adjustments, Equation \ref{eqn:chl_state_update} must be run iteratively for two sets of $T$ steps. The first set of steps (the positive clamped phase) entails clamping both the bottom and top layers to input and output patterns, i.e., $\mathbf{z}^0 = \mathbf{x}$ and $\mathbf{z}^L = \mathbf{y}$; this results in layer-wise activities $\{\mathbf{z}^{0,+} = \mathbf{x}\} \cup \{\mathbf{z}^{\ell,+}\}^{L-1}_{\ell=1} \cup \{\mathbf{z}^{L,+} = \mathbf{y}\}$. The second set of steps (the negative free phase) entails, starting from the current values of activities produced from the first phase, clamping only the bottom layer, i.e., $\mathbf{z}^0 = \mathbf{x}$, to produce the layer-wise activities $\{\mathbf{z}^{0,-} = \mathbf{x}\} \cup \{\mathbf{z}^{\ell,-}\}^{L}_{\ell=1}$. Once these two sets of neural activity values have been obtained, a change in plasticity is computed as follows:
\begin{align}
    \Delta \mathbf{W}^\ell = \frac{1}{\gamma^{L-\ell}} \bigg( \frac{\partial E^+}{\partial \Theta} - \frac{\partial E^-}{\partial \Theta} \bigg) 
    = \frac{1}{\gamma^{L-\ell}} \bigg( \Big(\mathbf{z}^{\ell,+} \cdot (\mathbf{z}^{\ell-1,+})^{\mathsf{T}}\Big) - \Big(\mathbf{z}^{\ell,-} \cdot (\mathbf{z}^{\ell-1,-})^{\mathsf{T}}\Big) \bigg)
\end{align}
where we notice that the recurrent dampening factor $\gamma$ (to introduce stability and prevent feedback synapses from being perfectly symmetric to the forward ones) appears in the final synaptic update, as a function of the layer index it is applied to \cite{xie2003equivalence}. % this is to introduce stability from?
The above plasticity equation demonstrates that CHL is effectively a difference between two dual-term Hebbian adjustments.

%NOTE: In Movellan's paper, his sketch at the very back of the CHL procedure starts from negative and goes to positive....so which is better??? Plus he has an exponential smoothing term for the forward Euler used to evolve the unit activities...
In \cite{movellan1991contrastive}, varieties of CHL were presented, with some instances that flipped the order of the positive/clamped and negative/free phases; each were argued to have different issues and strengths. The version presented above, as discussed in  \cite{movellan1991contrastive}, (empirically) leads to quicker training and comes with some theoretical guarantees with respect to how the energy function is optimized in different phases. Interestingly enough, different generalizations of CHL either follow this same ordering of phases (clamped then unclamped, e.g., contrastive divergence \cite{hinton2002training}, or the opposite (unclamped then clamped), e.g., equilibrium propagation \cite{scellier2017equilibrium}. %another phase-based algorithm, operates in the opposite fashion -- it starts from a negative or free phase and then proceeds to a clamped (technically weakly-clamped) phase. 
The GeneRec process \cite{oreilly_biologically_1996} (part of the Leabra framework \cite{oreilly2000computational}), as mentioned earlier in this survey, combines principles from both recirculation and CHL to construct a phase-based form of credit assignment that further approximates recurrent backprop \cite{almeida1987learning}. It can, as argued in \cite{oreilly_biologically_1996}, recover CHL when treating it as an approximation of second-order Runge-Kutta integration in tandem with a symmetry preserving constraint (connecting it to the symmetric delta rule \cite{lecun1991new}).  % symmetric midpoint variant of GeneRec

%The GeneRec setup, in essence, generalizes the circulation propagation cycle of recirculation (reviewed earlier) in the context of a flipped CHL credit assignment scheme; the `prediction phase' required clamping only the input pattern $\mathbf{z}^{0,+} = \mathbf{x}$ and letting the network find the fixed point activities for its internal and output nodes (at equilibrium, while the `training phase' further clamped the output target $\mathbf{y}$ to the output layer and let the hidden/internal nodes evolve until equilibrium (synaptic plasticity followed the equation above -- a difference between two correlation matrices were: one from a Hebbian update calculation and the other from an anti-Hebbian calculation).

\noindent
\textbf{Contrastive Divergence and Wake-Sleep.}  An early biologically-plausible connectionist model that built on the ideas of Hopfield memory was the harmonium \cite{hinton1986learning,smolensky1986information} -- also later referred to as the restricted Boltzmann machine (RBM) \cite{hinton2002training} -- trained with an algorithm known as contrastive divergence (CD), which built on the credit assignment scheme of contrastive Hebbian learning.
%algorithm proposed for learning certain connectionist graphical models, e.g. restricted Boltzmann machines (RBMs), was Contrastive Divergence \cite{hinton2002training}.  
Through a Markov Chain Monte Carlo (MCMC) sampling process, or rather, alternating block Gibbs sampling, ``fantasy'' data vectors are generated from the model after initializing a Markov chain from a clamped vector drawn from the empirical distribution / training data.  By propagating the data up to the latent variables and back down to the visible units, the model is run to ``thermal equilibrium'', and a sample is taken/drawn at this special state (serving as an approximation of the model's internal distribution) and subsequently used in what reduces to a CHL subtractive synaptic adjustment. One interpretation of this credit assignment process is that it is attempting to minimize the Kullback-Leibler (KL) Divergence \cite{hinton2002training} between two probability distributions -- the model's fantasy distribution and the data's real underlying distribution. 

%semi-restricted BMs \cite{}
Concretely, the synapses of a single-layer harmonium \cite{sejnowski1989hebb,hinton2002training} would be adjusted in accordance to optimizing its defined energy functional. Specifically, an RBM can be fully specified in terms of this functional (biases omitted):
\begin{align}
    E(\mathbf{z}^{\ell-1},\mathbf{z}^{\ell};\Theta) = -\Big( (\mathbf{z}^{\ell-1})^T \cdot \mathbf{W}^\ell \cdot \mathbf{z}^\ell \Big). \label{eqn:rbm_energy}
\end{align}
The above joint energy function can further be used to compute the marginal probability that the RBM assigns its sensory input layer $\mathbf{z}^{\ell-1}$. Formally, this means that the underlying model computes:
\begin{align}
    p(\mathbf{z}^{\ell-1}) = \frac{1}{C} \sum_{\mathbf{z}^\ell \in \mathcal{Z}_\ell} \exp\Big( -E(\mathbf{z}^{\ell-1},\mathbf{z}^{\ell};\Theta) \Big)
\end{align}
where $\mathcal{Z}_\ell$ is the set of all possible (binary) configuration values that a vector $\mathbf{z}^\ell$ could take and $C$ is the normalizing constant (or partition function); $C = \sum_{\mathbf{z}^{\ell-1} \in \mathcal{Z}_{\ell-1},\mathbf{z}^\ell \in \mathcal{Z}_\ell} \exp\Big( -E(\mathbf{z}^{\ell-1},\mathbf{z}^{\ell};\Theta) \Big)$.\footnote{An important drawback of harmoniums is the difficulty in tracking its log likelihood objective; this requires computing the partition function \emph{C} which is intractable to calculate for any reasonably-sized RBM.  Expensive approximations of the partition function, such as annealed importance sampling, or metrics like reconstruction cross entropy can be used as proxies for tracking actual progress, but do not reflect how well or how poorly an RBM model is learning.} Taking the gradient of the above marginal probability (with respect to synapses) results in a rather simple adjustment recipe:
\begin{align}
    \frac{\partial \log p(\mathbf{z}^{\ell-1})}{\partial \mathbf{W}^\ell} = \Big< \mathbf{z}^{\ell,+} \cdot (\mathbf{z}^{\ell-1,+})^{\mathsf{T}} \Big>_{data} - \Big< \mathbf{z}^{\ell,-} \cdot (\mathbf{z}^{\ell-1,-})^{\mathsf{T}} \Big>_{model} \label{eqn:energy_grad}
\end{align}
where $<\circ>_{d}$ denotes taking an expectation under the distribution $d$ specified in the subscript (such as under the $d = $`data' or $d = $`reconstruction' distribution, as in the above). The above plasticity rule only requires access to the neural activities of the RBM's input and output layers and further means that the neural system can learn to raise the probability of the input patterns it encounters by adjusting its synaptic efficacies to lower the energy on those inputs while raising the energy on others (those it does not encounter), particularly manipulating the energy for patterns that make a large contribution to the normalizing constant. 

%%%%%%%%%%%%%%%%%%%%%%%%%%%%%%%%%%%%%%%
%% CD-K Process for RBM
\begin{algorithm}[!t]
\caption{K-step contrastive divergence (CD-K) for computing synaptic updates for a two-layer harmonium.}
\label{alg:cdk}
\fontsize{8.5}{9}\selectfont
\begin{algorithmic}[1]
\State {\bfseries Input:} $ \Theta = \{ \mathbf{W}^1 \}$, $K$

\Function{ComputeUpdates}{$\mathbf{x}, \Theta$}
\LineComment Compute positive-phase activity values
\State $\mathbf{z}^{0,+} = \mathbf{x}$ \Comment Clamp input layer to sensory pattern $\mathbf{x}$
\State $\mathbf{p}^{1,+} = \sigma\Big(\mathbf{W}^1 \cdot \mathbf{z}^{0,+}\Big)$, $\mathbf{z}^{1,+}_j \sim \mathcal{B}(z^{\ell,+}_j = 1; \mathbf{p}^{1,+}_j)$
\State $\mathbf{z}^{0,-} = \mathbf{z}^{0,+}, \quad \mathbf{z}^{1,-} = \mathbf{z}^{1,+}$
\LineComment Compute negative-phase activity values
\For{$k = 1$ to $K$} \Comment Run $K$ steps of persistent block Gibbs sampling
    \State $\mathbf{p}^{0,-} = \sigma\Big((\mathbf{W}^1)^{\mathsf{T}} \cdot \mathbf{z}^{1,-} \Big)$, $\mathbf{z}^{0,-}_i \sim \mathcal{B}(z^0_j = 1; \mathbf{p}^{0,-}_i)$
    \State $\mathbf{p}^{1,-} = \sigma\Big(\mathbf{W}^1 \cdot \mathbf{z}^{0,-}\Big)$, $\mathbf{z}^{1,-}_j \sim \mathcal{B}(z^{\ell,-}_j = 1; \mathbf{p}^{1,-}_j)$
\EndFor

\LineComment Compute synaptic updates:
\State $\Delta \mathbf{W}^1 = \mathbf{z}^{1,+} \cdot (\mathbf{z}^{0,+})^{\mathsf{T}} - \mathbf{z}^{1,-} \cdot (\mathbf{z}^{0,-})^{\mathsf{T}}$ \Comment Update based on Equation \ref{eqn:cd_update}
\State \Return $\{\Delta \mathbf{W}^1\}$
\EndFunction
\end{algorithmic}
\end{algorithm}
%%%%%%%%%%%%%%%%%%%%%%%%%%%%%%%%%%%%%%%

Measuring the neural activities in an RBM is also straightforward and efficient since we may exploit a key simplification made to its synaptic structure -- there are no lateral connections between its units in any layer. This lack of cross-layer connections facilitates the use of block Gibbs sampling \cite{gelfand2000gibbs} (since the individual units in a layer can be assumed to be conditionally independent of one another given the other layer) to first collect samples of the RBM's data-dependent (i.e., positive or clamped) phase and then samples of its data-independent phase (i.e., negative or unclamped). As a consequence, sampling the model's activities is done as follows:
\begin{align}
    \mathbf{p}^\ell &= \phi^\ell(\mathbf{W}^\ell \cdot \mathbf{z}^{\ell-1}), \; \text{where } \mathbf{p}^\ell_j = p(z^\ell_j | \mathbf{z}^{\ell-1}) \; \text{and} \; 
    \mathbf{z}^{\ell}_j \sim \mathcal{B}(z^\ell_j=1; \mathbf{p}^{\ell}_j) \label{eqn:hid_state} \\
    \mathbf{p}^{\ell-1} &= \phi^\ell((\mathbf{W}^\ell)^{\mathsf{T}} \cdot \mathbf{z}^{\ell}), \; \text{where } \mathbf{p}^{\ell-1}_j = p(z^{\ell-1}_j | \mathbf{z}^{\ell}) \; \text{and} \; \mathbf{z}^{\ell-1}_i \sim \mathcal{B}(z^{\ell-1}_i=1; \mathbf{p}^{\ell-1}_i) \label{eqn:vis_state}
\end{align} %% make tiny algo to show CD-K
noting that $\sim \mathcal{B}(z=1|p)$ denotes sampling from the Bernoulli distribution (with probability that the value is one parameterized by a value $p$). Values collected from the positive phase are tagged with `+' as in $\mathbf{z}^{\ell,+}$ whereas those obtained a negative phase are tagged with `-', as in $\mathbf{z}^{\ell,-}$. To find the values needed for Equation \ref{eqn:energy_grad}, Equations \ref{eqn:hid_state} and \ref{eqn:vis_state} are run several times in succession; however, computing the second term of Equation \ref{eqn:energy_grad} (taking the expectation under the model distribution) is still very difficult, requiring the construction of a Markov chain Monte Carlo process, running Equations \ref{eqn:hid_state} and \ref{eqn:vis_state} many times until until (statistical) equilibrium has been reached. Instead, as a fast approximation to the log gradient update, \cite{hinton2002training,hinton2012practical} argued that, instead, a far simpler contrastive (correlational) scheme could be leveraged, replacing the second model distribution term with a reconstruction distribution term. This rule uses only two subsequent applications of \ref{eqn:hid_state} and \ref{eqn:vis_state} (to obtain positive and negative phase statistics):
\begin{align}
    \Delta \mathbf{W}^\ell \approx \Big< \mathbf{z}^{\ell,+} \cdot (\mathbf{z}^{\ell-1,+})^{\mathsf{T}} \Big>_{data} - \Big< \mathbf{z}^{\ell,-} \cdot (\mathbf{z}^{\ell-1,-})^{\mathsf{T}} \Big>_{recon}. \label{eqn:cd_update}
\end{align}
The above update rule is referred to as contrastive divergence (CD, or CD-1); a consequence of this fast rule is that it is technically optimizing the difference between two KL divergences, sans a tricky term, rather than the actual negative log likelihood of $p(\mathbf{x})$ \cite{sutskever2010convergence,hinton2012practical}. Nevertheless, in practice, this plasticity rule works well for training RBMs. A variation of CD is presented in Algorithm \ref{alg:cdk}, where we depict a slightly more accurate implementation that uses $K$ sampling steps to estimate the reconstruction term, i.e., \emph{CD}-K. Note that computing the synaptic updates/gradients via CD does entail using a biased sample, however, the variance introduced through this approximation is not significant enough to reduce the learning process's effectiveness. 
Further improvements to CD-centric credit assignment have been explored \cite{tieleman2009using,tieleman2008training,bengio2009justifying,romero2019weighted}; notably, a useful improvement is known as parallel tempering \cite{desjardins_tempered_2010}, which exploits the advantages offered by persistent Gibbs (block) sampling, alternatively referred as stochastic maximum likelihood. In essence, one maintains one or more continuously running chains in the background, using either random chain swapping or an expectation calculated across these ``fantasy particles'', to facilitate a broader and faster exploration of the RBM's internal configuration state space \cite{desjardins_tempered_2010}. 

Although this survey's treatment of harmoniums and contrastive divergence-based learning has been presented in the context of arbitrary, multi-layer systems, training `deep harmoniums' has been shown to be a challenging endeavor \cite{bengio2006greedy,salakhutdinov2008quantitative,hinton2012better,montavon2012deep,ororbia_deep_hybrid_2015b}; note that such systems are also referred to as deep belief networks (DBNs) \cite{hinton2006fast,mohamed2009deep,salakhutdinov2009deep} and, when additional top-down recurrent connections are introduced, as deep Boltzmann machines (DBMs) \cite{salakhutdinov2010efficient,cho2013gaussian}. 
For a DBM, the neuronal activities are a function of bottom-up and top-down pressures as in:
\begin{align}
    \mathbf{z}^\ell = \phi^\ell\Big(\mathbf{W}^\ell \cdot \mathbf{z}^{\ell-1} + (\mathbf{W}^{\ell+1})^T \cdot \mathbf{z}^{\ell+1} \Big) \label{eqn:dbm_activity}
\end{align}
where the top-down recurrent weights are set to be the transpose of the weight matrix of the layer above. Given their probabilistic, sampling-based nature, one specific scheme generalizes contrastive divergence to what is known as the stochastic approximation procedure (SAP). In effect, SAP entails carefully crafting a block Gibbs sampler that alternates the sampling of evenly-numbered and oddly-numbered layers in the harmonium system. The SAP learning process preserves a synergistic form of locality in the credit assignment although this comes at the price of greater external control (to coordinate the interaction of layers); SAP further requires custom initialization procedures for the layer-wise activities, such as greedy layer-wise pre-training of a stack of RBMs. %, to ensure effective operation.

%%%%%%%%%%%%%%%%%%%%%%%%%%%%%%%%%%%%%%%%%%%%%%%%%%%%%%%%%%%
%% Helmholtz machine - sleep phase
\begin{wrapfigure}{r}{0.5\textwidth}
\vspace{-0.45cm}
  \begin{center}
    \includegraphics[width=0.27\textwidth]{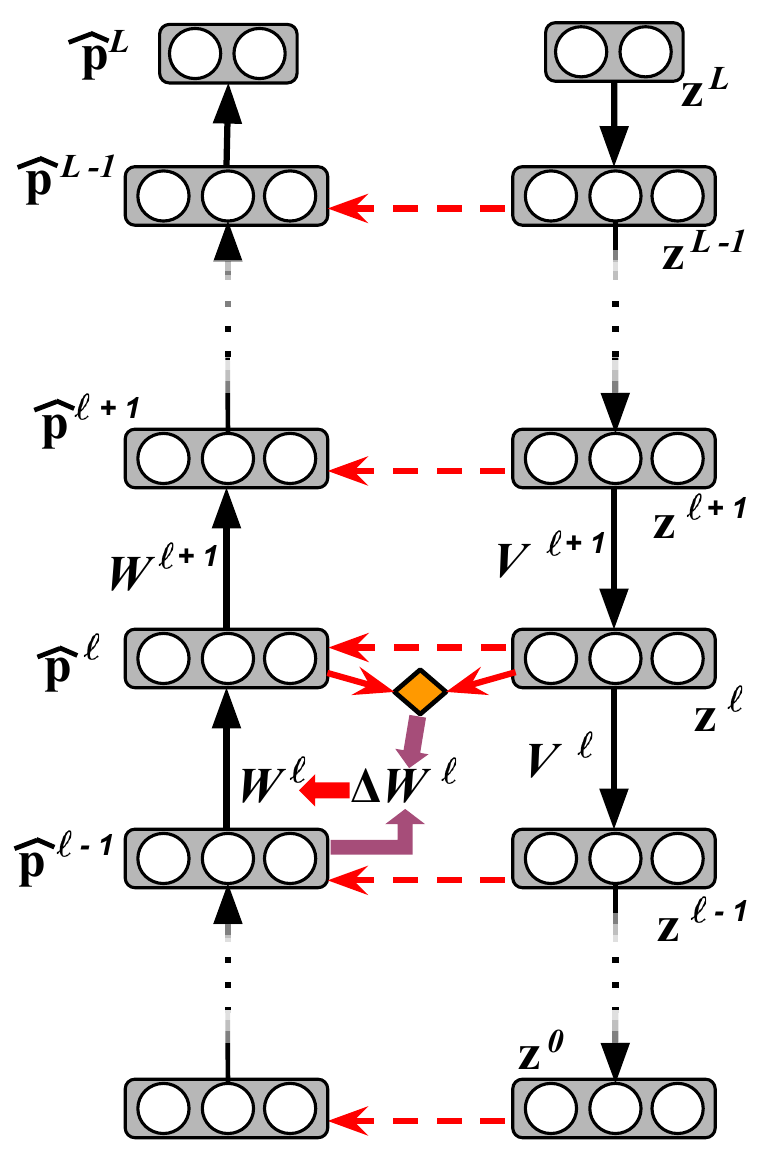}
  \end{center}
  \vspace{-0.255cm}
  \caption{\small{
  The sleep phase of a Helmholtz machine: the top-down directed generative model (right network) is sampled ancestrally to obtain a set of neural activities that drive the recognition model (left network) to produce layer-wise probabilities. Connections between $\ell-1$ and $\ell$ are updated via a local product of pre-synaptic activity $\mathbf{z}^\ell$ and post-synaptic probabilities $\mathbf{\widehat{p}}^\ell$.
  }}
  \label{fig:helmholtz_sleep}
  %\vspace{-0.25cm}
\end{wrapfigure}
%%%%%%%%%%%%%%%%%%%%%%%%%%%%%%%%%%%%%%%%%%%%%%%%%%%%%%%%%%%

The wake-sleep algorithm \cite{hinton1995wake,bornschein2014reweighted} was originally proposed as a simple way to learn a generative model composed of two opposite flowing directed sub-models; the Helmholtz machine \cite{hinton1993autoencoders,dayan1995helmholtz,hinton1995helmholtz,dayan1996varieties,dayan2000helmholtz} and the sigmoid belief network \cite{saul1996mean,sutskever2008deep} are among some of the more prominent models that adhere to this framework. Helmholtz machines, which are neural systems trained by optimizing a (Helmholtzian) free energy functional (average energy minus entropy; which also corresponds to the surprise of the generative model weighted by the probability of the data) \cite{hinton1995helmholtz,kirby2006tutorial}, consist of a network that is used to generate samples (known as the `generative distribution') and another network that is used for inference (known as the `recognition distribution'). 
%% wake-phase of Helmholtz machine
The ``wake phase'' of the learning algorithm entails presenting a data point to the inference model, propagating the activities forward across this network (co-model), and then updating the generative network (co-model), at each layer, to make it more likely to generate the values found in the inference network. Formally, this step of the wake-sleep process requires running the recognition and generative co-models (with $\mathbf{z}^0 = \mathbf{x}$) in the following manner:
\begin{align}
    \mathbf{p}^\ell &= \phi^\ell(\mathbf{W}^\ell \cdot \mathbf{z}^{\ell-1}), \;  \text{and, } \; \mathbf{z}^\ell_j \sim \mathcal{B}(z^\ell_j = 1; \mathbf{p}^\ell_j) \; \text{for} \; \ell = 1,2,...,L \label{eqn:ws_recog_wake} \\ 
    \mathbf{\widehat{p}}^L &= \phi^L(\mathbf{b}^L), \; \text{and, } \; \mathbf{\widehat{p}}^\ell = \phi^\ell(\mathbf{V}^\ell \cdot \mathbf{z}^\ell) \; \text{for} \; \ell=L-1,L-2,...,0 \label{eqn:ws_gen_wake}
\end{align}
which then triggers the following synaptic adjustments to the generative model:
\begin{align}
    \Delta \mathbf{b}^L &= \mathbf{z}^L - \mathbf{\widehat{p}}^L \; \text{and} \; 
    \Delta \mathbf{V}^\ell = (\mathbf{z}^\ell - \mathbf{\widehat{p}}^\ell) \cdot (\mathbf{z}^{\ell+1})^{\mathsf{T}}. \label{eqn:ws_gen_update}
\end{align}
Notice that we have introduced a particular bias vector $\mathbf{p}^L$ which serves as the parameters of the prior of the generative co-model.  
%% sleep-phase of Helmholtz machine
The ``sleep phase'' involves sampling the generative model to obtain a set of ``fantasy'' samples and the inference model is adapted to better infer the fantasy samples' causes.  Formally, this entails:
\begin{align}
    \mathbf{p}^L &= \phi^\ell(\mathbf{b}^L), \; \text{and } \; \mathbf{z}^L_j \sim \mathcal{B}(z^L_j = 1; \mathbf{p}^L_j) \\
    \mathbf{p}^\ell &= \phi^\ell(\mathbf{V}^\ell \cdot \mathbf{z}^{\ell+1}) \; \text{and } \; \mathbf{z}^\ell_j \sim \mathcal{B}(z^\ell_j = 1; \mathbf{p}^\ell_j) \; \text{for} \; \ell = L-1,L-2,...,0 \label{eqn:ws_gen_sleep} \\ 
    \mathbf{\widehat{p}}^\ell &= \phi^\ell(\mathbf{W}^\ell \cdot \mathbf{z}^{\ell-1}), \; \text{for} \; \ell = 1,2,...,L
    \label{eqn:ws_recog_sleep}
\end{align}
which then triggers the following synaptic adjustments to be made to the recognition model:
\begin{align}
    \Delta \mathbf{W}^\ell = (\mathbf{z}^\ell - \mathbf{\widehat{p}}^\ell) \cdot (\mathbf{z}^{\ell-1})^{\mathsf{T}}. \label{eqn:ws_recog_update}
\end{align} 
The above process for a Helmholtz machine -- Equations \ref{eqn:ws_recog_wake}-\ref{eqn:ws_gen_wake}, \ref{eqn:ws_recog_sleep}-\ref{eqn:ws_gen_sleep}, and synaptic updates via Equations \ref{eqn:ws_gen_update} and \ref{eqn:ws_recog_update} -- can be tracked by measuring the KL divergence between probability of the data and the probabilities produced by the top-down directed generative model \cite{kirby2006tutorial}. Beyond the specialized wake-sleep credit assignment scheme applied to Helmholtz machines, the up-down back-fitting algorithm \cite{hinton2006fast} was developed as a generalization of wake-sleep for fine-tuning a deep belief network (after it was constructed via a greedy, layer-wise pre-training phase). Notably, this scheme addressed/circumvented one of the wake-sleep algorithm's issues: the problem of ``mode averaging''. In wake-sleep, it is possible for the inference weights to pick a particular mode of the layer above and remain ``stuck'' to it, even if other modes would be useful to infer when learning to generate sensory data; this hinders the effectiveness of the sleep phase (downward pass), constraining the training of the recognition weights to whatever mode it uncovers first without exploring other useful modes of the data's underlying distribution. In contrast, in the up-down algorithm, instead of starting the generative pass using a completely random sample that is input to the topmost layer of the generative co-model, instead, the sample synthesized at the top is biased by running a Gibbs sampler for a few steps using the associative memory formed by the top-most RBM of the DBN.

Beyond DBNs, DBMs, and Helmholtz machines, many other models and algorithms for learning generative systems take inspiration from contrastive divergence and wake-sleep, including the variational walkback algorithm \cite{bengio2013generalized,goyal2017variational} and equilibrium propagation \cite{scellier2017equilibrium} (reviewed next). Furthermore, research efforts related to Helmholtz machines and harmoniums not only helped to spark the `deep learning revolution', through the stacking of RBMs to pre-train deep neural architectures \cite{bengio2006greedy,mohamed2009deep}, but they also inspired later incarnations of neural generative models, e.g., those learned through the framework of neural variational inference, e.g., variational autoencoders \cite{kingma2013auto,serban2017piecewise}.

\noindent
\textbf{Equilibrium Propagation.} The equilibrium propagation (EProp) procedure \cite{bengio2015early,scellier2016towards,scellier2017equilibrium,scellier2018generalization}, like the other schemes within this family, centers around an energy functional, specifically one that corresponds to the continuous Hopfield network model \cite{hopfield1984neurons,bengio2015early}. Although EProp shares many similarities to contrastive Hebbian learning, there are some subtle yet important differences. % that allow addressal of theoretical issues tied to CHL and CD. % Talk about these issues here...
A setting where EProp looks markedly different from CHL is when supervised learning is the goal; in this context, an external potential energy $\mathcal{L}(\mathbf{z}^L,\mathbf{y})$ is introduced to drive the neural system's output units towards target values, e.g., encoded labels.\footnote{Note that \cite{scellier2016towards,scellier2017equilibrium} presented the neural system in terms of a full Hopfield model, from which a feedforward network structure could be extracted. This survey directly presents EProp in terms of the feedforward, hierarchical structure for simplicity; however, the credit assignment scheme applies to arbitrarily connected systems.} For instance, \cite{scellier2017equilibrium} employed a quadratic cost for $\mathcal{L}(\mathbf{z}^L,\mathbf{y}) = \frac{1}{2}||\mathbf{z}^L - \mathbf{y}$. This external energy (or `force') is then combined with the usual Hopfield function $E$ to create a total energy functional of the form: $\mathcal{F} = E(\{\mathbf{z}^\ell\}^L_{\ell=1}) + \gamma \mathcal{L}(\mathbf{z}^L,\mathbf{y})$ (or, in a simplified format: $F = E + \gamma \mathcal{L}$). The external energy value is itself further weighted by the influence factor $\gamma \geq 0$ which controls the degree to which the output units $\mathbf{z}^L$ are pushed towards the target $\mathbf{y}$; this yields the notion of  `weakly-clamped' (or softly clamped) outputs as to the opposed to fully/hard-clamped output values inherent to the earlier CHL/CD schemes. The total energy function $\mathcal{F}$ accounts for both the internal interactions within the network (via the term $E$) as well as how the target variables influence/nudge the output units towards desired states (via $\mathcal{L}$).

As in CHL, neural activities are described with a differential equation (for motion) which takes the form:
\begin{align}
    \frac{\partial \mathbf{h}^\ell}{\partial t} = \frac{\partial E}{\partial \mathbf{h}^\ell} - \gamma \frac{\partial \mathcal{L}}{\partial \mathbf{h}^\ell} \label{eqn:eprop_activities}   
\end{align}
where the total energy of the system decreases as time progresses (notice that the above equation is specifically expressed in terms of pre-transformed neural activities $\mathbf{h}^\ell$).  
The equation for neuronal evolution -- the first term on the right hand side of Equation \ref{eqn:eprop_activities} -- can take on one of the following forms:
\begin{align}
    \frac{\partial E}{\partial \mathbf{h}^\ell} &= \partial\phi^\ell(\mathbf{h}^\ell) \odot \Big( \mathbf{W}^\ell \cdot \phi^{\ell-1}(\mathbf{h}^{\ell-1}) \Big) - \mathbf{h}^\ell, \; \text{or, } \label{eqn:eprop_state_diff}\\
    \frac{\partial E}{\partial \mathbf{h}^\ell} &= \Big( \mathbf{W}^\ell \cdot \phi^{\ell-1}(\mathbf{h}^{\ell-1}) \Big) - \mathbf{h}^\ell \label{eqn:eprop_state_rate}
\end{align}
where the first form (Equation \ref{eqn:eprop_state_diff}) follows from \cite{scellier2016towards} whereas the second one is the simpler point neuron model \cite{dayan2001theoretical}, used in \cite{scellier2018generalization} (the rate-based leaky integrator neuronal model); using this second form avoids requiring the first derivative of the activation, i.e., $\partial \phi^\ell(\circ)$. 
Crucially, the second term of Equation \ref{eqn:eprop_activities} depends on the choice of cost function used for the external force; in the case of the quadratic cost of \cite{scellier2017equilibrium}, this simply becomes $-\gamma \frac{\partial \mathcal{L}}{\partial \mathbf{h}^L} = \gamma (\mathbf{y} - \mathbf{h}^L)$ for the output units and $-\gamma \frac{\partial \mathcal{L}}{\partial \mathbf{h}^\ell} = 0$ for all other units.

To compute synaptic weight updates under the EProp scheme, like CHL/CD, two phases of computation are used (desirably using the same neural machinery in both instances) but, unlike CHL/CD, in a different order. In the first phase, we clamp the inputs to a data vector, set $\gamma = 0$ -- this is the negative or free (data / context-independent) phase -- and iteratively update each layer of (pre-transformed) neural $\mathbf{h}^\ell$ activities using the forward Euler method until a fixed point $\mathbf{h}^{\ell,-}$ is reached. Prediction can also be carried out upon reaching this fixed point as well; at the end of the negative, the output unit values are read out from $\mathbf{h}^L$. The second (positive) phase -- the data/context-dependent phase --- is then initiated by setting $\gamma > 0$ to a small positive value, permitting the external energy term to play a role in the measure of the total energy $F$, critically nudging the output units to the desired target or context $\mathbf{y}$. Since this force only directly impacts the output units (at their free fixed point), the effect of this nudging will reach the internal hidden neurons as time progresses, by iterating one of the neuronal differential equations above (Equation \ref{eqn:eprop_state_diff} or \ref{eqn:eprop_state_rate}). This (activation) spreading of the external force over the internal units of the neural system can also be viewed as conducting a form of backprop itself \cite{scellier2016towards,scellier2018generalization} but without the required global feedback pathway that uses different neural calculations. In fact, the second phase of this algorithm has been shown to approximate backprop, or rather, it ``back-propagates'' error derivatives across the network \cite{bengio2015early,bengio2017stdp}. Nevertheless, once the network settles to a new fixed point during the weakly-clamped positive phase we can then use the activities of the neurons acquired here, i.e., $\{\mathbf{x}\} \cup \{\mathbf{h}^{\ell,+}\}^{L-1}_{\ell=1} \cup \{\mathbf{y}\}$, and compare with those found at the end of the free/negative phase, i.e., $\{\mathbf{x}\} \cup \{\mathbf{h}^{\ell,-}\}^{L}_{\ell=1}$. As a result, the ultimate update applied to connection weights can then be calculated in one of two ways:
\begin{align}
\Delta \mathbf{W}^\ell &= \phi^\ell(\mathbf{h}^{\ell,+}) \cdot \Big(\phi^{\ell-1}(\mathbf{h}^{\ell-1,+})\Big)^{\mathsf{T}} - \phi^\ell(\mathbf{h}^{\ell,-}) \cdot \Big(\phi^{\ell-1}(\mathbf{h}^{\ell-1,-})\Big)^{\mathsf{T}}, \; \mbox{or, } \label{eprop:chl}\\
\Delta \mathbf{W}^\ell &= (\mathbf{h}^{\ell,+} - \mathbf{h}^{\ell,-}) \cdot \Big(\phi^{\ell-1}(\mathbf{h}^{\ell-1,-})\Big)^{\mathsf{T}} \mbox{.} \label{eprop:stdp}
\end{align}
The first rule is a contrastive rule, taking a form similar to the plasticity update induced by CHL; note that, however, the output unit values in the positive phase are the result of nudging towards a target whereas, in CHL, the outputs would be fully clamped to the output context/target $\mathbf{y}$, i.e., $\gamma = \infty$.\footnote{This means that, in CHL, the output units are completely insensitive to the internal force induced by the rest of the units of the neural system. This is an issue that EProp directly resolves.} This difference in clamping (`weak' in EProp versus `hard' in CHL) clearly stems from the fact that each optimizes a different energy functional, i.e., $\mathcal{L} = E^+ - E^-$ in CHL and $F = E + \gamma \mathcal{L}$ in EProp. CHL's energy optimization suffers from the (theoretical) fact that the objective could take on negative values if the clamped and free fixed points stabilize in different modes of the energy function; EProp, in contrast, does not face this issue given that its energy functional's form guarantees that the weakly-clamped (positive phase) fixed point will be close to the free (negative phase) fixed point, i.e., both fixed points lie within the same mode of the energy function -- this is also partly the result of reordering the negative phase to come before the positive phase \cite{movellan1991contrastive}. The second rule (Equation \ref{eprop:stdp}) is a simplified Hebbian rule based on the proposed spike-timing dependent plasticity-like rule of \cite{bengio2017stdp}. The central idea of this rule, which assumes no synaptic modifications during the negative/free phase, is to multiply the neuronal pre-activities with the rate of change of the post-activities, or simply $\Delta \mathbf{W}^\ell = \Delta \mathbf{h}^\ell \cdot (\phi^{\ell-1}(\mathbf{h}^{\ell-1}))^{\mathsf{T}}$, where the first term would ideally be a temporal derivative (which is what encodes the necessary information for driving weight plasticity in spiking neural systems). In effect, the rule shown in Equation \ref{eprop:stdp} is a simple secant approximation of this temporal change whereas the first rule Equation \ref{eprop:chl} is derived by integrating over the full trajectory taken by the neural activities during the each phase \cite{scellier2018generalization}. Desirably, it is important to note that equilibrium propagation has been generalized to operate with finer-grained dynamics, such as those that characterize spiking neuronal units, as in \cite{o2019training,martin2021eqspike}.

% connection to recurrent backprop
An important classical algorithm that is intimately related to EProp is recurrent backprop \cite{almeida1987learning,pineda1987generalization,pineda1989recurrent}, where, for a given set of (clamped) inputs and target states, the neural model is trained to settle into a stable activation state where the output units correspond to/align with desired target activity. Recent work has notably revised recurrent backprop, proposing useful  modifications to improve its performance/effectiveness \cite{liao2018reviving}. One key biological implausibility of recurrent backprop, however, is that, although it considers the same objective that EProp optimizes, its constrained (Lagrangian) formulation of the optimization problem leads to computing a fixed point for its second phase using a linearized form the recurrent network itself. This crucially means that the second phase of the algorithm does not follow the same dynamics as the first phase, thus implausibly (from a biological standpoint) requiring two different sets of neural computation for its inference and learning.

\subsubsection{Forward-Only Learning} 
\label{sec:forward_only}

One of the more recent families of credit assignment schemes to have been proposed -- forward-only processes \cite{williams1990gradient,heinz1995pipelined,hirasawa1996forward,ohama2004forward,kohan2018error,kohan2023signal,hinton2022forward,ororbia2023predictive,zhao2023cascaded,lee2023symba} -- focus on producing synaptic adjustments exclusively with a neural system's inference machinery, particularly working to offer a computational explanation of how neurons might adapt/receive learning signals without the presence of recurrent feedback parameters (as in predictive coding). Notably, forward-only approaches also, like those under the discrepancy reduction and energy-based families, employ a set of local loss functionals but particularly enjoy their global synergy as a consequence of two or more applications of forward propagation of information through the neural system itself. In effect, a second `forward pass' replaces the usual `backward pass' (or reverse set of calculations) and, desirably, is done so in parallel; this further circumvents the need for chained/conditioned relaxation phases as in energy-based schemes (CHL/CD). %\textcolor{red}{Resolve a major weakness in phase-based learning approaches, that require conditioning negative phase on positive phase (phases are now run in parallel)}
Forward-only schemes have been argued to provide the fewest constraints on the neural system (including both its inference and learning processes) \cite{kohan2018error,kohan2023signal}, given that its central imperative is to utilize one kind of computation for inference and learning and thus does not require feedback connectivity (further side-stepping the weight transport problem). In addition, synaptic parameters for any layer of neurons are updated as information is propagated to that layer; learning signals ``travel'' with the inputs across the ANN simultaneously. A positive side-effect of forward-only credit assignment is that, in addition having a lower computational complexity (e.g., very cheap or no iterative inference/relaxation is needed), it reduces the cost of additional memory, a price paid by many algorithms reviewed in the previous sections (for storing activation patterns or holding extra feedback connectivity structures/parameters), making it an ideal candidate for implementation on neuromorphic, edge-computing hardware \cite{kohan2023signal,ororbia2023learning}.

While the development of forward-only credit assignment is rather new (\cite{kohan2018error} was one of the earlier scalable proposals of forward-only learning and \cite{linsker1988self} was the earliest proposal of local self-supervised forward adaptation), having become more prominent over the last few years, two particular classes of forward-only learning procedures currently exist. Specifically, there are those that leverage a `supervised context' to drive a parallel (target-oriented) pathway and there are those that center around `self-supervised context' (typically relying on contrastive objectives) to drive theirs. In algorithms that are supervised context-driven \cite{kohan2018error,kohan2023signal,zhao2023cascaded}, the hidden representation of a sensory input and the representation of its corresponding context, e.g., label, are brought closer together (while also pushed farther apart from other pairs of sensory inputs and their contexts) via a set of local loss functionals. Specifically, the total loss for neural systems learned in this manner can be specified as  $\mathcal{L}(\Theta) = \sum^L_{\ell=1} \mathcal{L}^\ell(\mathbf{z}^\ell,\mathbf{c}^\ell)$ (notice that this expression is similar to the form that discrepancy reduction losses take -- a sum of local measurements). 
%provide guiding signals to push these two sets representations to be separate and distinct. 
Concretely, an algorithm such as signal propagation \cite{kohan2018error,kohan2023signal} runs both a sensory data pattern $\mathbf{x}$ and its corresponding context label $\mathbf{y}$ as input through the neural model (our example MLP), producing a set of compound activities $\mathcal{Z} = \{ [\mathbf{z}^\ell, \mathbf{c}^\ell ]\}^L_{\ell=1}$, as follows:
\begin{align}
    \big[\mathbf{z}^\ell, \mathbf{c}^\ell\big] = 
    \begin{cases} 
      \Big[ \phi^\ell(\mathbf{W}^\ell \cdot \mathbf{z}^{\ell-1}), \phi^\ell(\mathbf{C}^\ell \cdot \mathbf{c}^{\ell-1})\Big] & \ell = 1
      \\
      \phi^\ell\Big(\mathbf{W}^\ell \cdot \big[\mathbf{z}^{\ell-1}, \mathbf{c}^{\ell-1}\big]\Big) & \mbox{otherwise,}  %\phi^\ell\Big(\big[\mathbf{W}^\ell, \mathbf{C}^\ell\big] \cdot \mathbf{z}^{\ell-1}\Big)
    \end{cases} \label{eqn:sigprop_inf}
\end{align}
where we notice that a neuronal layer is represented as a composition of two separate vectors of values, i.e., $[\mathbf{z}^\ell, \mathbf{c}^\ell] = \mathbf{m}^\ell \in \mathbb{R}^{2\mathcal{J}_\ell \times 1}$ (except for input layer $\ell=0$, where $[\mathbf{x}, \mathbf{y}] = \mathbf{m}^0 \in \mathbb{R}^{(\mathcal{J}_0 + C) \times 1}$), since, as mentioned above, there are two forward pathways run in parallel. Immediately, after each layer $\ell$ of neural activity values have been computed as in Equation \ref{eqn:sigprop_inf}, the synapses may be adjusted in accordance with the following rule:
\begin{align}
    \Delta \mathbf{W}^\ell = \frac{\partial \mathcal{L}^\ell(\mathbf{z}^\ell, \mathbf{c}^\ell)}{\partial \big[\mathbf{z}^\ell, \mathbf{c}^\ell\big]} \cdot \Big([\mathbf{z}^{\ell-1},\mathbf{c}^{\ell-1}]\Big)^{\mathsf{T}}
\end{align}
where the local loss could take on many forms such as a distance function $\mathcal{L}^\ell(\mathbf{z}^\ell, \mathbf{c}^\ell) = ||\mathbf{c}^\ell - \mathbf{z}^\ell||^2_2$ or a (cosine) similarity measurement $\mathcal{L}^\ell(\mathbf{z}^\ell, \mathbf{c}^\ell) = \frac{(\mathbf{c}^\ell)^{\mathsf{T}} \cdot \mathbf{z}^\ell}{||\mathbf{z}^\ell||_2 \mathbf{z}^\ell||_2}$. Notice that the local loss at $\ell$ separates the context-oriented pathway $\mathbf{c}^\ell$ from the input-oriented pathway activity $\mathbf{z}^\ell$ so as to extract the layer-wise target that will drive/trigger the synaptic update for the $\ell$th layer.\footnote{At test time inference, note that, to make a prediction in the above model, one would need to cycle through possible context values $\mathbf{y}$ and select the one that yields the lowest total loss (a prediction scheme often used in energy-based models such as harmoniums \cite{hinton2012practical}). 
Other schemes for crafting the dual forward pathways are possible, beyond the one presented here, if one did want to spare the expense for this type of test-time inference \cite{kohan2018error,kohan2023signal}.} 
Error forward-propagation (EFP) is notable variation of the supervised context-driven theme presented above, where an error signal \cite{kohan2018error} or error-modulated input pattern is propagated through the network instead of the context \cite{heinz1995pipelined,hirasawa1996forward,ohama2004forward,ren2022scaling,dellaferrera2022error}. For example, a scheme may, such as the one in \cite{dellaferrera2022error}, introduce a single set of fixed random feedback connections that wires the output error directly to the input, producing an error-perturbed version of the sensory input, i.e., $\mathbf{\tilde{x}} = \mathbf{x} + \mathbf{B} \cdot \frac{\partial \mathcal{L}(\mathbf{z}^L, \mathbf{y})}{\partial \mathbf{z}^L}$, which is then subsequently run through the neural model to produce context-driven target values for the hidden layers.\footnote{Note that this type of scheme yields spatially local but not temporally local synaptic updates; this stems from the fact that the second forward pass is not completed in parallel but constrained to operate sequentially. Furthermore, this scheme has the unfortunate drawback that the feedback synapses are not learned, as in feedback alignment \cite{lillicrap2016random}; though weight mirrors \cite{akrout2019deep} could be employed to rectify this issue.}

%%%%%%%%%%%%%%%%%%%%%%%%%%%%%%%%%%%%%%%
%% forward-only schemes
\begin{figure}[!t]
\centering     %%% not \center
\begin{subfigure}{0.495\textwidth}
    \centering
    %\vspace{-5mm}
     \includegraphics[width=0.565\linewidth]{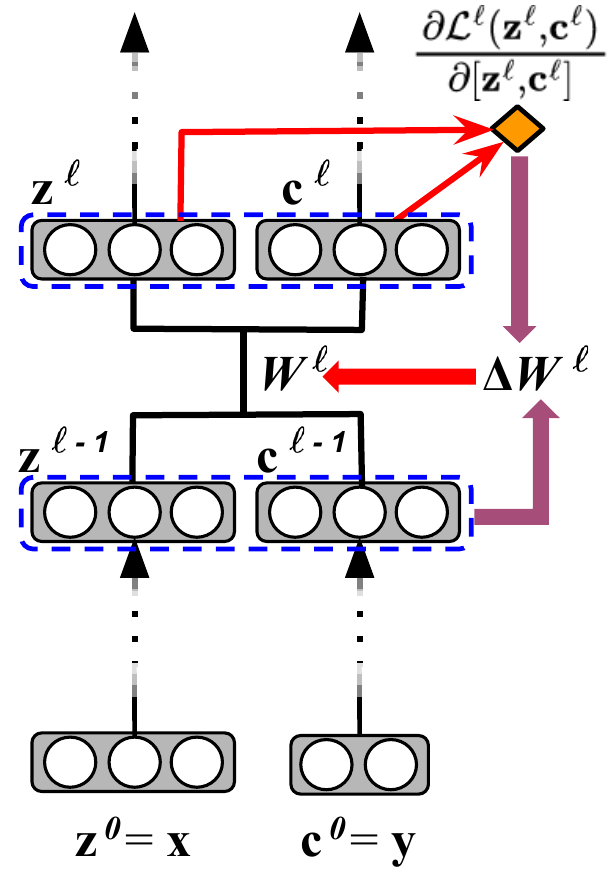} 
    % \caption{Signal propagation.}
    % \label{fig:sigprop}
\end{subfigure}
\begin{subfigure}{0.495\textwidth}
    \centering
    %\vspace{-5mm}
     \includegraphics[width=0.75\linewidth]{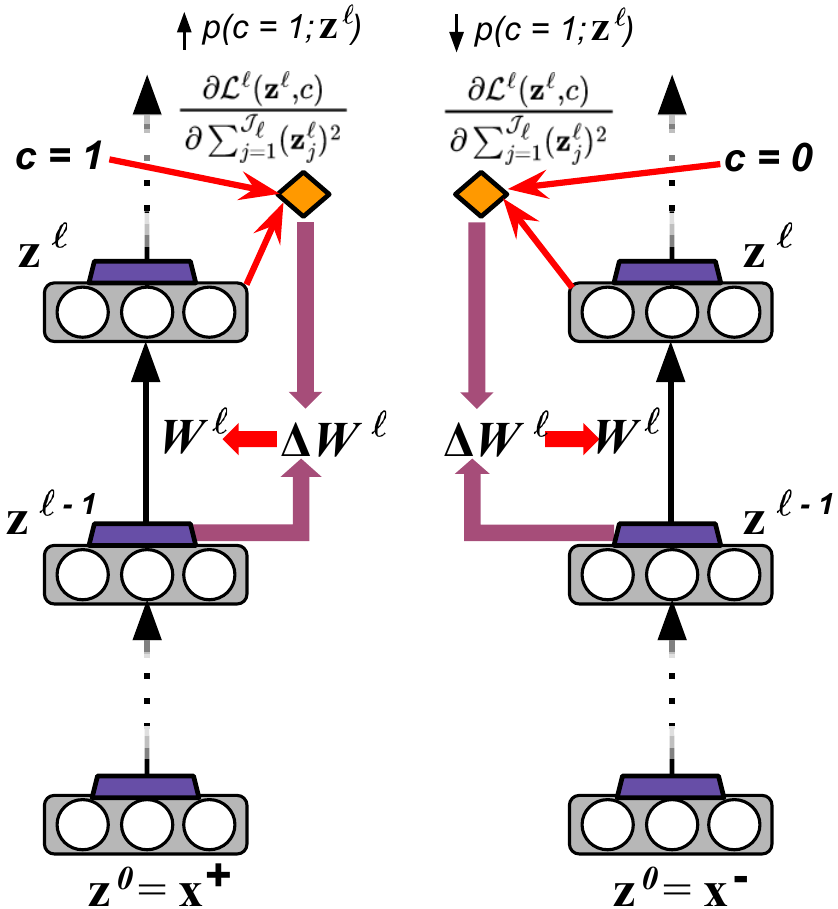} 
    % \caption{Forward-forward learning.}
    % \label{fig:ffprop}
\end{subfigure}
%\vspace{-0.3cm}
\caption{\small{\textbf{Forward-Only Credit Assignment.} 
Depicted are types of forward-only learning: 
(Left) supervised context-centered with signal propagation (sigprop) as an example. The dashed blue box indicates that the activation at $\ell$ is composed (via concatenation) producing a larger vector that feeds into the weight update. 
(Right) Self-supervised-context with the forward-forward algorithm (FFA) as an example. Notice that, in FFA, the kind of the sensory input provided to the network affects what the synaptic update will do:  a positive (original) datapoint $\mathbf{x}^+$ raises ($\uparrow$) the goodness probability $p(c=1;\mathbf{z}^\ell)$ of layer $\ell$ while a negative (confabulated) datapoint $\mathbf{x}^-$ lowers ($\downarrow$) the goodness probability $p(c=1;\mathbf{z}^\ell)$ of layer $\ell$. In FFA, Layer normalization is implied by the purplish trapezoids that appear right above each neuronal layer of activities; note that the weight update $\Delta \mathbf{W}^\ell$ uses the layer-normalized pre-synaptic activities but the raw unnormalized post-synaptic activities are used in the local goodness functional.
In either scheme (sigprop or FFA), a local functional produces a signal that triggers a synaptic update for $\mathbf{W}^\ell$. 
}}
\label{fig:forward_only}
\vspace{-0.6cm}
\end{figure}
%%%%%%%%%%%%%%%%%%%%%%%%%%%%%%%%%%%%%%%

In schemes that take a self-supervised approach \cite{hinton2022forward,ororbia2023predictive,lee2023symba,ahamed2023forward} (including \cite{giampaolo2023investigating} which mixes \cite{hinton2022forward} with local backprop), such as the forward-forward (FF) and predictive forward-forward (PFF) procedures, the focus is on distinguishing between a `positive sample', or data pattern, and a `negative sample', or an automatically generated adversarial pattern, through the use of two parallel inference pathways (instead of depending on a context label). The goal of these forms of credit assignment are to maximize the `goodness' for data points taken from the data distribution and to minimize the goodness for those that are drawn from outside of it adversarially. In an FF-based scheme, the total loss of a neural system can, like in signal propagation described earlier, be expressed as a sum of local functionals; however, these functionals focus on comparing and contrasting positive and negative activity patterns: $\mathcal{G}(\Theta) = \sum^L_{\ell=1} \mathcal{G}^\ell(\mathbf{z}^\ell,\mathbf{c}^\ell)$. We use the symbol $\mathcal{G}$ to imply `goodness', or the determination of the probability that the activities of a group of neurons is indicative that an incoming signal comes from the (training) data distribution (is an instance of a positive class) as opposed to those come from outside of it, e.g., confabulations, adversarial examples, etc. Formally, for any layer $\ell$ in a neural model, its `goodness probability' is dependent upon the sum of its squared activities compared to a threshold value $\theta^\ell_z$:
\begin{align}
    p(c=1;\mathbf{z}^\ell) &= \frac{1}{1 + \exp \big( -(\sum^{J_\ell}_{j=1} (\mathbf{z}^\ell_j)^2 - \theta_z ) \big)}, \label{eqn:goodness_prob}
\end{align}
where $p(c=1;\mathbf{z}^\ell)$ indicates the probability that the activity value at $\ell$ comes from the data distribution -- the positive class is indicated by $c = 1$ -- while the probability that the data does not is $p(c=0;\mathbf{z}^\ell) = 1 - p(c=1;\mathbf{z}^\ell)$. The goodness functional can then be formulated akin to binary class logistic regression:
\begin{align}
    \mathcal{L}(\mathbf{z}^\ell, c) = -\Big( c \log p(c=1;\mathbf{z}^\ell) + (1 - c) \log p(c=0;\mathbf{z}^\ell) \Big) \label{eqn:goodness_loss} 
\end{align}
where the binary label $c$ (a goodness label for sensory datapoint $\mathbf{x}$) is produced by generative process that synthesizes negative data samples. Patterns sampled from the data distribution are labeled as $c = 1$ while patterns sampled from the negative data generating process are labeled as $c = 0$; key to FF's effectiveness is the design or access to a useful negative data distribution, much as is done for related statistical procedures such as noise contrastive estimation \cite{gutmann2010noise}. % and negative sampling \cite{goldberg2014word2vec}.
With goodness defined, the process undertaken by FF learning entails first running a sensory sample, regardless of whether it is positive or negative, through the network according to the following slightly modified forward pass: $\mathbf{z}^\ell = \phi^\ell\Big(\mathbf{W}^\ell \cdot \text{LN}(\mathbf{z}^{\ell-1}) \Big)$ 
where $\text{LN}(\mathbf{z}^{\ell-1}) = \frac{\mathbf{z}^{\ell-1}}{(||\mathbf{z}^{\ell-1}||_2 + \epsilon)}$ denotes a (parameter-less) layer-wise normalization \cite{ba2016layer} operation and $\epsilon$ is a small numerical stability factor for preventing division by zero.\footnote{Layer normalization has been argued to be an important operation for forward-forward credit assignment as well as in biology \cite{carandini2012normalization}, as it, in effect, forces any given layer to use relative activities in judging goodness; the length of a layer of neural activity vector is corresponds to its goodness while its orientation is information that is passed on to the next layer \cite{hinton2022forward}.} Then, as soon as the data pattern has been propagated through layer $\ell$, its goodness label may be immediately fed in so as to trigger that layer's local functional, and the consequent synaptic update, as follows:
\begin{align}
    \Delta \mathbf{W}^\ell &= \Big( 2 \frac{\partial \mathcal{L}(\mathbf{z}^\ell, c)}{\partial \sum^{J_\ell}_{j=1} (\mathbf{z}^\ell_j)^2} \odot \mathbf{z}^\ell \Big) \cdot \big( \text{LN}(\mathbf{z}^{\ell-1}) \big)^T.
\end{align}
Desirably, the above FF plasticity rule facilitates the adjustment of the synapses of any particular neuronal layer that is independent of the others but, unlike Hebbian rules, requires knowledge across a group of neurons, given that a layer's goodness depends on the sum of squares of its set of activity values rather on that of any individual unit. Notice that the updates computed for a positive or negative data sample may done at any time and independently, facilitating a flexible asynchronous, variable scheduling of when negative datapoints are used versus positive ones (though current prominent efforts \cite{hinton2022forward,ororbia2023predictive} have focused on simultaneous adjustments based on balanced batches of positive and negative samples). Finally, it is important to point out that the flexibility of goodness-based (contrastive) learning lends itself to an important recurrent generalization that further resolves the locking problems inherent to backprop-based neural models \cite{hinton2022forward,ororbia2023predictive,ororbia2023learning}; concretely, this means the neuronal dynamics can take on an iterative form (similar to schemes of the previous two credit assignment families):
\begin{align}
    \mathbf{z}^\ell(t) = (1-\beta)\mathbf{z}^\ell(t-1) + \beta\bigg(
    \phi^\ell \Big( \mathbf{W}^\ell \cdot \text{LN}( \mathbf{z}^{\ell-1}(t-1) ) + \mathbf{V}^\ell \cdot \text{LN}( \mathbf{z}^{\ell+1}(t-1) ) - \mathbf{L}^\ell \cdot \text{LN}( \mathbf{z}^\ell(t-1) ) \Big)
    \bigg) \label{eqn:ff_state_update}
\end{align}
which is the generalized form \cite{ororbia2023predictive} of the model in \cite{hinton2022forward}. The dynamics of Equation \ref{eqn:ff_state_update} define not only bottom-up and top-down synaptic (local) pressures but also lateral/cross-layer synaptic parameters\footnote{Neurobiologically, the connectivity patterns of these lateral synapses can be designed to model inhibitory and self-excitatory cross-layer effects and neural competition patterns \cite{ororbia2023predictive,ororbia2023learning}.}; all of which can be locally adjusted with the same goodness functional $\mathcal{L}^\ell(\mathbf{z}^\ell,c)$. 
%% general context / other variations
Other self-supervised forward-only schemes either design local predictive contrastive losses \cite{illing2021local} (which further mixes a local prediction of neural activity to produce a modulatory signal per neuron) or mutual information preservation \cite{linsker1988self,lowe2019putting,xiong2020loco} functionals (with modern inspirations such as \cite{siddiqui2023blockwise,fournier2023preventing}, which mixes local backprop with decoupled, block-local self-supervised functionals).

%% temporal and spiking generalizations of forward-only
We remark that the principles underlying forward-only learning (e.g., learning without feedback neural circuitry) have been seen application in temporal credit assignment, such as learning time-varying sequence models of data as in recurrent networks. Several of these are often based on approximations/formulations of computing gradients in accordance with forward-mode differentiation \cite{williams1990gradient,jaeger2002tutorial,tallec2017unbiased,mujika2018approximating,ren2022scaling}. Finally, it is important to point out that forward-only algorithms, notably signal propagation and the recurrent form FF/PFF, have been generalized to operate in the context of spiking neural networks \cite{lee2023exact,kohan2023signal,ororbia2023learning}. This marks the beginning of an important move to investigation of such biological credit assignment schemes in the context of energy-efficient hardware; for example, an instantiation of FF learning has been demonstrated to operate well on an optical (fiber) platform \cite{oguz2023forward}.

\section{Discussion} % and Synthesis
\label{sec:discussion}

We motivated the general organization and direction of this survey with a single important question: where do the signals that induce credit assignment with respect to the individual processing units of neuronal systems come from and how are they created? Throughout this review, we have examined six general ways, centering around different algorithmic ``groupings'' or ``families'', to answer this question: 
implicit signals, explicit global signals, non-synergistic explicit local signals, and synergistic explicit local signals in three flavors -- discrepancy reduction, energy-based, and forward-only schemes.

We furthermore began this work with a treatment of backpropagation of errors (backprop), the popular workhorse algorithm behind the advances of deep learning today, and its core criticisms/central issues (with respect to neurobiology as well as practical considerations; see Section \ref{sec:backprop_problems}). In Table \ref{tab:problem_resolution}, we mark/annotate the six general credit assignment families with respect to how well each resolves each of the six prominent problems of backprop, if they do so at all; problem 5 (P5) is further subdivided into three sub-problems. A \xmark{ }denotes that not any of the algorithms within the family offer anything to resolve//alleviate the given problem, \cmark{ }indicates the opposite (all algorithms within the family resolve the problem), and \nmark{ }indicates that some, but not all, procedures offer a possible solution. In addition, we added three columns corresponding to three general properties that would prove most useful for any one family of biologically-plausible credit assignment schemes to have.\footnote{These three properties, by no means, do not make up an exhaustive list of all that might be desired, as this would largely depend on the research or engineering problem context and its constituent goals.} 
These three properties were:
\textbf{1)} whether the algorithm grouping was \emph{architecture-agnostic} (it did not require any particular structural forms or design patterns in the neural architecture it would conduct credit assignment with respect to); 
\textbf{2)} whether the credit assignment family could effectively (in an online manner, without unfolding) handle time-varying data sequences; and 
\textbf{3)} whether the algorithm grouping worked in the context of goal-directed behavior, decision-making, or control. 
The final column -- `All' -- is marked whether or not an item met all eleven criteria. Finally, notice that under each credit assignment family is a single row examining the one algorithm within it that satisfies the most criteria by itself:
\begin{itemize}[noitemsep,nolistsep]
    \item \textit{Implicit Signals} (\textbf{`Imp'}): two-factor Hebbian adaptation (Hebbian (2F));
    \item \textit{Explicit Global Signals} (\textbf{`EG'}): three-factor Hebbian adaptation (Hebbian (3F));
    \item \textit{Non-Synergistic Explicit Local Signals (\textbf{`NSEL'})}: synthetic local updates (SLU); 
    \item \textit{Synergistic Explicit Local Signals, Discrepancy-reduction} (`SEL:DR'): predictive coding/neural generative coding (PC/NGC); 
    \item \textit{Synergistic Explicit Local Signals, Energy-based} (\textbf{`SEL:EB'}): equilibrium propagation (EProp);
    \item \textit{Synergistic Explicit Local Signals, Forward-only} (\textbf{`SEL:FO'}): recurrent forward-forward/predictive forward-forward (RFF/PFF) learning.
\end{itemize}

%%%%%%%%%%%%%%%%%%%%%%%%%%%%%%%%%%%%%%%%%%%%%%%%
%% problems to resolve 
%% \cmark
%% \xmark
\begin{table}[!t]
    \begin{center}
    \begin{tabular}{|l | c c c c |c c c| c ||c c c|c|} 
     \hline
      &  \multicolumn{8}{|c|}{\textbf{Problem to Resolve}}  & \multicolumn{3}{|c|}{\textbf{Property}} & \textbf{All} \\ 
      & \textbf{P1} & \textbf{P2} & \textbf{P3} & \textbf{P4} & \multicolumn{3}{|c|}{\textbf{P5}} & \textbf{P6} & \textbf{AA} & \textbf{Time} & \textbf{Ctrl} & \\ %[0.5ex] 
     \textbf{Approach} & & & & & LP & FL & UL & & & & & \\
     \hline\hline
     Family: \textit{Imp} &  \cmark & \cmark & \xmark & \nmark & \nmark & \cmark & \nmark & \cmark & \nmark & \nmark & \xmark & \xmark \\ 
     \textit{Hebbian (2F)} & \cmark & \cmark & \xmark & \xmark & \cmark & N/A & \cmark & \cmark & \cmark & \cmark & \xmark & \xmark \\ 
     \hline
     Family: \textit{EG} & \nmark & \cmark & \xmark & \xmark & \xmark & \xmark & \nmark & \nmark & \cmark & \nmark & \xmark & \xmark \\ 
     \textit{Hebbian (3F)} & \cmark & \cmark & \xmark & \xmark & \cmark & N/A & \cmark & \cmark & \cmark & \xmark & \cmark & \xmark \\ 
     \hline
     Family: \textit{NSEL} & \cmark & \nmark & \xmark & \nmark & \nmark & \xmark & \nmark & \cmark & \cmark & \nmark & \xmark & \xmark \\ 
     \textit{SLU} & \cmark & \xmark & \xmark & \xmark & \cmark & \cmark & \cmark & \cmark & \cmark & \cmark & \xmark & \xmark \\ 
     \hline
     Family: \textit{SEL: DR} & \nmark & \nmark & \nmark & \nmark & \nmark & \nmark & \nmark & \xmark & \nmark & \nmark & \nmark & \xmark \\ 
     \textit{PC/NGC} & \cmark & \cmark & \cmark & \xmark & \cmark & \cmark & \cmark & \xmark & \xmark & \cmark & \cmark & \xmark \\ 
     \hline
     Family: \textit{SEL: EB} & \cmark & \nmark & \nmark & \nmark & \cmark & \nmark & \nmark & \cmark & \xmark & \nmark & \xmark & \xmark \\ 
     \textit{EProp} & \cmark & \xmark & \cmark & \xmark & \cmark & \cmark & \cmark & \xmark & \xmark & \xmark & \xmark & \xmark \\ 
     \hline
     Family: \textit{SEL: FO} & \cmark & \cmark & \nmark & \nmark & \nmark & \nmark & \nmark & \cmark & \cmark & \nmark & \nmark & \nmark \\ 
     \textit{RFF/PFF} & \cmark & \cmark & \cmark & \xmark & \cmark & \cmark & \cmark & \cmark & \xmark & \xmark & \xmark & \xmark \\ 
     \hline 
     All Families & \cmark & \cmark & \cmark & \cmark & \cmark & \cmark & \cmark & \cmark & \cmark & \cmark & \cmark & \cmark \\ 
     \hline
    \end{tabular}
    \end{center}
    \caption{\small{
    \textbf{What problems central to backprop-based learning (see Section \ref{sec:backprop_problems} for details) do different biologically-inspired credit assignment (bio-CA) algorithm families resolve?} 
    In each row, we consider one complete algorithm family as well as one procedure within it that satisifies the most criterion. For a family, \cmark{ }means all schemes within it satisfy the criterion, \nmark{ }means that at least one scheme within it clears the criterion, and \xmark{ } means that no scheme within it resolves the criterion. For a single algorithm, \cmark{ } means that it resolves the criterion directly whereas \xmark{ } means that it does not. 
    Problem names are abbreviated as follows:
    `P1' = global feedback pathway, `P2' = weight transport, `P3' = short-term plasticity, `P4' = constraint and sensitivity, `P5' = locality and locking, `P6' = inference-learning dependency. 
    Notice that `P5' has been broken down into three sub-problems: `LP' = local plasticity, `FL' = forward locking, `UL' = update locking. 
    Three extra criteria are introduced to determine if a scheme embodies particular properties; `AA' = architecture agnostic?, `Time' = naturally extends to sequences, `Ctrl' = facilitates behavioral/control-based learning.
    } \\
    \scriptsize{\textit{Note:} Some of the family naming labels have been abbreviated to: 
    `Imp' = Implicit (Signals), `S' = Synergistic, `NS' = Non-Synergistic,  `EG' = Explicit Global (Signals), `EL' = Explicit Local (Signals), while `DR' = Discrepancy Reduction, `EB' = Energy-based, and `FO' = Forward-only.}
    }
    \label{tab:problem_resolution}
    \vspace{-0.5cm}
\end{table}
%%%%%%%%%%%%%%%%%%%%%%%%%%%%%%%%%%%%%%%%%%%%%%%%

In light of the construction of Table \ref{tab:problem_resolution}, notice that, among all six credit assignment groupings, in aggregate, it is the \textbf{forward-only (FO) credit assignment class that meets all eleven criteria}, resolving all eight (sub)problems and meeting three useful generalization properties. However, this promising, positive synthesized result should be taken with a grain of salt -- no one single algorithm within the family resolves all eight problems, with RFF/PFF learning \cite{hinton2022forward,ororbia2023predictive} resolving only seven. When considering all six credit assignment families together, we further see, in the final row of Table \ref{tab:problem_resolution}, that all eleven criteria (eight problems and three properties) are resolved, indicating that positive progress has indeed been made despite the challenges facing research efforts in brain-inspired machine intelligence. In addition, this aggregate result hints that perhaps an important key to future success is in a confluence of algorithms/families -- a grand synergy or hybrid of approaches, where schemes work to compensate for each other's weaknesses by leveraging their particular strengths. 
We remark that comparing and contrasting this diverse set of credit assignment schemes usefully lends itself to highlighting important gaps that research in the domain might want to consider in future developments. For instance, we can see that control, which relates to behavioral and goal-oriented adaptation (reinforcement learning), poses the greatest difficulty; this is corroborated by the analysis in Table \ref{tab:algo_task_grouping}, as only the SEL:DR family has a stronger body of work in the area.\footnote{We remark that this state-of-affairs is likely a consequence of the work done on predictive coding that exploits its intimate relationship with the neurobiological process framework known as active inference \cite{friston2017graphical,parr2022active}.}

% next move to task table and analyze this briefly
Another way that we examined biologically-inspired credit assignment (bio-CA) was with respect to the types of (machine intelligence) contexts that algorithms have been adapted/applied to; the results of this treatment are in Table \ref{tab:algo_task_grouping}. Specifically, we grouped related works under each family based on (a non-exhaustive collection of) four challenging task types that test credit assignment scalability in different ways: `Vision' for evaluating whether any of a family's schemes have been generalized to work with convolutional/pooling structures that typically characterize computer vision models; 
`Time/Graph' to capture whether adaptations have been made to operate with inherently time-varying data sequences (e.g., video prediction tasks or language modeling) or relational knowledge modeling problems;  
`Control/RL' to determine if any algorithms in a family handle control/decision-making tasks (e.g., reinforcement learning or bandit problems); 
and `Spike' to specifically focus on whether generalizations of any algorithms in a group exist for spiking neural networks and neuromorphic architectures. Several patterns emerge from this investigation. First, many attempts have been to develop generalizations of various credit assignment routines in the category of `Time/Graph', with the primary bulk of studies focused on temporal modeling tasks, often attempting to deal with capturing distal correlations in sequences without requiring the unrolling inherent to backprop through time (in contrast, very few efforts have engaged with relational/graph problems). Not surprisingly, many studies have engaged with the scaling of various algorithms to vision problems, likely because the general supervised learning setup remains unchanged, e.g., classification/regression, requiring only changing the underlying neural transformations, e.g., the use of convolution or pooling-type operations instead of fully-connected synpatic structures. In terms of control, far less work exists to date in determining how various credit assignment schemes handle the difficult challenges posed by reinforcement learning (RL) -- this might partially relate to the fact that even current state-of-the-art backprop-based approaches struggle with sample efficiency, requiring many episodes (or replayed action/event sequences) to uncover useful structure about an agent environment as well as facing difficulty in successfully navigating the exploration-exploitation \cite{sutton2018reinforcement} trade-off central to RL. Thus, adding a typically more complex, noisier backprop-alternative, such as one of the schemes studied in this survey, would only serve to complicate an already complicated problem. Very little work in RL/control exists for most credit assignment families \emph{except} for the SEL:DR family, specifically due to predictive coding as mentioned before, and the EG family, largely due to three-factor Hebbian plasticity. 
Finally, and perhaps surprisingly, there is a great deal more work with respect to generalization towards the realm of spiking neural networks, arguably one of the most challenging spaces for application of biological adaptation schemes (despite these schemes seemingly being more of a fit for neurobiological contexts, at least with respect to neuronal architecture \cite{eliasmith2012large,eliasmith2013build}). The greater concentration of more successful developments in spiking systems include studies in four out of the six families, i.e., Imp, EG, SEL:DR, and EL:EB. We highlight the importance of this last observation, given that being able to successfully adapt spiking neuronal systems would have major implications in terms of energy efficiency; neuromorphic chips \cite{davies2018loihi} and related hardware platforms stand to impact edge-based computing and (neuro)robotics greatly \cite{hagras2004evolving,hwu2017self}.

%%%%%%%%%%%%%%%%%%%%%%%%%%%%%%%%%%%%%%%%%%%%%%%%%%%%%%%%%%%%%%%%%%%%
% task type organization of bio-CA
\begin{table}[!t]
    \centering
    \begin{tabular}{l | p{0.135\linewidth} | p{0.135\linewidth} | p{0.135\linewidth} | p{0.135\linewidth} } % 
    \toprule
    \textbf{CA Family} & \textbf{Vision} & \textbf{Time/Graph} & \textbf{Control/RL} & \textbf{Spike}\\ 
    \midrule
    %%%%%%%%%%%%%%%%%%%%%%%%%%%%%%%%%%%
    \textit{Implicit Target} & 
    \begin{tabular}[t]{@{}l@{}}
         \cite{gupta2021hebbnet,lagani2022comparing,amato2019hebbian} \\
         \cite{fukushima1982neocognitron,fukushima1988neocognitron}
    \end{tabular}
    & 
    \begin{tabular}[t]{@{}l@{}}
         \cite{brunel1996hebbian,wennekers2006language,jaeger2002tutorial} \\
         \cite{panda2017learning}
    \end{tabular}
     & 
    \begin{tabular}[t]{@{}l@{}}
         None
    \end{tabular}
     & 
    \begin{tabular}[t]{@{}l@{}}
         \cite{levy1983temporal,abbott2000synaptic,izhikevich2007solving} \\\cite{fremaux2016neuromodulated,legenstein2008learning,brzosko2017sequential} 
    \end{tabular}
    \\
    \midrule
    %%%%%%%%%%%%%%%%%%%%%%%%%%%%%%%%
    \textit{Explicit Global Target} % , Global Feedback
    & 
    \begin{tabular}[t]{@{}l@{}}
         \cite{han2020extension,pogodin2021towards,moskovitz2018feedback} \\
         \cite{teng2020layer}
    \end{tabular}
    & 
    \begin{tabular}[t]{@{}l@{}}
         \cite{liao2016bridging,han2020extension,launay2020direct} \\
         \cite{rombouts2015attention,kruijne2021flexible,evanusa2020deep} \\
         \cite{murray2019local}
    \end{tabular}
     & 
    \begin{tabular}[t]{@{}l@{}}
         \cite{izhikevich2007solving,salimans2017evolution,kusmierz2017learning} \\
         \cite{legenstein2008learning,fremaux2010functional,aljadeff2019cortical}
    \end{tabular}
     & 
    \begin{tabular}[t]{@{}l@{}}
         \cite{urbanczik2009reinforcement,samadi2017deep,fremaux2010functional} \\
         \cite{chase2009functional,pawlak2010timing,gerstner2018eligibility}
    \end{tabular}
    \\
    \midrule
    %%%%%%%%%%%%%%%%%%%%%%%%%%%%%%%%%%%%%%%%%%%%
    \begin{tabular}[t]{@{}l@{}}
         \textit{Explicit Local Target} \\
         Non-Synergistic
    \end{tabular}
    & 
     \begin{tabular}[t]{@{}l@{}}
         \cite{rueda2015supervised,tavakoli2021tourbillon,belilovsky2020decoupled} %\\
         %\cite{}
    \end{tabular}
    & 
    \begin{tabular}[t]{@{}l@{}}
         \cite{jaderberg2016decoupled,czarnecki2017understanding,murray2019local} %\\
         %\cite{}
    \end{tabular}
     & 
    \begin{tabular}[t]{@{}l@{}}
         None
    \end{tabular}
     & 
    \begin{tabular}[t]{@{}l@{}}
         \cite{ma2022deep,andrew2003spiking,neftci2017event} \\
         \cite{neftci2019surrogate}
    \end{tabular}
    \\
    \midrule
    %%%%%%%%%%%%%%%%%%%%%%%%%%%%%%%%%%%%%%%%
    \begin{tabular}[t]{@{}l@{}}
         \textit{Explicit Local Target} \\
         Synergistic \\ 
         Discrepancy-reduction
    \end{tabular}
    & 
     \begin{tabular}[t]{@{}l@{}}
         \cite{bartunov2018assessing,zee2022robust,ororbia2023backpropagation} \\
         \cite{ororbia2022convolutional,ernoult2022towards}
    \end{tabular}
    & 
    \begin{tabular}[t]{@{}l@{}}
        \cite{schmidhuber1989local,ororbia2017learning,wiseman2017training} \\
        \cite{manchev2020target,ororbia2018continual,jiang2022dynamic} \\
        \cite{santana2017exploiting,pinchetti2022,salvatori2022reverse} \\
        \cite{mali2021investigating,roulet2021target}
    \end{tabular}
     & 
    \begin{tabular}[t]{@{}l@{}}
         \cite{ororbia2022active,gklezakos2022active,ororbia2022backprop} \\
         \cite{rao2022active,ororbia2022cogngen,ororbia2022maze} \\
         \cite{zhong2018afa,gupta2021structural}
    \end{tabular}
     & 
    \begin{tabular}[t]{@{}l@{}}
         \cite{ororbia2019spiking,o2019training,mesnard2016towards} \\
         \cite{zhang2004single,rao2004hierarchical,n2023predictive}
    \end{tabular}
    \\
    \midrule 
    %%%%%%%%%%%%%%%%%%%%%%%%%%%%%%%%%%%%%%%%%%%%%%%%%
    \begin{tabular}[t]{@{}l@{}}
         \textit{Explicit Local Target} \\
         Synergistic \\ 
         Energy-based
    \end{tabular}
    & 
     \begin{tabular}[t]{@{}l@{}}
         \cite{gao2016novel,ernoult2020continual,laborieux2021scaling} \\
         \cite{norouzi2009stacks,lee2009convolutional,lee2011unsupervised}
    \end{tabular}
    & 
    \begin{tabular}[t]{@{}l@{}}
         \cite{taylor2006modeling,sutskever2008recurrent,mittelman2014structured} \\
         \cite{gan2015deep,niehues2012continuous,hinton1995helmholtz} \\
         \cite{kendall2020training}
    \end{tabular}
     & 
    \begin{tabular}[t]{@{}l@{}}
         \cite{crawford2016reinforcement,widrich2021modern,kubo2022combining} %\\
         %\cite{}
    \end{tabular}
     & 
    \begin{tabular}[t]{@{}l@{}}
         \cite{hinton1999spiking,xie2000spike,mesnard2016towards} \\
         \cite{zoppo2020equilibrium,neftci2014event,martin2021eqspike}
    \end{tabular}
    \\
    \midrule 
    %%%%%%%%%%%%%%%%%%%%%%%%%%%%%%%%%%%%%%%%%%%%%%%%
    \begin{tabular}[t]{@{}l@{}}
         \textit{Explicit Local Target} \\
         Synergistic \\ 
         Forward-only
    \end{tabular}
    & 
     \begin{tabular}[t]{@{}l@{}}
         \cite{lowe2019putting,hinton2022forward,kohan2023signal} \\
         \cite{xiong2020loco,siddiqui2023blockwise,fournier2023preventing}
    \end{tabular}
    & 
    \begin{tabular}[t]{@{}l@{}}
         \cite{steil2004backpropagation,hinton2022forward,paliotta2023graph} \\
         \cite{williams1990gradient,tallec2017unbiased,mujika2018approximating}
    \end{tabular}
     & 
    \begin{tabular}[t]{@{}l@{}}
         \cite{lemmel2023real} %\\
         %\cite{}
    \end{tabular}
     & 
    \begin{tabular}[t]{@{}l@{}}
         \cite{ororbia2023learning,kohan2023signal,lee2023exact} \\
         \cite{oguz2023forward}
    \end{tabular}
    \\
    \bottomrule
    \end{tabular}
    \caption{\small{\textbf{Biological Credit Assignment as Applied to Task Settings.} 
    Overview of neurobiologically-inspired credit assignment (bio-CA) schemes organized by task setups where they have been utilized/evaluated, i.e., `Vision' refers to scaling up to problems applied to natural images, `Time / Graph' refers to temporal or relational modeling tasks, `Control' refers to reinforcement learning control tasks, and `Spike' refers to generalizing the bio-CA scheme to adapt spiking neural networks.
    }}
    \label{tab:algo_task_grouping}
    \vspace{-0.5cm}
\end{table}
%%%%%%%%%%%%%%%%%%%%%%%%%%%%%%%%%%%%%%%%%%%%%%%%%%%%%%%%%%%%%%%%%%%

Taken together, the synthesis of Tables \ref{tab:problem_resolution} and \ref{tab:algo_task_grouping} demonstrates both promise as well as  directions for growth. While biological credit assignment has come a long way since some of its earliest mathematical/algorithmic incarnations \cite{hebb1949organization,rosenblatt1958perceptron}, giving rise to the design of computational agents processing various kinds of information sources -- ranging from natural images/videos to event-based streams -- much work remains to bridge the performance gap between current state-of-the-art biological approaches to credit assignment and backprop-based DNNs \cite{bartunov2018assessing}. Furthermore, more concerted effort will be needed to facilitate the construction of powerful, energy-efficient neuro-mimetic machines, as desired by the field. One of the domains that we argue might be most important will be the realm of RL/control, as this often encourages/lends itself to the design of modular neural agent architectures, often ones that share many similarities with the cognitive architectures typically crafted \cite{anderson1998actr,laird2012soar,sharma2016large,ororbia2022cogngen} in Cognitive Science.

%%%%%%%%%%%%%%%%%%%%
% bio-CA Algorithmic Connections and Equivalences to backprop and others
\noindent
\textbf{Equivalence to Backprop.} % tp to gauss-newton
A notable theoretical direction that has been pursued in service of neuro-mimetic credit assignment has with respect to convergence properties, comparing the results of particular brain-inspired adaptation schemes to the capabilities of backprop-trained ANNs \cite{lee2015difference,ororbia2019biologically,bartunov2018assessing,choromanska2019beyond,kunin2020two,ororbia2022ngc,ororbia2023backpropagation}. This trend in theoretical effort, particularly for the case of supervised learning setups, has lead to the establishment of valuable equivalencies between some approaches in the SEL algorithms, such as PC, and backprop. In effect, this work can be viewed as establishing that the credit assignment conducted by a non-backprop routine ``biologically approximates'' the backprop algorithm, i.e., treating backprop as a sort of gold standard given its modern success in catalyzing the progress made in DNN-driven systems \cite{lecun2015deep}. Specifically, \cite{whittington2017approximation, millidge2020predictive} have shown, under particular conditions (such as small output error), that PC approximates the weight updates yielded by backprop when training MLPs and for computational graphs in general; furthermore, this approximate equivalency can be demonstrated if a temporal scheduling of PC's weight updates, with respect to the underlying neural activity dynamics, is carefully crafted \cite{song2020brain,salvatori2022reverse}. Similar equivalencies between biologically-inspired credit assignment and backprop have also established for other algorithms in the families reviewed in this survey; this notably includes target propagation \cite{le1986learning,lecun1991new,bengio2014auto,lee2015difference,bengio2020deriving,millidge2022backpropagation,roulet2021target}, contrastive Hebbian learning \cite{movellan1991contrastive,xie2003equivalence}, and equilibrium propagation \cite{scellier2017equilibrium,millidge2022theoretical,millidge2022backpropagation}. A key step underpinning many established theoretical equivalencies is to show that the updates to parameters calculated by a biological learning process \cite{xie2003equivalence,bengio2015stdp,bengio2017stdp,bengio2016feedforward,scellier2017equilibrium,scellier2018generalization} align within $90^{\circ}$ of the direction suggested by the gradients computed by backprop \cite{lillicrap2016random,nokland2016direct}. 
Other results from this line of theoretical research include equivalences between bio-CA algorithms themselves \cite{oreilly_biologically_1996,ororbia2018conducting,scellier2017equivalence,scellier2018generalization} -- for example, showing that predictive coding and contrastive Hebbian learning (and equilibrium propagation) can be equivalent to one another \cite{millidge2022backpropagation} or target propagation is (approximately) a form of Gauss-Newton optimization \cite{bengio2020deriving,meulemans2020theoretical}. In \cite{hinton2022forward,ororbia2023predictive}, relationships were established between forward-forward learning and contrastive Hebbian learning, contrastive divergence, and predictive coding.  
%%%%%%%%%%%%%%%%%%%%
%% Transition paragraph
%In light of the above families of algorithms and considering how they answer the question of where targets for learning might come from, we now end this survey by turning our attention to two central issues that face all of them, which will require the attention of the research community that drives work on neurobiologically-motivated algorithms at large.

% next briefly move into bio-behavior and physiological realism (or move this to unifying framework?
\noindent
\textbf{Biological Behavioural versus Physiological Realism.} 
%A learning algorithm developed that more faithfully embodies  natural neural computation tends to address one or more of the above issues related backprop, though none currently exist that address all of these issues, let alone completely, despite many recent efforts at doing so \cite{bengio2015towards,scellier2016towards}. 
To lay additional foundation for future progress in biologically-inspired credit assignment, and in the spirit of arguments put forth in \cite{bartunov2018assessing,gupta2022bio}, we next highlight an important consideration for research in the area. 
In particular, when developing biologically-motivated algorithms and frameworks and studying how they scale, we argue that design and development will need to account for: 
\textbf{1)} the sufficiency of the learning algorithm in question (in terms of a task or class/set of tasks), 
\textbf{2)} the impact that incorporating biological constraints into a neural system has (with respect to task performance or the kinds of policies uncovered for solving the task), and 
\textbf{3)} the necessity of other mechanisms for neural adaptivity (beyond those that explicitly work in service of credit assignment/plasticity). 
These three complementary notions indicate a possible pathway for developing more unified, effective benchmarks for evaluating and contextualizing biological credit assignment algorithms with respect to one another. Furthermore, these notions shift the focus from a more strict characterization in terms of and comparison to backprop itself (though this type of comparison will still likely play an important role in the near future) and towards an examination of the behavior that biological credit assignment would facilitate when tackling certain problems and task contexts -- a particular class of tasks/problems the would complement this shift, as suggested by our earlier synthesis (Table \ref{tab:problem_resolution}), is that of reinforcement learning and control, further motivated empirically by the gap we observed in the literature through our synthesis tables (Table \ref{tab:algo_task_grouping}). This particular emphasis might prove useful as development in RL often entails crafting agent architectures that have multiple interacting modules -- this would test the flexibility and computational expense of various credit assignment schemes and might further promote hybrid designs that enjoy a heterogeneity of biological credit assignment processes as opposed to the use of one single scheme as the uniform explanation of learning across neuronal units within an architecture. Furthermore, tackling RL will usefully leverage the more substantial accumulation of work in computer vision and temporal sequence learning (as we saw, in Table \ref{tab:algo_task_grouping}, more efforts have provided contributions towards these) and notably lead to intersections with important complementary research threads in cognitive science and neurorobotics. This might also motivate work to fill the gap in relational, graph-centered learning (or `knowledge-graph learning'), as we observed there is far less work in this area as compared to temporal modeling. %% this also connects to embodiment and enactivism (via RL/AIF), affecting neurorobotics in tandem with energy-efficient neuromorphics -- could connect to interroceptive AI and mortal compute -- this could connect to the development of more software libraries that standardly implement these (can cite ngc-learn as one) as well hardware emulators/platforms to test things out

In complement to the above behavioral/task-family central perspective, a valuable shorter-term direction would involve cross-family empirical analyses and (simulation) studies as well as further cross-algorithm theoretical establishments, ideally for more than two credit assignment schemes, similar in spirit to the work done in \cite{millidge2022backpropagation}. % (which provided a theoretical bridge between three algorithms; although this only connects two families, i.e., SEL:DR and SEL:EB). 
Empirical research into algorithmic scalability has so far revealed mixed results, either in the negative \cite{bartunov2018assessing}, or in the positive \cite{xiao2018biologically}. However, very few of the current studies have done proper, comprehensive comparisons of algorithms across families (though a few minor efforts exist \cite{bartunov2018assessing,ororbia2019biologically,zee2022robust}); most modern studies typically narrow down to a more recent, small subset of schemes (with typical choices being those from the feedback alignment family as benchmark candidates). %
%We emphasize one message put forth at the start of \cite{bartunov2018assessing}, specifically emphasizing that all three aspects of it need to be pursued: 
%1) existing algorithms need to be modified/added to/optimized to account for how the brain actually learns, 2) research needs to continue in searching for newer, more physiologically realistic learning procedures that can scale up to large-scale, complex tasks, and 3) we must appeal to other aspects of the brain's adaptive capacities to account for the fact that humans can perform well on complex, data-starved tasks.  
In essence, studying larger aggregates of algorithms across a greater diversity of credit assignment families will help in teasing out further strengths and weaknesses inherent to each scheme beyond their comparison to backprop; this might possibly aid us, if more tightly integrated with results in computational cognitive neuroscience, in further understanding the intricacies of how brains and natural neuronal networks conduct learning, with both physiological and behavioral elements accounted for and evaluated properly. 
Furthermore, it will become ever more important to more deeply characterize the failure cases of each algorithm in turn, e.g.., examining stability and robustness of the learning process \cite{ororbia2019biologically,mali2021investigating}, as well as how the underlying neural architectural complexity (e.g., number of layers/depth, types of connectivity patterns, design modularity and the resulting information communications that will be needed) interacts with the efficacy of various credit assignment process. In addition, investigating the effect of adversarial perturbations has on credit assignment scheme could prove useful in uncovering further strengths/weaknesses \cite{farinha2023efficient}. % insect fruit-fly model?

% \textcolor{red}{Desiradata for benchmarking/evaluating bio-CA algos} 
\noindent
\textbf{Towards A Unified Story of Neurobiological Credit Assignment.} %Set of Principles: } 
As we have seen throughout this review, algorithms within different credit assignment families address issues/criticisms of backprop to varying degrees while addressing other aspects of neurobiology along the way.
As a result, the next question that arises is: what other criterion need to be addressed by current and future neurobiological-motivated algorithms? Criteria have been offered \cite{oreilly1998sixprinciples,bengio2015towards,whittington2017approximation,ororbia2022ngc}, but no one clear consensus seems to exist, largely due to the fact that the world of brain-inspired credit assignment, despite the many efforts across several decades, is still rather ``young'' in comparison to other lines/threads of inquiry in machine intelligence research. On the path to developing a complete theory of learning, credit assignment, and inference, it will be important to develop a more complete set of standards with which we can fully judge and compare algorithms and/or architectures, with complementary benchmarks and problems/tasks that rigorously and thoroughly test these criteria. In particular, as discussed above, it will be crucial to move beyond supervised learning, specifically classification (on databases such as MNIST, CIFAR-10, and ImageNet), when designing bio-CA schemes for neuro-mimetic agents as this will require researchers and engineers to consider biologically-plausible architectural design elements, such as lateral and cross-layer recurrent connectivity patterns\footnote{The interaction between distributed representations and lateral competition, via cross-neuronal excitation/inhibition, is among the most important known biological mechanisms that could benefit neuro-mimetic computational models the most, facilitating natural emergent sparsity \cite{olshausen1997sparse} (fewer neuronal units are active at any time step), which is useful for energy efficiency, as well as possibly aiding in combating the grand challenge of catastrophic forgetting in ANNs \cite{french1999catastrophic}. Models that compromise between a winner-take-all style approach and full lateral inhibition appear to work well in practice \cite{majani1989induction,fukai1997simple}, though much work remains to integrate them properly and completely in probabilistic, neural frameworks.} 
and mechanisms for memory storage and knowledge consolidation. 

Constructing coherent cognitive architectures that combine one or more of the algorithms from the credit assignment families reviewed in this work might prove invaluable in achieving the above goals -- an important continuance of the historical aspirations of classical connectionism and parallel distributed processing \cite{rumelhart1986parallel} -- where inspiration can be drawn from early work in neural cognitive architectures such as Leabra \cite{oreilly2016leabra}, CogNGen \cite{ororbia2022cogngen}, or Spaun/Nengo \cite{eliasmith2013build}. In complement to this, further results in computational and cognitive neuroscience could aid to shape and corroborate the principles developed in brain-inspired machine intelligence models, possibly aiding researchers and theoreticians in crafting a story and spectrum between the higher-level, coarser-grained dynamics simulated by rate-coded models, learned under different forms of bio-CA, and the finer-grained dynamics underwriting intricate systems of spiking neuronal cells. Interesting questions arise from this cognitive systems-level premise: how would a complex dynamical system that combines different credit assignment schemes function and how might it generalize differently than systems that use only a single scheme uniformly? Beyond this, where and how should different credit assignment schemes be applied (in what ``artificial brain region'' would any specific approach work best)? For example, SEL:EB schemes, e.g., equilibrium propagation, might be better suited to adaptation in memory modules, which might operate a slower time-scale of adaptation, whereas SEL:DR, e.g., predictive coding, and EG schemes, e.g., multi-factor Hebbian plasticity, might be useful for faster processing time-scales. It is known 
that Hebbian correlational-learning can usefully constrain task-learning \cite{hancock1991biologically,oreilly1998sixprinciples}, helping bias a neural circuit's distributed representations by incorporating additional co-occurrence information that might further improve task-specific performance; for instance, it has been shown that even backprop benefits from being combined with Hebbian update rules \cite{soo1996merging} (offering a form of implicit regularization, at the very least). 

An emergent pattern or principle that we observed throughout this review of various algorithms is what we label here as \emph{the degree of entanglement between inference and learning}. %
% degree of entanglement --> lead to embodiment and enactivism
% This makes me think that defining a set of principles/criteria means we need to also think about architecture and algorithm simultaneously - almost all of the explicit-target algorithms involves extending the architecture in some way to directly or indirectly embody these principles... --> THIS ALSO CONNECTS TO ARCH VS UPDATE and how the two are intimately related -> can connect to compartment neurons and two sites of synaptic integration too (maybe 3-factor Hebb rule??)
Specifically, this refers to the distinction that is typically applied between inference, learning, and structure (in this survey's case, the computational structure or arrangement of neuronal units within a communication network) in machine intelligence. Many algorithms observed throughout this review, such as those within the discrepancy reduction (predictive coding, target propagation, etc.) and energy-based (sub-)families (contrastive Hebbian learning and equilibrium propagation), introduce additional architectural components such as feedback weights, error units, decoding mechanisms, additional cost functionals, etc. This aspect of biological learning highlights that credit assignment does not need to be regarded as a separate computational process that ``floats'' outside of the neuronal processing architecture -- the form and arrangement of the neuronal processing units, as well as the synaptic information message passing pathways connecting them, can be intertwined with the learning/plasticity process itself. In many instances, by imposing particular architectural constraints on a system's neural computation, e.g., local top-down and bottom-up forms of feedback\footnote{The cortex has been shown to feature quite a bit of (top-down-bottom-up) bidirectional connectivity \cite{white1989cortical,felleman1991distributed} and has been argued to facilitate constraint-satisfaction processing, where low-level (perceptual) and high-level (conceptual) constraints can be brought to be bear simultaneously in the context of neuronal processing. This top-down-bottom-up notion has been shown to explain a variety of cognitive phenomena \cite{mcclelland1981interactive}, e.g., higher-level word processing can influence lower-level letter perception, and has been shown to aid in resolving ambiguity in visual inputs \cite{vecera1998figure}.}, simple, efficient, and local synaptic adjustment rules may be readily employed (as in contrastive Hebbian learning or the recurrent form of forward-forward learning). This somewhat contradicts the properties that we evaluated credit assignment schemas for in Table \ref{tab:problem_resolution} (and possibly works against resolving inference-learning dependency problem); we remark that, although architectural agnosticism might be a valuable property, providing maximal freedom to the experimenter or engineer when constructing their model for a given problem, giving up some degree of this freedom appears to bring with it easy ways of emulating short-term plasticity as well as brain-like forms of parallelism (e.g., layer-wise parallelism) in not only the learning process but also in the very inference itself. Take, for example, predictive coding -- such a scheme offers parallelism across layers in both the inference and synaptic adjustment steps, resolving the forward-locking and upate-locking problems cleanly, due to the message passing induced by its feedback synaptic structure. Characterizing this entanglement further could prove useful for biological credit assignment and brain-inspired machine intelligence research. 
This opens the door to energy-efficient implementations of asynchronous computation \cite{belilovsky2021decoupled,salvatori2022incremental}. If our goal is to develop intelligent agents that generalize in the ways that humans do, then computationally formalizing and integrating structural/architecture properties of the human brain is nearly if not as important as computationally formalizing and integrating properties of the brain's synaptic plasticity and adaptation. We note, however, that some particular aspects of architectural agnosticism would likely remain even in the neural circuitry that features some entanglement of its structure with learning/inference -- many of the credit assignment schemes we reviewed did not require the neuronal activities to be differentiable themselves, such as in forward-only learning, which could, in heterogeneous hybrid systems, permit the adaptation of black-box building blocks \cite{hinton2022forward} or `sub-circuits'.

Going forward, the process of crafting neurobiological forms of neuronal computation and learning will need to strike a balance between how closely the details of actual neurobiological circuitry and biochemical/electrophysiological processes are modeled and how much effort is placed in constructing fast, efficient methodology for practical applications. This leads us to ask the following questions: 
\emph{How much neurobiological realism do we need to sufficiently generalize? How much is too much?}  
If we lean too heavily towards exactly modeling (neuro)biology, which carries with it tremendous value in computational cognitive neuroscience, % for simulation study (as is the case for spiking neural networks, for example), 
we might risk constructing architectures and learning procedures that are far too slow to simulate, restricting how well we may be to exploit the unreasonable effectiveness \cite{halevy2009unreasonable} of large-scale datasets and streams. On the other hand, if we lean too heavily towards building algorithms that are only meant to be fast and handle the demands of industrial applications, we risk eschewing important neurophysiological mechansims that might be invaluable in allowing our artificial neural systems to generalize in the right ways, e.g., generalizing from small samples as humans do, gradually forgetting old task information instead of catastrophically forgetting it, and transferring knowledge across task domains to work with datasets with little to no human annotations/labels. This balance between biophysical realism/faithfulness will be critical to consider going forward in brain-inspired machine intelligence research and might possibly span concurrent, complementary lines of inquiry (and related communities of theoreticians, scientists, and engineers), guided by each thread's particular scientific directives and questions. 
%It might be valuable to develop methods to allow researchers to take a particular aspect of neurobiology, such as an important chemical process, and formalize it in a useful computational manner without trying to simulate its exact particularities and properties.
%%%%%%%%%%%%%%%%%%%%%%%%%%%%%%

\section{Conclusion}
\label{sec:conclusion}

A core mission of brain-inspired machine intelligence, and modern initiatives such as ``NeuroAI'' \cite{richards2019deep,payeur2021burst,zador2023catalyzing,momennejad2023rubric}, is to construct a unified theoretical and computational framework for the neural processing and dynamics that underwrite inference and credit assignment in the brain. Achieving this aspiration would lay the groundwork for constructing ``thinking machines'' \cite{turing1950computing} and artificial general intelligence, based on artificial neural systems, capable of emulating the complex, energy-efficient behavior of humans and animals while further providing scalable models useful for computational neuroscience and cognitive science. With the hope of foregrounding research in biologically-inspired, neuro-mimetic credit assignment, this survey examined and synthesized the scientific efforts, ranging from the theoretical to the empirical, of countless researchers who have made important contributions to this grand overarching objective. In essence, the pursuit of biologically-inspired credit assignment entails looking to neurobiology and cognitive neuroscience for inspiration and a grounding in to how things might operate in the neuronal circuitry that makes up brains. As presented in this review, this kind of credit assignment can be broken down into several families or classes centering around answering one of many possible questions: where do the signals that drive learning, or credit assignment, in neuronal networks come from and how are they produced? 
Studying this question resulted in six general families of algorithms and procedures, each embodying different answers to this central question. We furthermore examined the mechanics of and characterized several prominent issues/criticisms \cite{grossberg1987competitive,zipser1988back,crick1989recent,oreilly2000computational,harris2008stability,urbanczik2009reinforcement} underlying backpropagation of errors \cite{linnainmaa1970representation,rumelhart1985feature} (backprop), the credit assignment scheme that has served as the catalyst behind the ever-growing set of benchmark-breaking, state-of-the-art results in modern-day machine intelligence \cite{lecun2015deep}. In light of these, we examined how the current body of biological credit assignment research resolves these issues/criticisms and furthermore examined what problems in artificial intelligence these schemes have been applied to. %the available results related to each in tackling particularly challenging problem domains in artificial intelligence.

Based on this review of ``backprop-free'' biologically-inspired approaches to credit assignment and neural computation, one might ask which family or which procedure is a possible ``master algorithm'' \cite{domingos2015master} that underlies the brain's adaptive capabilities? To this question, the answer, as one might suspect based on the synthesis offered by this work, is that no single scheme or family developed, to date, offers the perfect, viable candidate. Furthermore, no one scheme completely and resoundingly resolves all of the issues that backprop has been criticized for, especially when it comes to neuro-physiological plausibility. Nevertheless, the ``galaxy of algorithms'' that currently exist, of which we suspect are only but a subset of a much larger space of possibilities, represent promising alternatives to the very learning algorithm that the domain of deep learning has become so well-accustomed to using. As further argued in this review, it might be the combination and synchrony of this heterogeneous set of schemes, i.e., in the form of general neuro-mimetic cognitive architectures \cite{eliasmith2012large,oreilly2016leabra,roelfsema2018control,ororbia2022cogngen,salvatori2023brain,bernaez2023incorporate} further engendering the design of embodied and enactive agents \cite{friston2008hierarchical,friston2010free,parr2022active}, that might generate new pathways towards far greater energy efficiency, compatibility and viable integration into neuromoprhic edge-computing and low-energy analog platforms as well as robotic systems \cite{davies2018loihi,grollier2020neuromorphic,kumar2022dynamical,yi2023activity}, and different, more neurobiologically-motivated forms of generalization than what today's intelligent technological artifacts are capable of. %, ranging from natural pattern completion abilities, robustness to noisy/adversarial data patterns, out-of-sample zero-shot generalization, and online temporal credit assignment. 
Besides offering mathematical formulations of ``tricks'' used by the brain that might benefit current backprop-based machine learning \cite{marblestone2016toward}, the development of brain-inspired approaches to machine intelligence might open the door to powerful intelligent computational tools that not only tackle machine learning tasks that modern-day deep neural networks struggle with but also serve as useful, credible hypotheses for credit assignment \cite{bengio2017stdp} in the brain that could be validated and extended from a neuroscience and cognitive sciences points-of-view.

\subsection*{Acknowledgments}
\small{
We would like to thank Alexander Ororbia (Sr.) for reviewing and providing comments for an early draft of this article. 
Furthermore, we acknowledge the support of the Cisco Research Gift Award \#26224.
}

\bibliographystyle{acm}
\bibliography{ref}

\newpage
\section*{Supplemental Material}

\subsection*{Potential Research Questions}
\label{sec:questions}

Other potentially useful research questions related to research in biologically-inspired credit assignment, ranging from scientific to engineering considerations, include the following:
\begin{itemize}[noitemsep,nolistsep]
\item When does a credit assignment scheme only apply to a certain level of abstraction of the neural (assembly) dynamics? What changes or modifications are needed to make an algorithm operate at different time-scales (e.g., moving from rate-coded dynamics to spiking dynamics and vice versa)?  %\textcolor{red}{single (cortical) neurons as deep neural networks themselves \cite{beniaguev2021single}}
\item Does a credit assignment scheme or process work uniformly well across different types of learning setups, e.g., supervised, semi/weakly-supervised, unsupervised, and reinforcement learning? Which cases/setups highlight weakness or strengths of specific algorithm families/procedures? What algorithm types complement each other in hybrid/combined setups? %Or are some algorithms particularly suited to certain types of learning, indicating certain pairings/combinations?
\item To what degree of entanglement between (computational) neural architecture and inference/credit assignment is needed to result in reasonable generalization in certain task contexts? What flexibility must be ceded to obtain particular properties, such as energy efficiency or massive parallelism, and how is generalization impacted by different degrees of entanglement?
%To what level is the lack of sufficient neurobiological plausiblity due to the update rule and due to the design of the architecture?
\item What other criterion drawn from neuro-physiology/biology would help to improve the stability and generalization ability of artificial neural systems and their processes of credit assignment?
\item How do different update computations, produced by different biological credit assignment schemes, mix or interact with outer optimization procedures, e.g., RMSprop \cite{tieleman2012lecture}, Adam \cite{kingma2014adam}, etc.?
%\item What issues must be overcome for any one credit assignment to operate effectively with both rate-coded and spiking neuronal dynamics?
\item %The Issue of Time: 
What modifications or generalizations are needed for a credit assignment scheme to efficiently and effectively handle time-varying information, particularly without unrolling or unfolding in time? %\cite{liu2022beyond}
\end{itemize}

\newpage 
\subsection*{Terms and Symbology}
In Tables \ref{table:symbology} and \ref{table:terms}, we provide a table that collects and briefly defines several acronyms, abbreviations, key symbols, and operators used throughout the survey.

\begin{table*}[!ht]
\footnotesize
\begin{center}
\caption{Key symbols and operators.}
\label{table:symbology}
%\rowcolors{2}{gray!25}{white} % Add alternating row colors
\begin{tabular}{ll} 
 \toprule
 \textbf{Item} & \textbf{Explanation} \\ 
 \midrule
 $\cdot$ & Matrix/vector multiplication \\
 $\odot$ & Hadamard product (element-wise multiplication) \\
 %$\oslash$ & Element-wise division \\
 $\mathbf{z}_j$ &  $j$th scalar of vector $\mathbf{z}$ \\
 $||\mathbf{v}||_2$ & Euclidean norm of vector $\mathbf{v}$ \\
 $L$ & Number of layers \\
 $\mathbf{z}^{\ell}$ &  A layer $\ell$ of neural activity values\\
 %$\mathbf{e}^{\ell}$ &  Error/mismatch signal at layer $\ell$ \\
 $\phi^\ell()$ &  Nonlinear activation/transfer function applied to a vector/matrix \\
 $\mathcal{L}$ & Cost/objective functional \\
 $M$ & Neuromodulatory signal \\
 $\mathsf{T}$ &  Transpose operation \\
 $\mathbf{W}^\ell$ & A matrix of synaptic weight values \\
 $E()$ & Energy functional \\
 $\mathcal{F}()$ & Free Energy functional \\
 \bottomrule
\end{tabular}
\vspace{0.25cm}
\caption{Key abbreviations/acronyms and definitions.}
\label{table:terms}
%\rowcolors{2}{gray!25}{white} % Add alternating row colors
\begin{tabular}{llll} 
 \toprule
 \textbf{Item} & \textbf{Explanation} & \textbf{Item} & \textbf{Explanation}\\ 
 \midrule
 2F & Two-Factor & 3F & Three-Factor\\
 Backprop & Backpropagation of errors & CD-K & Contrastive divergence (K steps)\\
 CHL & Contrastive hebbian learning & SAP & Stochastic approximation procedure\\
 EProp & Equilibrium propagation & SLU & Synthetic local updates \\
 PC & Predictive coding & DBN & Deep belief network \\
 RFF & Recurrent forward-forward & RBM & Restricted Boltzmann machine\\
 PFF & Predictive forward-forward & DBM & Deep Boltzmann machine\\
 SNN & Spiking neural network & ANN & Artificial neural network \\
 STDP & Spike-timing-dependent plasticity & DNN & Deep neural network \\
 R-STDP & Reward-modulated STDP & SA & Simulated annealing \\
 LRA & Local representation alignment & GeneRec & Generalized recirculation \\
 SigProp & Signal Propagation & rec-LRA & Recursive LRA\\
 Targetprop & Target propagation & DTP & Difference target propagation  \\
 MLP & Multilayer perceptron & HSIC & Hilbert-Schmidt independence criterion \\
 KP & Kolen-Pollack method & WTA & Winner-take-all\\
 RFA & Random feedback alignment & DFA & Direct feedback alignment \\
 SOM & Self-organizing map & ART & Adaptive resonance theory\\
 RNN & Recurrent neural network & BPTT & Backprop through time \\
 SGD & Stochastic gradient descent & CIFAR10, SVHN, ImageNet & Natural image datasets/benchmarks \\
 RMSprop & Root mean squared propagation & Adam & Adaptive moment estimation\\
 SLU & Synthetic Local Updates & HISC &  Hilbert-Schmidt indepedence criterion\\
 EFP & Error Forward Propagation & PEPITA &
      \begin{tabular}[t]{@{}l@{}}
         Present the error to perturb the input \\
         to modulate activity
       \end{tabular}\\
 Neuromod & Neuro-modulation &  Competitive & Competitive (Hebbian) learning\\
 DRTP & Direct random target projection & WM & Weight Mirrors\\
 \bottomrule
\end{tabular}
\end{center}
\end{table*}

\end{document}